\documentclass[times,twocolumn, final,preprint]{elsarticle}

\usepackage{framed,multirow}

\usepackage{amssymb}
\usepackage{latexsym}

\usepackage{url}

\usepackage{hyperref}

\usepackage{array}
\usepackage{graphicx}
\usepackage{amsmath}
\usepackage{amsfonts}
\usepackage{hyperref}
\usepackage[table]{xcolor}
\usepackage{subcaption}
\usepackage{multirow}
\usepackage{microtype}
\usepackage{soul}
\usepackage{xspace}
\usepackage{booktabs}
\usepackage{tablefootnote}
\usepackage[bottom]{footmisc}
\usepackage{adjustbox}
\usepackage{tabularx}
\usepackage[section]{placeins}
\usepackage{balance}
\usepackage{stfloats}
\usepackage[para]{threeparttable}
 \newcommand{\outline}[1]{} % use this to hide comments

\newcommand*{\p}{\emph{p}\xspace}
\newcommand*{\np}{\emph{np}\xspace}
\newcommand*{\vfep}{VF$\mathrm{^{EP}}$\xspace}
\newcommand*{\vfiz}{VF$\mathrm{^{IZ}}$\xspace}

\graphicspath{{figures}}

\journal{journal}

\begin{document}

%\verso{Pietro Astolfi \textit{et~al.}}
\begin{frontmatter}

\title{Supervised Tractogram Filtering using Geometric Deep Learning\tnoteref{mytitlenote}}%
\tnotetext[mytitlenote]{Extended version of MICCAI publication \cite{astolfi_tractogram_2020}.}

\author[fbk,iit,cimec]{Pietro Astolfi}
\author[fbk]{Ruben Verhagen}
\author[gin]{Laurent Petit}
\author[fbk,cimec]{Emanuele Olivetti}
\author[apss,cimec]{Silvio Sarubbo}
\author[nnaisense]{Jonathan Masci}
\author[fbk]{Davide Boscaini}
\author[fbk,cimec]{Paolo Avesani\corref{correspondingauthor}}
\cortext[correspondingauthor]{Corresponding author: 
  e-mail: \href{mailto:avesani@fbk.eu}{\url{avesani@fbk.eu}}}
%\ead{avesani@fbk.eu}

\address[fbk]{NILab, TeV, Fondazione Bruno Kessler, Trento, Italy}
\address[iit]{PAVIS, Istituto Italiano di Tecnologia, Geonva, Italy}
\address[cimec]{Center for Mind/Brain Sciences (CiMeC), University of Trento, Rovereto, Italy}
\address[gin]{GIN, IMN, CNRS, CEA, Université de Bordeaux, Bordeaux, France}
\address[nnaisense]{NNAISENSE, Lugano, Switzerland}
\address[apss]{Department of Neurosurgery, Azienda Provinciale per i Servizi Sanitari, ``Santa Chiara" Hospital, Trento, Italy}
%
%\received{}
%\finalform{}
%\accepted{}
%\availableonline{}
%\communicated{}
%
\begin{abstract}
\begin{center}
	\vspace{-.7cm}
	\centering
	\includegraphics[width=.835\textwidth]{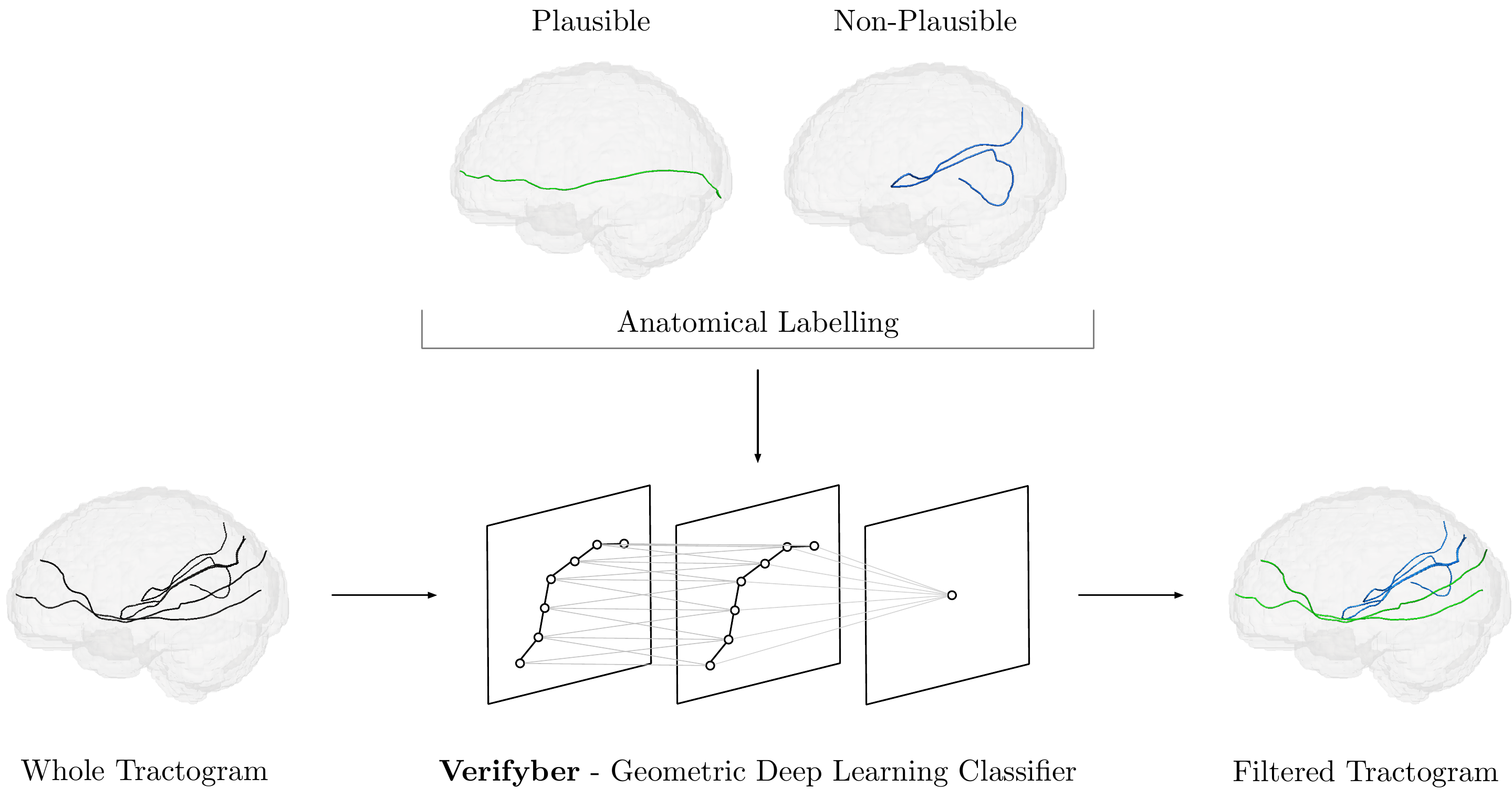}
	\vspace{-.1cm}
\end{center}
\small
A tractogram is a virtual representation of the brain white matter. It is composed of millions of virtual fibers, encoded as 3D polylines, which approximate the white matter axonal pathways. To date, tractograms are the most accurate white matter representation and thus are used for tasks like presurgical planning and investigations of neuroplasticity, brain disorders, or brain networks. However, it is a well-known issue that a large portion of tractogram fibers is not anatomically plausible and can be considered artifacts of the tracking procedure.  
With Verifyber, we tackle the problem of filtering out such non-plausible fibers using a novel fully-supervised learning approach. Differently from other approaches based on signal reconstruction and/or brain topology regularization, we guide our method with the existing anatomical knowledge of the white matter. Using tractograms annotated according to anatomical principles, we train our model, Verifyber, to classify fibers as either anatomically plausible or non-plausible.
The proposed Verifyber model is an original Geometric Deep Learning method that can deal with variable size fibers, while being invariant to fiber orientation. Our model considers each fiber as a graph of points, and by learning features of the edges between consecutive points via the proposed sequence Edge Convolution, it can capture the underlying anatomical properties. The output filtering results highly accurate and robust across an extensive set of experiments, and fast; with a 12GB GPU, filtering a tractogram of 1M fibers requires less than a minute. Verifyber implementation and trained models are available at \url{https://github.com/FBK-NILab/verifyber}.
\end{abstract}

\end{frontmatter}

%\linenumbers


\section{Introduction}%
\label{sec:introduction}

\outline{Preamble/synopsis with purpose of the work/paper: keys anatomy/gdl\\}
The purpose of this work is to leverage the brain anatomical knowledge to design a method for tractogram filtering based on the notion of non-plausible pathways of the white matter. This challenge is approached with a geometric deep learning model to better capture and learn the structural properties of the brain fibers and to provide fast tractogram filtering at run time.

\outline{Tractography impact: bundle and connectome (add clinical)\\}
A tractogram provides an explicit representation of the brain connectivity structure in the white matter~\citep{basser_vivo_2000}. It is composed of a collection of fibers, usually of the order of $10^{5-6}$, which encode the main axonal pathways. Each fiber, sometimes also called streamline, is represented as a 3D polyline by a sequence of points of variable length. Tractograms play a key role in both neuroanatomical studies~\citep{henderson_tractography_2020, maffei_topography_2018, jeurissen_diffusion_2017, hau_revisiting_2017, de_benedictis_new_2016} and brain network studies~\citep{zhang_quantitative_2022, yeh_mapping_2020}. The characterization of the structural brain connectivity aims to identify the bundles of fibers with a specific functional purpose. A neuroanatomical bundle is obtained by segmenting the relevant portion of fibers from the tractogram. The detection of main bundles represents an important step in the process of neurosurgical planning~\citep{yang_diffusion_2021, henderson_tractography_2020, essayed_white_2017}. Also brain network studies take as input a tractogram to compute the connectome. Given a parcellation of the brain cortex, the connectome is obtained by computing the adjacency matrix containing in each cell an estimate of the connectivity between two parcels based on the fibers connecting them. Both bundle segmentation and connectome computation are very sensitive to the accuracy of the tractogram ~\citep{buchanan_testretest_2014,rheault_tractostorm_2020}. For this reason the quality assurance of the fiber pathways is being an open challenge for the scientific community~\citep{jorgens_challenges_2021, zhang_quantitative_2022, rheault_common_2020, jeurissen_diffusion_2017, maier-hein_challenge_2017, thomas_anatomical_2014}.

\outline{Issue of Tractography: artifacts, noise, data derivative\\}
Tractograms are data derivatives. This type of data is the outcome of a complex pipeline of data processing. After a step of diffusion MRI preprocessing~\citep{glasser_minimal_2013, fischl_freesurfer_2012, jenkinson_fsl_2012, tournier_mrtrix3_2019} and a step of diffusivity model reconstruction~\citep{pierpaoli_diffusion_1996, tournier_robust_2007, descoteaux_high_2015}, a further step of tracking~\citep{mori_three-dimensional_1999, basser_vivo_2000, jeurissen_diffusion_2017} is in charge of the computation of the fiber pathways. The process of tracking requires the definition of many parameters like the policy of seeding, the strategy of stepping, the stop criterion, the constraints on curvature and length of fibers. Slightly different choices may produce quite different tractograms~\citep{thomas_anatomical_2014}, and the evaluation of their accuracy is not straightforward~\citep{neher_strengths_2015}. 

\outline{Survey of tractography limitations\\}
In the last years the issues of tractography reliability and reproducibility have been approached with several data analysis contests: FiberCup on 2011~\citep{fillard_quantitative_2011}, Fiberfox on 2015~\citep{maier-hein_challenge_2017}, Traced on 2017 ~\citep{nath_tractography_2020} and Votem on 2018~\citep{schilling_limits_2019}. The design of these contests is quite similar and it is based on the use of a phantom to define the ground truth in advance~\citep{cote_tractometer_2013}. The evaluation is carried out by measuring the mismatch between the synthetic model and the tractograms computed using the state of the art methods. These initiatives achieved a general agreement on the main limitations of the tractography techniques. 

A recurrent weakness of tracking algorithms is the generation of false positive fibers ~\citep{maier-hein_challenge_2017}. Indeed, the evolution of the diffusivity model reconstruction from DTI~\citep{pierpaoli_diffusion_1996} to HARDI~\citep{descoteaux_regularized_2007} improved the sensitivity of tracking (missing less existing pathways), but at the cost of increasing the number of fibers that are not anatomically plausible~\citep{thomas_anatomical_2014}. Also the common practice of overtracking  contributes to the decreasing of tracking specificity. One of the reasons for this issue is that fiber density is not 
coherent with the physiological distribution of axons~\citep{raffelt_apparent_2012, yeh_correction_2016}. While for neuroanatomical studies it is crucial to have high sensitivity tractograms, brain connectivity studies might be meaningfully affected by low specificity tractograms where false positive fibers may impact twice with respect to false negatives \citep{zalesky_connectome_2016}. Finding a balance between sensitivity and specificity in fiber tracking algorithms is still an open problem.

\paragraph{State of the art of tractogram filtering}\mbox{}\\
\outline{Solution strategies: ex-ante vs ex-post\\}
The recent literature provides many contributions to improve the accuracy of the tractograms. The proposed methods can be divided into two main groups according to their strategy. The first strategy could be referred to as {\em ex-ante} and includes all the attempts to revise the tracking algorithms with the goal of reducing the generation of non-plausible fibers. The other strategy is approaching the problem {\em ex-post}, by filtering out artifactual fibers once the tractogram is computed.

\outline{Solution ex-ante: tracking driven by anatomy priors, bundle-specific\\}
According to the {\em ex-ante} strategy, a quantitative study~\citep{bastiani_human_2012} was carried out  to investigate how the choice of hyperparameters might impact the results of tracking algorithms. The tuning of heuristics, especially in stochastic methods for probabilistic tracking, introduces significant source of variability and a critical dependency from the choice of parameters’s values. The general trend to improve the tracking is to make sure that such heuristics are anatomically informed. For example, seeding and stopping criteria have been revised to be driven by the gray matter and white matter interface~\citep{smith_anatomically-constrained_2012, girard_towards_2014, lemkaddem_global_2014} or the tracking has been constrained using topographic regularity \citep{aydogan_tracking_2018}. The challenge becomes how to inject anatomical priors in the tracking algorithms. It turned out to be easier and effective to elicit anatomical constraint in the case of bundle specific tracking~\citep{yendiki_automated_2011, thomas_anatomical_2014, chamberland_active_2017, rheault_bundle-specific_2019}. In these restricted cases the fiber pathways are driven by volumetric ROI defined according to the anatomical knowledge of a specific bundle. However, {\em ex-ante} strategies remains non-successful for the whole brain tractography~\citep{schilling_brain_2020}. 

\outline{Solution ex-post: task of filtering (signal vs tractography only)\\}
The task of tractogram filtering, which approaches the problem with an {\em ex-post} strategy, adopts a global view and considers as input the whole brain tractogram. We may consider tractogram filtering as a complementary step to be combined with a better tuning of the tracking algorithms, since {\em ex-ante} and {\em ex-post} strategies are not in contrast or mutually exclusive. We may distinguish two types of filtering solutions: signal-based and tractography-based. The former solutions formulate the filtering task as an inverse problem of signal reconstruction, the latter solutions adopt filtering criteria based only on the tractography data. Both of them carry out an unsupervised strategy.  

\outline{Signal-based SIFT, LIFE, COMMIT\\}
In the signal-based solutions the plausibility of fibers is estimated by computing how much their pathways are explained by the diffusion signal. The most common methods cast the filtering task as a global regularization problem by assigning a weight to each fiber. In 
SIFT~\citep{smith_sift:_2013} and SIFT2~\citep{smith_effects_2015}, the weights are a proxy of the fibers density. In LiFE~\citep{pestilli_evaluation_2014} the weights capture how much the fiber pathways are related to the diffusion signal. COMMIT~\citep{daducci_commit_2015} extends the estimate of weights by including microstructure information. In all these methods the thresholding of weights to discriminate between plausible and non-plausible fibers is managed with heuristics. Nevertheless, as remarked by \citet{smith_quantitative_2020, frigo_diffusion_2020, rheault_bundle-specific_2019}, the filtering operated with a regularization approach might remove fibers whose pathway is anatomically plausible.  

\outline{Tractography-based Unsupervised Grouping-informed [Xia, Wang]\\}
The alternative approaches are based only on tractography. Their basic assumption is that the topographic regularity of tractogram structures across individuals might be a good proxy of anatomical plausibility. Different unsupervised methods, by leveraging the groupwise consistency of fiber bundles, have been proposed to detect outlier pathways~\citep{odonnell_automatic_2007, wang_modeling_2018, xia_groupwise_2020}. They differ in the definition of the proximity metrics for the computation of topographic regularity. The filtering in these cases is subject to the population bias due to the lack of general anatomically-informed priors. For this reason, the groupwise analysis is usually limited to the anatomy of well-known fiber bundles. While a large population of tractograms provides a more robust estimate of structure regularity, the consistency constraints tend to eliminate inter-individual differences. To contrast the smoothing effect due to population averaging, other unsupervised methods refer to the fiber density map~\citep{yeh_automatic_2019} or to the local fiber agreement~\citep{chandio_fiberneat_2022} as a proxy of anatomical plausibility. The focus in these methods is to exploit the intra-individual information.

\outline{Mixed-based Signal+Bundles filtering (regularization) [nie, Yeh, schiavi, ocampo, neher, aydogan]\\}
The two distinct approaches, signal-based and tractography-based, have been combined to design mixed solutions~\citep{aydogan_track_2015, neher_anchor-constrained_2018, nie_topographic_2019, schiavi_new_2020, ocampo-pineda_hierarchical_2021}. The basic intuition is that where a priori knowledge of neuroanatomical bundles is not available, the fibers are regularized by a signal-based filtering, while along the pathways of known bundles, the fibers are filtered out when they do not meet the expected topographic regularity.  

\outline{Deep learning unsupervised embedding (bundle driven?) [legarreta]\\}
The most recent trend in unsupervised methods for tractogram filtering is the investigation of deep learning techniques. FINTA~\citep{legarreta_filtering_2021} proposes a convolutional neural network to learn an embedded representation of the fibers. After the learning procedure, the fibers are projected into a new latent space where the computation of nearest neighbors might easily detect the similarity as proximity. However, the learning of the embedding is not driven by neuroanatomical knowledge and the subsequent filtering of fiber is not guaranteed to properly capture the notion of anatomical plausibility. %\comment{Indeed they demonstrate to filter properly when considering specific bundles}

\outline{Recap: unsupervised only and trend of anatomy\\}
As a general remark of the state of the art, we may notice that all methods following an {\em ex-post} strategy to tractogram filtering are adopting an unsupervised approach, and the design of a loss function suitable to capture the notion of anatomical plausibility remains an open challenge. For this reason, the common trend is to integrate additional neuroanatomical constraints, both in {\em ex-post}~\citep{neher_anchor-constrained_2018, nie_topographic_2019, schiavi_new_2020, ocampo-pineda_hierarchical_2021} and {\em ex-ante}~\citep{rheault_bundle-specific_2019} strategies.     

\outline{Proposal Verifyber: ex-post, trk-based, supervised, gdl, anatomy driven\\}
\paragraph{Our contributions}\mbox{}\\
In this work we propose Verifyber, a novel tractography-based method to perform {\em ex-post} filtering of non-plausible fibers from a tractogram. The task of tractogram filtering is shaped as a supervised learning problem where a binary classifier takes in input a fiber and provides as output either the category anatomically plausible or anatomically non-plausible. We present an original learning model based on geometric deep learning (GDL)~\citep{masci_geometric_2016, bronstein_geometric_2017}, which better fits the learning on 3D data without forcing Euclidean vector representations. The notion of anatomical plausibility is derived from fiber examples, labelled either as anatomically plausible or anatomically non-plausible.  

\outline{Anatomy ground truth: inclusion vs exclusion\\}
We envision the task of elicitation of brain knowledge as a binary labelling of fibers. Despite the evolutionary nature of the knowledge of the human brain, we may conceive two main labelling policies: {\em inclusive} and {\em exclusive}. The {\em inclusive} policy leans to be more conservative and aims to prevent false positives. According to this prior only fibers following the pathways of well known bundles are labelled as anatomically plausible, non-plausible otherwise. Conversely, the {\em exclusive} policy is more sensitive to the false negative, in this case only fibers with clear artifactual pathways are labelled anatomically non-plausible. It is out of the scope of this work to establish which policy might be more effective and appropriate. Our goal is to investigate whether the proposed method is equally robust for the two policies. 

\outline{Inclusive Anatomy: known bundles (evolutionary/biased) [Zhang]\\}
Our empirical analysis is considering datasets labelled with both {\em inclusive} and {\em exclusive} policies. As a reference example of {\em inclusive} policy we point to an anatomically curated white matter atlas~\citep{zhang_anatomically_2018}. This atlas provides a whole brain tractogram averaged over 100 individuals. A team of experts manually curated the annotation of 74 bundles. For our purpose we considered anatomically plausible all the fibers of those bundles, non-plausible otherwise. 

\outline{Exclusive Anatomy: heuristic rules (spurious only) [Extractor]\\}
In the literature the {\em exclusive} policy is less common. As an instance of this kind we consider Extractor~\citep{petit_structural_2022}. In such a work, the notion of anatomical non-plausibility is defined by a set of heuristic rules based on the current knowledge of the human white matter. Well-known artifactual pathways based on geometric properties or brain locations are labelled as anatomically non-plausible, usually half portion of the whole tractogram. This declarative knowledge can be applied to any tractogram enabling the annotation of training and test sets for learning purposes. 

\outline{Learning issue: fixed point representation\\}
Regardless of the source of labelling, the challenge of supervised learning is to train a binary classifier based only on a digital representation of fibers. The choice of an appropriate representation of a fiber, suitable for the learning process, is a crucial step. Usually a fiber is encoded as an ordered sequence of a variable number of 3D points. Previous works on supervised learning for tractography had to deal with the constraint of learning algorithms that require a fixed length embedding. The most common solutions are the computation of an Euclidean embedding such as dissimilarity representation~\citep{olivetti_supervised_2011, berto_classifyber_2021}. Unfortunately these fiber embeddings are lossy. %\comment{add other example of embeddings (berto, yeh, odonnel) and mention workaround of custom distance metrics (gori-varifolds, quickbundle, recobundle), mention that other deep learning approaches uses voxel maps (tractseg, neuro4neuro) or they assume streamlines to be 2d maps and then uses standard CNN (fibernet, fibernet2, finta, deepwma)}

\outline{Geometric deep learning fits streamlines representation\\}
To overcome such limitations and to preserve the full geometrical information encoded in the fiber pathways, we propose to investigate the use of Geometric Deep Learning models like PointNet~\citep{qi_pointnet_2017} and Dynamic Graph CNN~\citep{wang_dynamic_2019}, which by construction can deal with variable size inputs like point clouds and graphs. GDL architectures are based on layers of permutation invariant/equivariant operators whose combination allows a model to perform convolution in a non-grid (and non-Euclidean) representation. Our working hypothesis is that GDL might be more accurate in capturing the geometrical properties of pathways associated with the notion of anatomical plausibility.  

\outline{Contribution: novel model\\}
Verifyber is designed as a novel end-to-end trainable GDL model to deal with traditional encoding of fibers as ordered sequences of a variable number of 3D points. Our proposal extends the Edge Convolution (EC) layer by \citet{wang_dynamic_2019} to take into account the information encoded by the edges between two subsequent 3D points in a fiber. Then, the architecture is composed by a global pooling layer that compresses each fiber to a single descriptor and a classification head to discriminate between the two categories, either anatomically plausible or non-plausible. Differently from EC, our model is sequence sensitive, i.e., not permutation invariant, while it remains invariant to the orientation of the input fibers. %\comment{mention contribution sEC as in the old drafting}

\outline{Outline experiment and results\\}
We provide the results of a broad set of experiments aimed at proving the properties of the proposed model and assessing the efficacy in discriminating anatomically plausible and non-plausible fibers. We show that Verifyber outperforms in accuracy competing deep learning methods such as bidirectional LSTM~\citep{graves_framewise_2005, huang_bidirectional_2015}, PointNet~\citep{qi_pointnet_2017}, and Dynamic Graph CNN~\citep{wang_dynamic_2019}. These results are robust with respect to different types of tractography and equally effective on {\em inclusive} and {\em exclusive} policies to elicit the notion of neuroanatomical plausibility. An additional comparison is aimed to show the different behaviour of supervised and unsupervised filtering approaches, these lasts represented by FINTA \citep{legarreta_filtering_2021}. We also investigate how a trained model behaves across different sources of tractograms, when the computation of tracking is not necessarily homogeneous. We show some preliminary results of this kind of analysis on a clinical dataset. An additional simulation study allows the evaluation of the behaviour of the method when the labeling of fiber is dynamically evolving over time.

%%% Local Variables: 
%%% mode: latex
%%% TeX-master: "tractogram_filtering_main"
%%% End:
\section{Method}%
\label{sec:method}

\outline{premise\\}
In this section, we describe our method, Verifyber. For the sake of comprehension, we also summarize the Edge Convolution (EC) layer ~\citep{wang_dynamic_2019}, which is a building block of our model on top of which we built the proposed {\em sequence EC}. 

\subsection{Edge Convolution Layer}
\label{ssec:t_filt_ec}
\begin{figure}[th]
    \centering
        \begin{subfigure}[t]{0.49\linewidth}
        \centering
        \includegraphics[width=0.73\linewidth]{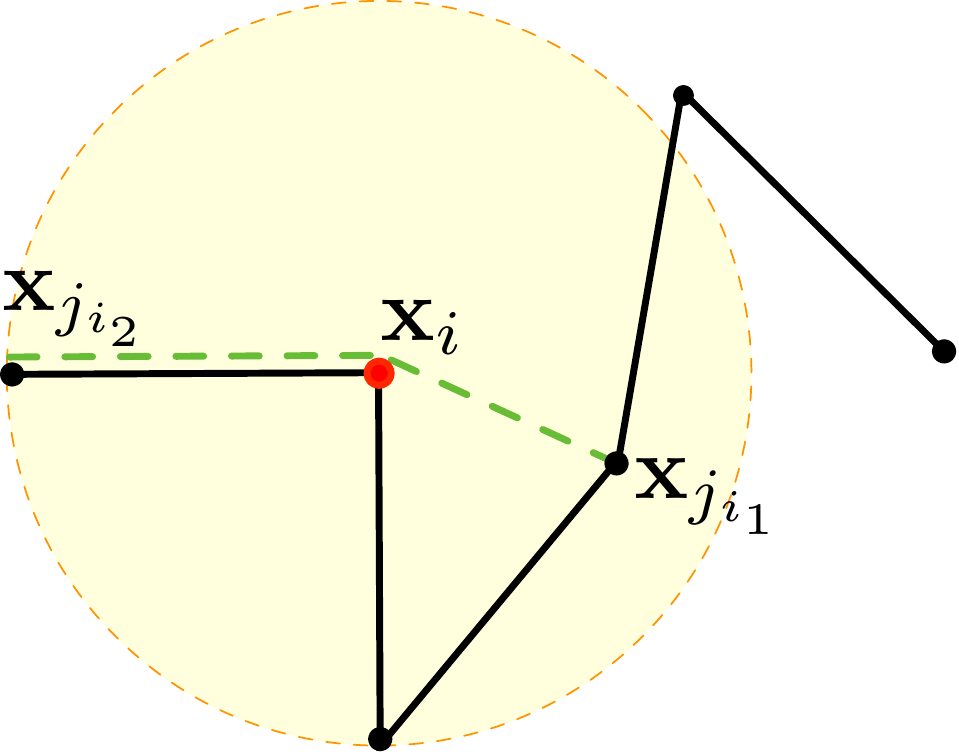}
        \caption{}
        \label{fig:knn_a}
    \end{subfigure}
    \begin{subfigure}[t]{0.49\linewidth}
        \centering
        \includegraphics[width=0.7\linewidth]{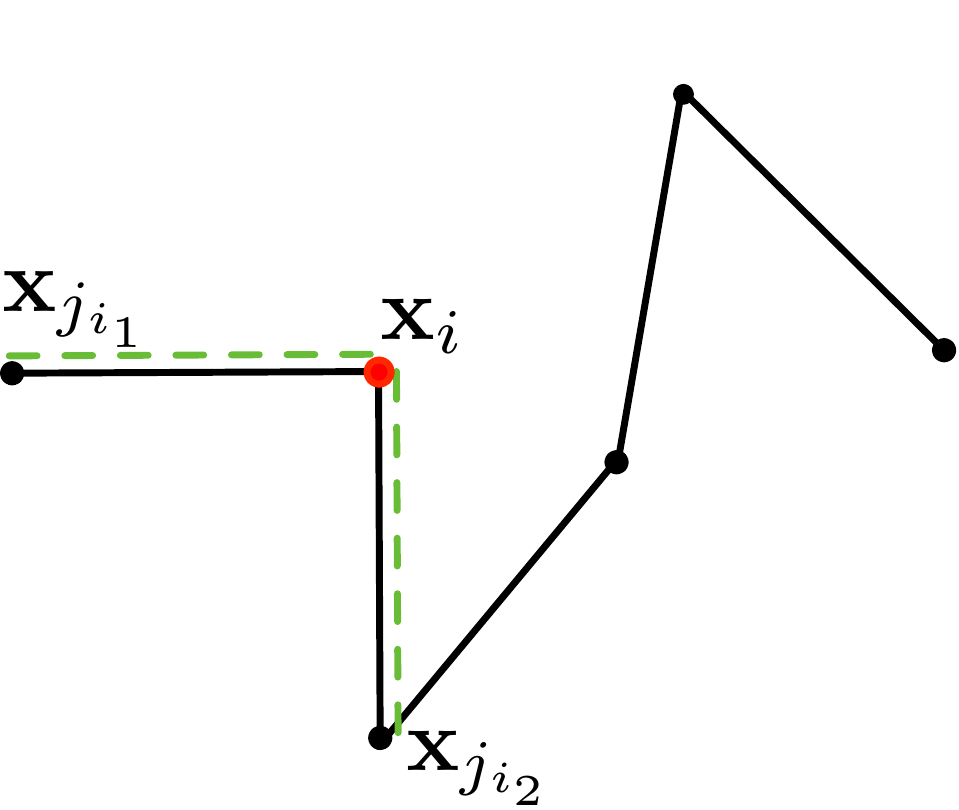}
        \caption{}
        \label{fig:knn_b}
    \end{subfigure}
    \caption[Euclidean $k$-nn vs. graph $k$-nn]{Comparison between Euclidean $k$-nn (a), and graph $k$-nn on the streamline (b).}
\end{figure}

Considering a point cloud $\mathcal{X} = \{ \textbf{x}_1, \textbf{x}_2, \dots, \textbf{x}_n\}$, $\textbf{x}_i \in \mathbb{R}^3$, an Edge Convolution (EC) layer first induces a graph structure for $\mathcal{X}$ by retrieving for each point $\textbf{x}_i$ the set of $k$ nearest neighbors, $\mathrm{knn}(\textbf{x}_i) = \{\textbf{x}_{j_{i_1}}, \dots, \textbf{x}_{j_{i_k}}\}$, using the Euclidean distance as metrics (see Figure \ref{fig:knn_a}). The result is a $k$-nn graph composed of $\mathcal{V}$ nodes and $\mathcal{E}$ edges:
\begin{equation}
\label{eq:graph_ec}
\mathcal{G(V,E)},\; \mathcal{V}=\mathcal{X},\; e_{ij} \in \mathcal{E} \colon \textbf{x}_i \to \textbf{x}_j \iff \textbf{x}_j \in \text{knn}(\textbf{x}_i).
\end{equation}
Then, each point representation, $\textbf{x}_i$, is enriched with the representation of each of its neighbors, $\textbf{x}_{j_i}$, to obtain edge features, $\textbf{e}_{ij}$, which are learnt through a neural network $h_{\boldsymbol{\mathrm{\Theta}}}$. Specifically,   
\begin{equation}
\textbf{e}_{ij}=h_{\boldsymbol{\mathrm{\Theta}}}(\textbf{x}_i \oplus (\textbf{x}_j - \textbf{x}_i)),
\end{equation}
where $\oplus$ denotes the concatenation operator. Finally, a new representation of a point, $\textbf{x}_i^\prime$, is obtained by aggregating all the learned edge features with a pooling operator, i.e., $\textbf{x}_i^{\prime}={\mathrm{pool}}(\textbf{e}_{ij})$, $j \colon (i,j) \in \mathcal{E}$, where $\mathrm{pool}$ is either $\mathrm{max}$ or $\mathrm{mean}$. 

\begin{figure*}[th]
    \centering
    \includegraphics[width=.835\linewidth]{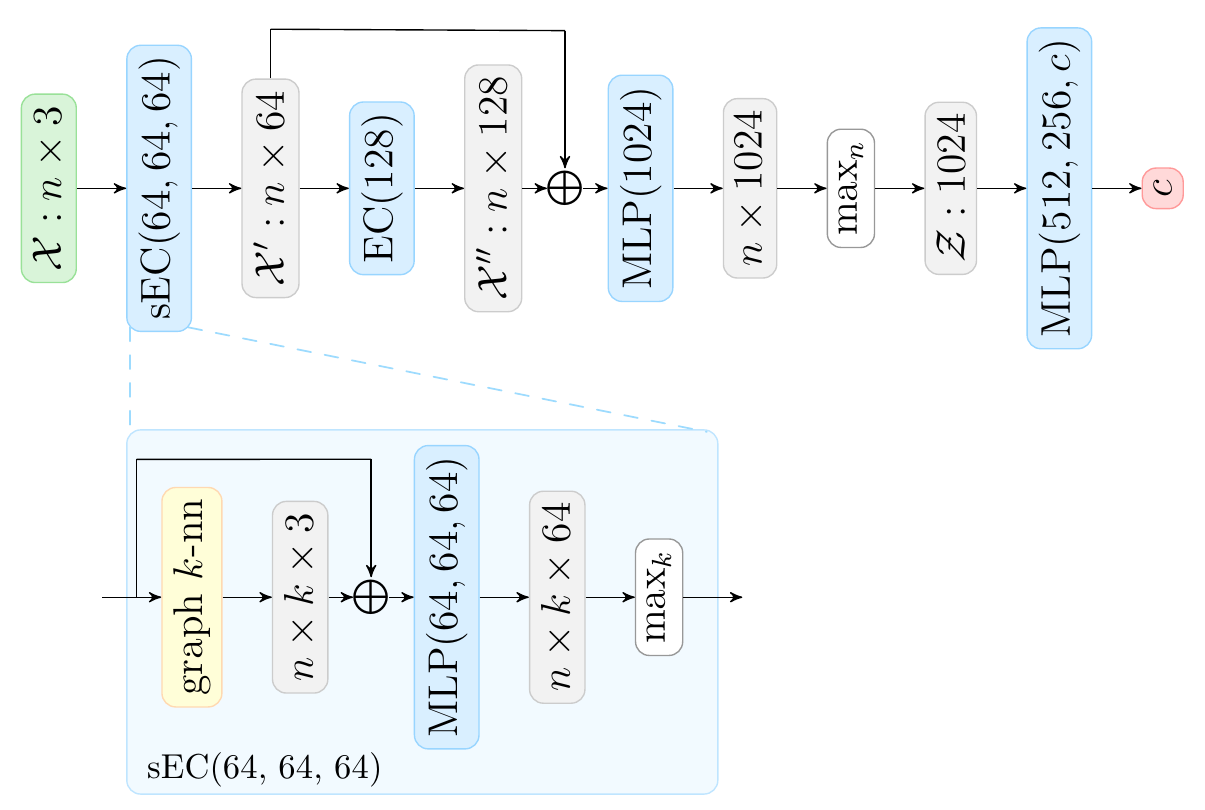}
    \caption{Block diagram of the Verifyber architecture. Green, gray, and red blocks represent input, intermediate, and output tensors, respectively. Parametric layers are colored in blue, while non-parametric layers in white. In yellow we highlight the graph $k$-nn that allows the model to be sequence sensitive. }
    \label{fig:dec_arch}
\end{figure*}

\subsection{Sequence Edge Convolution Layer}
A remarkable property of the EC layer is the invariance to the permutation of the points in the input point cloud. Indeed, the layer contains only operators invariant to the points ordering, e.g., FC layers, max / mean pooling, Euclidean $k$-nn. Although this property is fundamental in the point cloud domain, it becomes undesired if the input is a sequence as in our case. To solve this issue, we propose a simple but well-motivated modification: we substitute the Euclidean $k$-nn, which was inducing a graph structure based on euclidean distance of the points, with a graph-based $k$-nn (see Figure \ref{fig:knn_b}) that instead computes neighbors along the existing input graph. Considering the streamline structure, the graph-based $k$-nn preserves the input graph to be a bidirectional sequence of points where each non-terminal point, $\textbf{x}_{i \neq 0,n}$, has two neighbors: the previous and the next point in the sequence, while the terminal points, $\textbf{x}_0, \textbf{x}_n$, have just one neighbor:   
\begin{equation}
\mathcal{G(V,E^\prime)},\, e^\prime_{ij} \in \mathcal{E^\prime} \colon \textbf{x}_i \to \textbf{x}_j, \iff j=i+1 \vee j=i-1.
\end{equation}
By using this graph structure an EC layer loses the invariance to the input permutations, while maintaining the invariance with respect to the input flipping (a crucial property when dealing with streamlines), thanks to the bidirectionality of the edges. For this reason, we define this modified EC layer as \emph{sequence} EC (sEC) layer.

\subsection{Verifyber Model}
The Verifyber (VF) model is characterized by the stacking one sEC layer with one or more EC layers. Figure~\ref{fig:dec_arch} shows the block diagram of the architecture of the Veryfiber model we used for our experiments in Section~\ref{sec:experiments}. One sEC layer and one EC layer are stacked to produce new representations $\mathcal{X}^{\prime}$ and $\mathcal{X}^{\prime\prime}$ with 64 and 128 features respectively. The stacking of these two layers guarantees the model to be both sequence sensitive and \emph{dynamic}: as shown in~\citep{wang_dynamic_2019}, the computation of knn in latent space allows a dynamic adjustment of the local neighborhood of points guided by the optimization of the task at hand. Then, $\mathcal{X}^{\prime}$ and $\mathcal{X}^{\prime\prime}$ are concatenated, encoded to 1024 features with a learning layer $g_{\boldsymbol{\mathrm{\Phi}}}$, and pooled to obtain a single descriptor of the whole point cloud, 
\begin{equation}
\textbf{z}=\mathrm{pool}(g_{\boldsymbol{\mathrm{\Phi}}}(\mathcal{X}^{\prime} \oplus \mathcal{X}^{\prime\prime})).
\end{equation}
Finally, the 1024-dimensional feature vector $\textbf{z}$ is classified using a fully connected (FC) network composed of three layers, which decreases the number of features to 512, 256, and $c$ (number of classes), respectively.

%%% Local Variables: 
%%% mode: latex
%%% TeX-master: "tractogram_filtering_main"
%%% End:
\section{Related Works}%
\label{sec:related_works}

\outline{Learning from streamlines: the issue of data representation\\}
The proposed Verifyber model is the result of a step-by-step investigation aiming to solve a well-known problem in tractography analysis: finding a data representation compliant with computational requirements. Performing automated analysis of a tractogram requires a method able to deal with the structure of streamlines. Such a structure presents some characteristics which differ from the typical neuroimaging data representations like images and volumes: a streamline is a sequence of points with variable length and no orientation. 
%Hence, the space of streamlines is non-Euclidean

\outline{heuristics -> resampling. Embedding handcrafted, sometimes domain independent\\}
\paragraph{Streamline embedding and traditional models} Streamlines' properties prevent traditional machine learning methods from being directly applied to them. One common requirement of machine learning methods is to have fixed length vectors as input, and thus existing works resorted to different preprocessing solutions to match such a requirement. A widely adopted heuristics consists in resampling the streamlines to a fixed number of points~\citep{garyfallidis_quickbundles_2012, garyfallidis_recognition_2018, gupta_fibernet_2017, odonnell_automatic_2007, legarreta_filtering_2021}. However, for traditional methods like support vector machines or linear classifiers, the resulting 3-dimensional vectors with fixed length, (\# points, $(x,y,z)$), are not suitable, and they need to employ a more advanced embedding technique to project streamlines in a new convenient space. An example are embeddings based on dissimilarity representation like ~\citet{olivetti_supervised_2011} and more recently ~\citet{berto_classifyber_2021} that also considers handcrafted features based on the white matter anatomy, or using non-linear dimensionality reduction techniques, e.g., t-SNE~\citep{van_der_maaten_visualizing_2008}, like in ~\citet{chandio_fiberneat_2022}. These embeddings enable the training of traditional classifiers at the cost of losing some geometrical/structural information of the streamlines, e.g., the presence of a loop.  

\outline{Unsupervised with grid-based conv -> no task related + controversial conv (no flip inv.)\\}
\paragraph{Convolutional Neural Networks for streamlines} The limitation of lossy embeddings might be bypassed using deep learning techniques that are able to learn embeddings based on the target task. Given the breakthrough of CNNs in computer vision, there have been recent attempts to apply them also to streamlines. FINTA~\citep{legarreta_filtering_2021} proposes an unsupervised approach, where the embedding is learned by means of a convolutional autoencoder, and then it is used as input for the downstream task, e.g., tractogram filtering. Even though the learned embedding might preserve the structural information of streamlines, the lack of a task-specific supervision does not guarantee its optimality for the downstream task. FiberNet~\citep{gupta_fibernet_2017} and FiberNet 2.0~\citep{gupta_fibernet_2018}, Deep CNN (DCNN) tract classification~\citep{xu_objective_2019, lee_novel_2020} and Deep White Matter Analisys (deepWMA)~\citep{zhang_deep_2020}, instead, train standard CNN models like AlexNet~\citep{krizhevsky_imagenet_2012} and ResNet~\citep{he_deep_2016} directly on streamlines using bundle supervision. In this way they learn an embedding specific for the bundle segmentation task. However, we notice a controversial use of convolutional filters in FiberNet and DCNN, as they treat streamlines of size (\# points, $(x,y,z)$) like images of size (height, width), i.e., width = 3. More correctly, as operated in FINTA and DeepWMA, $(x,y,z)$ should be considered channels like $(r,g,b)$ in images so that different filters are learned for each channel. Finally, we also notice a general drawback when using standard CNNs to perform learning on the streamline structure. Indeed, CNNs are translation invariant (or equivariant), a property that is crucial for the image domain. However, streamlines, unlike images, can be drastically affected by translating points, and thus the translation invariance/equivariance of CNNs is not a desired property. Streamlines, instead, being unoriented, require flip invariance, but unfortunately this is not a property of standard CNNs despite it is neglected by the approaches mentioned above.
%\todo TRAFIC~\citep{}
%liu deepbundle
% zhong autoencoded
% patil siamese

\outline{supervised + geometric conv -> task related + geometric learning (flip inv.)\\}
\paragraph{Recurrent Neural Networks and Geometric Deep Learning} Based on these premises we investigate supervised deep learning approaches different from standard CNNs. We seek for neural network architectures more suitable for the streamline structure. Guided by the sequentiality of the streamline structure, we start our investigation from setting a baseline using a Recurrent Neural Network (RNN) model. Then, with the aim to have a flip invariant model able that deals with size-varying input, we explore some methods from the family of Geometric Deep Learning (GDL)~\citep{masci_geometric_2016, bronstein_geometric_2017}. GDL comprises all the methods that extend convolution principles to non-Euclidean data, e.g., non grid-based data like point clouds and graphs. To deal with such data, GDL models exploit modules and operators that permutation invariant instead of translation invariant and that can be applied to batches of size-varying samples.

\outline{Reference DL method for sequence: bLSTM\\}
\paragraph{Bidirectional LSTM~\citep{graves_framewise_2005}} In the literature of RNN methods, especially in the field of Natural Language Processing where data has a sequential structure, a large number of methods is based on Long Short Term Memory (LSTM)~\citep{hochreiter_long_1997}. Among all, we individuate the bidirectional LSTM (bLSTM)~\citep{graves_framewise_2005, huang_bidirectional_2015} as a reference deep learning method to analyze streamlines. bLSTM is characterized by two LSTM layers, each of them fed with a different orientation of the input. It learns a shared embedding of both orientations by combining the two LSTM outputs with an aggregator operator (concatenation) and then forwards it to a FC network, which performs classification.
\outline{Drawback bLSTM: no orientation + fixed length\\}However, there are two main limitations of the bLSTM method when applied to streamlines. First, it requires a fixed-length vector as input to its LSTM layers, and second it is not invariant to the input flipping despite the bidirectional architecture. Indeed, the two LSTM layers learn two different set of parameters, which may produce different hidden states if fed with the same sequence. Nevertheless, the use of the two directions is still beneficial for the network because it improves the learning of local context information. Eventually, bidirectionality combined with an augmented training where streamlines are given in both orientations might mitigate the lack of flip invariant layers.

\outline{Streamlines as point cloud\\}
The limitations of bLSTM are not present in GDL methods, which by construction are flip invariant (special case of permutation invariance) and can be fed with size-varying point clouds or graphs. Hence, if we model a streamline as a point cloud we are neglecting its sequential structure (since point clouds do not assume any ordering of their vertices), but we maintain both the streamline spatial information and its invariance to the flipping of orientation.
%We may approximate a streamline to be a point cloud by neglecting its sequential structure as point clouds do not assume any order in the points. Despite such a representation loses sequentiality it maintains both the streamline spatial information and its invariance to the flipping of orientation. 

\outline{Point Net model\\}
\paragraph{PointNet~\citep{qi_pointnet_2017}} In our experiments, we investigate the pioneer and most adopted model for point cloud, namely PointNet (PN)~\citep{qi_pointnet_2017}. PN is characterized by a simple architecture composed only of FC layers and pooling layers that are by construction permutation invariant. In particular, for the task of classification PN presents a series of FC layers as encoder, a max pooling layer that generates a single global feature vector of the input point cloud, and another series of FC layers performing the output classification.
\outline{Drawback PN: no point relations\\}However, learning on streamlines using PN could be limited due to the non-consideration of point relations. Indeed, PN is only able to consider a global relation among all the points by performing the max pooling in latent space. For this reason we decided to investigate also a GDL model that consider points relation as encoded by graph structures, namely Dynamic Graph CNN (DGCNN)~\citep{wang_dynamic_2019}.  

\outline{DGCNN to capture point relations. Drawback DGCNN: permutation invariant\\}
\paragraph{Dynamic Graph CNN~\citep{wang_dynamic_2019}} The DGCNN model is, according to \citeauthor{wang_dynamic_2019} a generalization of PN. Instead of considering a single global (all to all) relation, DGCNN considers multiple local neighborhood relations, like in a $k$-nn graph structure, computed at different depths of the network, i.e., in different latent spaces. The model is based on Edge Convolution layers (explained in Section \ref{ssec:t_filt_ec}) which have deeply inspired our Verifyber. However, since DGCNN makes only use of EC layers (plus the classification decoder), it is a permutation invariant model as well as PN. These models cannot distinguish two streamlines whose points are randomly shuffled, and this is an undesired behaviour for the tractogram filtering task.          

\outline{Verifyber overcome several current limitations. Verifyber properties: capture relation, order sensitive, no fixed length\\}
\paragraph{Verifyber} Our contribution sEC allows the proposed Verifyber model to overcome the permutation invariance limitation while remaining orientation invariant. Also, VF inherits the other good properties of PN and DGCNN, resulting able to work with size-varying input and to consider point relations.

\section{Material}%
\label{sec:material}

\outline{describe section structure: subdivision in dataset to test the method performances, and dataset to check neuroscience impact of the method.\\}

In this section, we present the datasets used for the empirical analyses. A
summary is reported in Table~\ref{tab:datasets}. For each dataset, we report the data source and the type of labelling, either with inclusive or exclusive policies. The source data are mainly drawn from the Human Connectome Project (HCP)~\citep{van_essen_wu-minn_2013} repository.

\renewcommand{\thefootnote}{\fnsymbol{footnote}}
\renewcommand{\TPTnoteSettings}{\footnotesize}
\newcolumntype{L}{>{\centering\arraybackslash}p{1cm}}
\newcolumntype{W}{>{\centering\arraybackslash}p{1.6cm}}
\begin{table*}[ht]
    \centering
    \caption[Summary of Verifyber materials]{Summary of the adopted datasets. B: number of bundles. \#: number of tractograms. T: number of fibers in a tractogram.
    }
    \label{tab:datasets}
    \begin{adjustbox}{width=\textwidth}
    \tiny
    \begin{threeparttable}
%	\scalebox{2}
    \begin{tabular}{l | l | p{1.6cm} | c |c | c | p{.6cm} | p{1.6cm}}
        \toprule
        Name & Source & {\em p}/{\em np} label & B & \# & T & Track & DWI  \\
        \midrule[0.2pt]
%        \multirow{2}{*}
%        {\em HCP-EP} & HCP & Exclusive & - & 20 & 1M & CSD & ?\\
%        	 		 & & Petit et al.~\citep{petit_structural_2022} & & & & PF-ACT & ?\\
%        \midrule[0.2pt]
		{\em HCP-EP}\tnote{*} & HCP & Exclusive \newline \citet{petit_structural_2022} & - & 20 & 1M & CSD\newline  PF-ACT & 3T DWI, 1.25mm,\newline270g multi-b=(1,2,3)K\\
%		\midrule[0.2pt]
%        {\em BILGIN-EP} & BIL \& \newline GIN & Exclusive \newline \cite{petit_structural_2022} & - & 1 & 1.5M & ? & ?\\
        \midrule[0.2pt]
        {\em HCP-IZ}\tnote{$\dagger$} & HCP & Inclusive \newline \citet{zhang_anatomically_2018} & 74 & 1\tnote{\S} & 1M & HARDI\newline  UKF & 3T DWI, 1.25mm,\newline108g single-b=3K\\
        \midrule[0.2pt]
        {\em HCP-IW}\tnote{$\ddagger$} & HCP & Inclusive \newline \citet{wasserthal_tractseg_2018} & 72 & 23 & 10M & CSD\newline  iFOD2 & 3T DWI, 1.25mm,\newline270g multi-b=(1,2,3)K\\
        \midrule[0.2pt]
        {\em APSS-IS} & APSS & Inclusive \newline expert: S.S. & 4 & 5 & 100K & DTI\newline  EuDX & 1.5T DWI, 2.5mm,\newline60g single-b=1K\\
        \bottomrule
    \end{tabular}
    \begin{tablenotes}
    	\tiny
        \item[*] \url{https://doi.org/10.25663/brainlife.pub.13}
        \item[$\dagger$] \url{https://github.com/SlicerDMRI/ORG-Atlases}
        \item[$\ddagger$] \url{https://zenodo.org/record/1477956\#.Ya67UyzMKL8}
        \item[\S] averaged from 100 subjects.
    \end{tablenotes}
	\end{threeparttable}
    \end{adjustbox}
    \vspace{-.4cm}
\end{table*}

%\paragraph{HCP diffusion} 
%\outline{Shared base of other datasets. Brief description of spec.\\}
%The Human Connectome Project (HCP)~\citep{van_essen_wu-minn_2013} diffusion dataset is the base of most of the datasets we present in the next paragraphs. It is a public and widely adopted dataset where the acquisition and pre-processing pipeline for diffusion MRI is carefully validated~\citep{milchenko_obscuring_2013, sotiropoulos_effects_2013, glasser_minimal_2013}. The dataset comprises healthy subjects (all genders) aged between 24 and 35. Each subject has both the structural T1 image and the 3T Diffusion Weighted Image (DWI). The DWI has resolution 1.25mm with 270 gradients multi-shell~\citep{andersson_integrated_2016}, and is corrected with eddy currents~\citep{andersson_non-parametric_2015}.  

\paragraph{Dataset HCP-EP}
\outline{Technical: size, diffusion model, tracking, co-registration\\}
This dataset is composed of 20 individuals randomly selected from HCP repository. The processing pipeline carried out the estimation of the diffusivity model using the Constrained Spherical Deconvolution (CSD)~\citep{tournier_robust_2007}, and the fiber tracking using the algorithm for Particle Filtering Anatomically Constrained Tractography (PF-ACT)~\citep{girard_towards_2014}. More in detail, the tracking generated around $\sim$1M streamlines for each tractogram by seeding 16 points for each voxel with step size 0.5mm. Tractograms were normalized to the same space via non-linear co-registration to the MNI152 standard brain~\citep{fonov_unbiased_2011}. For computational purposes, all the streamlines have been compressed to the most significant points~\citep{presseau_new_2015}.

\outline{Extractor is an external labeling based on anatomy. Rule-based labeling. Test the model performance intra dataset.\\}
For this dataset, the labelling was performed using Extractor~\citep{petit_structural_2022}, a rule-based method that implements an {\em exclusive} policy to elicit the notion of anatomical plausibility. In particular, the criteria encoded by the rules are driven by the definition of non-plausibility. Non-plausible streamlines are identified with several heuristics: (i) those streamlines that are shorter than 20 mm, or contain a loop, or are truncated, i.e., they terminate before reaching the WM/GM interface; (ii) outlier streamlines with respect to clustering~\citep{cote_cleaning_2015} of the three main categories of pathways, i.e., associative, projective, commissural. According to a conservative approach, all the remaining streamlines are labeled as anatomical plausible.

%Among the non-plausible, the most prominent are short streamlines, representing the 31.8$\pm$1.2\% of the whole set of non-plausible.   
%can be basic properties of a streamline like the length e.g., streamlines shorter than 20mm are considered np, the presence of a loop, and the location of terminal points e.g. or they can be dependent from the white matter surrounding like truncation i.e.,    

\paragraph{Dataset HCP-IZ}
\outline{Technical: size, diffusion model, tracking, co-registration\\}
The second dataset adopted to test our method is again HCP-based, but in this case, composed of only one averaged brain: tractogram and structural T1 image. The average comes from a set of 100 HCP subjects, for which the tracking has been performed on the estimated diffusivity model~\citep{descoteaux_regularized_2007} using the Unscented Kalman Filter (UKF) Tractography~\citep{reddy_joint_2016}. 10K streamlines were randomly selected from each subject's brain, resulting in a merged tractogram composed of $\sim$1M streamlines. The merge was possible after a step of streamline-based linear registration~\citep{odonnell_unbiased_2012}, which moved all the tractogram to the space of one arbitrarily picked subject. The same affine transformation was applied to the structural T1w images of subjects. Finally, an average T1w was computed by merging all the subjects' T1w through a simple mean operation.      

\outline{Labeling based on known white matter bundles. Cluster-based labeling. Possibility to test the model performance in incremental learning setting. Learn from a single averaged subject.\\}
This dataset is presented in ~\citet{zhang_anatomically_2018} as an atlas of white matter bundles. The bundles are extracted from the average tractogram using the White Matter Analysis clustering~\citep{odonnell_automatic_2007}. First, 800 clusters are generated, and then they are visually inspected and merged to obtain 74 different classes of bundles (see~\citet{zhang_anatomically_2018} for the full list), including 16 classes of superficial U-shape streamlines. However, in this procedure, almost 300 clusters are not merged into a bundle because composed of unknown pathways. We consider all these unknown streamlines as anatomically non-plausible. Moreover, we use the very high number of bundle classes to create multiple split of \p and \np streamlines, emulating a real-world scenario where the labeling evolves incrementally in time. Each split considers the streamlines belonging to certain classes of bundles as plausible and all the others as non-plausible.

%\paragraph{BILGIN-EP}
%\outline{to be confirmed~\citep{mazoyer_bilgin_2016}\\}

%105 subjects from the Human Connectome Project (HCP) were used for the following experiments. The corresponding dMRI images have 1.25 mm isotropic resolution and 270 gradient directions with 3 b-values (1000, 2000, 3000 s/mm2) and 18 b = 0 images (Sotiropoulos et al., 2013). We used the HCP data that had already been processed by the minimal preprocessing pipeline (e.g. distortion correction, motion correction, registration to MNI space and brain extraction had already been completed) (Glasser et al., 2013)

\paragraph{Dataset HCP-IW}
\outline{Technical: size, diffusion model, tracking\\}
With the aim of proving the impact of our filtering method, we adopt a third HCP-based dataset published along with TractSeg~\citep{wasserthal_tractseg_2018, wasserthal_combined_2019}. This dataset is composed of 23 tractograms non-overlapping with the ones of the other HCP-based datasets. The tractograms were obtained using multi-shell multi-tissue CSD model estimation and the Second-order Integration over Fiber Orientation Distributions (iFOD2) probabilistic tracking with MRtrix~\citep{tournier_mrtrix3_2019}. The tracking was performed by: (i) random seeding within the masked brain, (ii) pruning streamlines shorter than 40mm, (iii) cropping streamlines at the GM/WM interface, and (iv) stopping after reconstructing 10M streamlines. Moreover, the tracking was executed twice per subject, once considering anatomical constraints and once not.

\outline{Public benchmark for bundle dissection. Use ground-truth bundles as proxy evaluator for the filtering impact. Test the model on another different type of tractography.\\}
This HCP-IW dataset is one of the few benchmark datasets for bundle segmentation. It contains the labeling of 72 white matter bundles per tractogram. Given the lack of \p versus \np categorization in this dataset (as in all the other publicly available datasets), we use bundles as proxy evaluators to quantitatively and qualitatively show the impact of our method.     

\paragraph{Dataset APSS-IS}
\outline{Technical: size, diffusion model, tracking\\}
The last dataset we adopt is a clinical dataset obtained from the Department of Neurosurgery at the Santa Chiara Hospital (APSS) in Trento (Italy). It comprises 5 patients affected by brain tumors. For each subject, we have available the DWI and the reconstructed tractogram. The DWI was acquired with a 1.5T MR scanner using 60 directions. Then, a single shell b=1000 s/mm$^2$ was extracted to reconstruct the diffusion model with DTI~\citep{pierpaoli_diffusion_1996}. The tracking was performed using Euler Delta Crossing (EuDX)~\citep{garyfallidis_dipy_2014} and produced approximately 100K fibers.

%5 with brain tumor. Tractography: 60 directions, single shell b=1000 s/mm2, diffusion tensor imaging (DTI) reconstruction, Euler Delta Crossing (EuDX) tracking method (Garyfallidis et al., 2014), 106 seeds, approximately 100K streamlines. Bundles: Left inferior fronto-occipital fasciculus (Left_IFOF) and left arcuate fasciculus (Left_AF). Expert-based segmentations: One expert neurosurgeon (S.S.) manually segmented the bundles in the lesioned hemisphere of the patients, who were affected by brain tumors. The lesion however did not affect the shape of the bundles consistently. Bundles where successively refined to remove outliers using the interactive segmentation tool Tractome (Porro-Muñoz et al., 2015) and visually inspected, remaining with 7 instances for each bundle.

\outline{In house (private) real world dataset. Impact of filtering in a practical use-case. Test the model on different tracking and data distribution.\\}
An expert neurosurgeon (S.S.) manually segmented bundles for clinical purposes in both the healthy and lesioned hemispheres of the patients. The manual segmentation followed an ROI-based procedure operated with TrackVis~\citep{wang_diffusion_2007}. Due to the different sizes and locations of tumors, the types of the segmented bundle were not consistent across patients or hemispheres. Among the available segmentations, we selected the ones that were in common with the 5 subjects. The selection resulted in three types of bundles for the healthy hemisphere: the Arcuate Fascicle (AF), the Superior Longitudinal Fascicle (SLF), and the Inferior Fronto-Occipital Fascicle (IFOF), and one bundle in the lesioned hemisphere: Pyramidal Tract (PT).

%%% Local Variables: 
%%% mode: latex
%%% TeX-master: "tractogram_filtering_main"
%%% End:

\section{Experiments and Results}%
\label{sec:experiments}

\outline{Experiments structured into two parts: (i) test model, (ii) test task\\}
The empirical analysis is organized into model-related and task-related experiments. Model-related experiments aim to assess the properties of the proposed model; task-related experiments are designed to investigate the effectiveness of Verifyber as a solution for the task of tractogram filtering. The performance of Verifyber are compared with a selection of the state of the art methods. The sensitivity to the sequential structure of the streamlines is carried out with an ablation study. Finally, the impact of the proposed solution on tractogram filtering is estimated both quantitatively and qualitatively by: (i) looking at the distribution of the misclassification error, (ii) considering the behavior on different types of tractograms, (iii) simulating the evolving definition of anatomical plausibility to test the adaptation in the case of concept drift, and (iv) comparing with unsupervised deep learning tractogram filtering.  

\subsection{Model related experiments}
\label{ssec:model_exp}
\outline{Cross-validation to evaluate performance of proposed method VF\\}
\paragraph{Cross-validation analysis}The first experiment is designed to measure the learning performances of Verifyber according to the usual setting of cross-validation for a supervised task. For this purpose, we consider the HCP-EP dataset with 20 annotated tractograms. The train and test splitting follows a 5-fold cross-validation scheme, where each fold is composed of 4 tractograms and 3.5 million fibers. For each run, the remaining 4 folds are randomly split into 4 buckets, 3 devoted to training and 1 to validation. The training procedure is designed as follows: 1K epochs; cross-entropy loss to optimize the classification; Adam optimizer with default alfa and beta momentum (0.9, 0.99); initial learning rate of $10^{-3}$ multiplied by a factor of 0.7 every 90 epochs until a minimum value of $5 \cdot 10^{-5}$ is reached. In each epoch, we define a mini-batch composed of 16K streamlines, randomly sampled from two subjects, 8K from each of them. A subject is sampled only once for each epoch. The evaluation of the binary classification task is carried out by measuring the accuracy, the precision, the recall, and the Dice-S\o rensen coefficient (DSC). Results are reported in Table \ref{tab:res_extractor}.

\begin{figure*}[th]
  \centering
  \includegraphics[width=.835\textwidth]{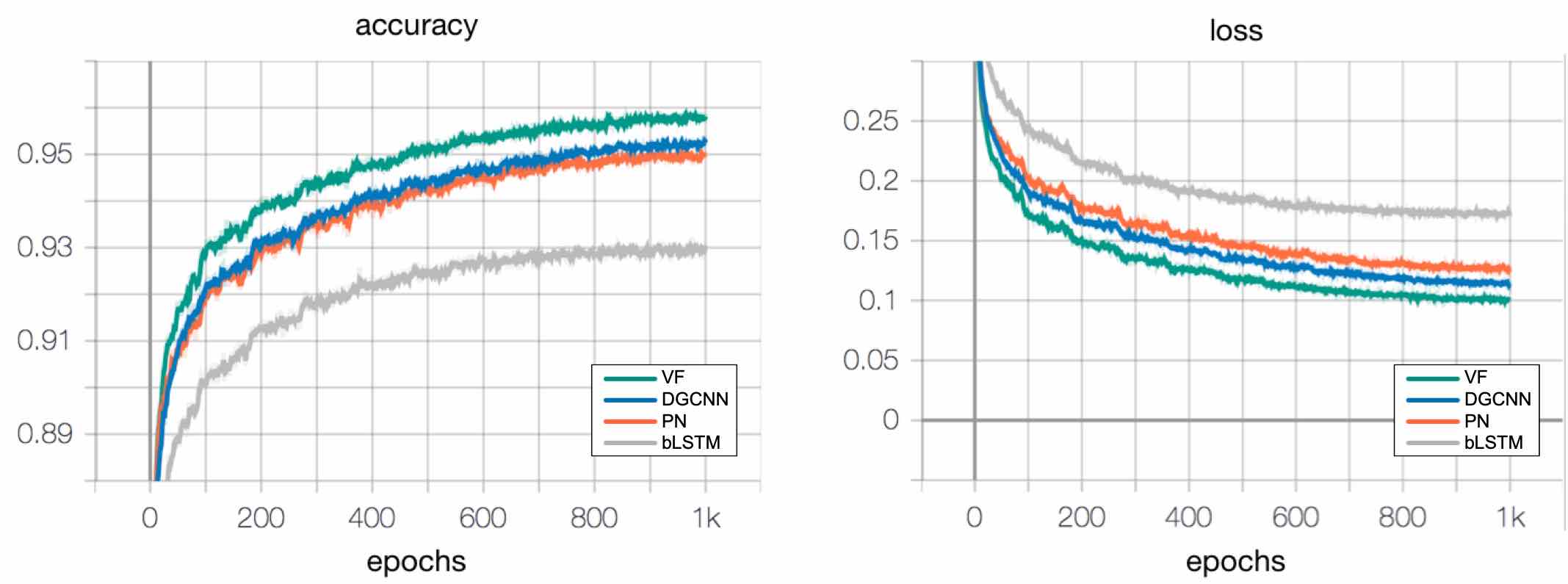} 
  \caption[Training curves on HCP-EP]{Training curves of the four methods compared. Despite the similar number of parameters of all the models we can observe that bLSTM (grey) has clear negative gap with respect to all the other models. Conversely, VF (green) has always a positive gap compared to all the others.
}
  \label{fig:training_ext}
\end{figure*}

\newcolumntype{C}{>{\centering\arraybackslash}p{1.9cm}}
\begin{table*}[ht]
    \centering
    \caption{Accuracy, precision, recall, and DSC 5-fold cross validation scores on HCP-EP dataset. Reported values are the mean and standard deviation across the 5-folds. Each fold of 4 subjects has been used once as a test.% Std columns refer to the mean std across the 4 test subjects, averaged over the 5 folds.
    }
    \label{tab:res_extractor}
    \begin{tabular}{ l | C | C | C | C }
        \toprule
        Method & Accuracy & Precision & Recall & DSC \\
        \midrule
        bLSTM   & 93.0 $(\pm0.1)$ & 93.8 $(\pm0.1)$  & 96.2 $(\pm0.2)$ & 95.0 $(\pm0.1)$ \\
        PN      & 94.7 $(\pm0.1)$ & 95.5 $(\pm0.2)$ & 96.9 $(\pm0.2)$ & 96.2 $(\pm0.1)$ \\
        DGCNN     & 94.4 $(\pm0.1)$ & 95.4 $(\pm0.2)$ & 96.5 $(\pm0.1)$ & 96.0 $(\pm0.1)$ \\
        \textbf{VF}    & \textbf{95.2} $(\pm0.1)$ & \textbf{96.1} $(\pm0.2)$ & \textbf{96.9} $(\pm0.1)$ & \textbf{96.6} $(\pm0.1)$ \\
        \bottomrule
%        DEC$_\mathrm{perm}$  & 94.3 & 95.4 & 96.5 & 95.9 \\
%        VF$_\mathrm{perm}$ & 30.0 & 87.7 & 00.0 & 00.6 \\
%        \hline
    \end{tabular}
\end{table*}

\newcolumntype{C}{>{\centering\arraybackslash}p{1.1cm}}
\begin{table*}[h!]
    \centering
    \caption{Models architecture used for experiments.
    }
    \label{tab:models_desc}
    \begin{tabular}{ l | l | c }
        \toprule
        Method & Architecture & params \\
        \midrule
        bLSTM   & \footnotesize \texttt{MLP(128)$\rightarrow$LSTM(256)$\oplus$LSTM$^{-1}$(256)$\rightarrow$ MLP(256,128)$\rightarrow$FC(2)} & \normalsize 800K \\
        \midrule[0.2pt]
        PN      & \footnotesize \texttt{MLP(64,64,64,128,1024)$\rightarrow$MAX$\rightarrow$MLP(512,256,40)$\rightarrow$FC(2)} & \normalsize 800K \\
        \midrule[0.2pt]
        \multirow{2}{*}
        {DGCNN}   & \footnotesize \texttt{ec1:EC(64,64,64)$\rightarrow$EC(64,64,64,128)$\rightarrow$ec1$\oplus$ec2$\rightarrow$} & \normalsize 800K \\
        	        & \footnotesize \texttt{$\rightarrow$MLP(1024)$\rightarrow$MAX$\rightarrow$MLP(512,256)$\rightarrow$FC(2)} & \\
%        VF    & \textbf{95.2} $(\pm0.1)$ & \textbf{96.1} $(\pm0.2)$ & \textbf{96.9} $(\pm0.1)$ & \textbf{96.6} $(\pm0.1)$ \\
        \bottomrule
    \end{tabular}
\end{table*}

\newcolumntype{C}{>{\centering\arraybackslash}p{1.9cm}}
\begin{table*}[ht]
    \centering
    \caption{Permutation test results on HCP-EP. Reported values refer to only one split of the 5-fold. The mean and standard deviation are computed across the 4 test subjects. Lower values are better. As expected PN and DGCNN proved the permutation invariance maintaining the same results of Table \ref{tab:res_extractor} of the paper.
    }
    \label{tab:res_permutation}
    \begin{tabular}{ l | C | C | C | C }
        \toprule
        Method & Accuracy & Precision & Recall & DSC \\
        \midrule
        bLSTM$_\mathrm{perm}$  & 64.1 $(\pm1.1)$ & 89.8 $(\pm1.0)$& 55.1 $(\pm1.1)$& 68.3 $(\pm0.9)$\\
        PN$_\mathrm{perm}$ & 94.5 $(\pm0.1)$ & 95.4 $(\pm0.2)$ & 96.8 $(\pm0.2)$ & 96.1 $(\pm0.2)$ \\
        DGCNN$_\mathrm{perm}$ & 94.3 $(\pm0.1)$ & 95.4 $(\pm0.3)$ & 96.5 $(\pm0.2)$ & 95.9 $(\pm0.2)$ \\
        \textbf{VF$_\mathrm{perm}$} & \textbf{30.0} $(\pm 2.8)$ & \textbf{87.7} $(\pm 0.7)$& \textbf{00.3} $(\pm 0.0)$& \textbf{00.6} $(\pm 0.1)$\\
        \bottomrule
    \end{tabular}
\end{table*}

\outline{Comparison with state of the art learning models: bLSTM, PN, DGCNN\\}
\paragraph{Comparison with  deep learning models} The second experiment aims to compare Verifyber with the competing deep learning models, namely bLSTM, PN, and DGCNN. As reference dataset we consider HCP-EP as above. In this experiment, we operate the 5-fold cross-validation setting adopted for Verifyber to all other methods and we measure the same evaluation metrics.
% For a fair comparison, we enforce a uniform configuration among the different models: the choice of the input representation, the number of parameters of the NN models, and the hyper-parameters for training.
We report the behavior of training curves for all models, both accuracy and loss, in Figure~\ref{fig:training_ext}. 

Regarding the input representation, GDL models can deal with size-varying input, e.g., streamlines with different numbers of points, while bLSTM requires a fixed vectorial representation in input, like common learning models. For this reason, we need to resample the points of all streamlines to be a fixed number. According to previous works~\citep{odonnell_automatic_2007, garyfallidis_quickbundles_2012}, the common choices are resampling to 12, 16, or 20 points per streamline. Since a side empirical assessment analysis did not provide any significant difference in performance, we operate a resampling to 16 points for all the subsequent experiments. To prevent the bias of heterogeneous size of models, we set the architecture of the different methods with a uniform number of parameters, as reported in Table~\ref{tab:models_desc}.

\outline{Results comparative analysis of accuracy: Table/Figure (ref?)\\}
The results of the comparison on the test set are reported in Table \ref{tab:res_extractor}. According to the related results at the end of the training illustrated in Figure~\ref{fig:training_ext}, we may conclude that there is no overfitting of the model during the learning process. While the performance of geometric deep learning models (PN, DGCNN, VF) are quite similar, there is a meaningful gap with respect to the recurrent neural network model. A t-test between bLSTM and PN reports a p-value $< 10^{-3}$.
 
%we show the performance achieved by all methods on the HCP-EP dataset. The scores obtained by Verifyber are the highest and most stable across different metrics, even though the gap compared to PN and DGCNN is small. The lowest performance is obtained by bLSTM, for which the computation of t-test provides a statistical significance, with a p-value $< 10^{-3}$. The inspection of behavior at training time, see the Figure~\ref{fig:training_ext}, confirms a poor minimization of the loss for bLSTM compared to other GDL models.

\outline{Test for sensitivity to points sequence of streamlines\\}
\paragraph{Permutation invariance test} A requirement for the learning model based on streamlines is the sensitivity to the sequence order of points. Both Verifyber and DGCNN capture the notion of context by taking into account the neighbors of a point. Nevertheless, the working assumption is that Verifyber is exploiting more carefully the sequential relation of points in a fiber. For this purpose, we design a simple permutation test where the order of points in a fiber is randomly permuted. The side effect is to generate pathways potentially anatomically non-plausible. We then operate the inference on this new test set using just one split of 5-fold cross-validation since this experiment is not sensitive to the selection of the individuals. The results evaluated with the previous metrics are shown in Table~\ref{tab:res_permutation}. As expected, the performance of Verifyber drops to $30.0\%$ of accuracy and $0.0\%$ of recall because permuted fibers are classified as anatomically non-plausible. On the other hand, both DGCNN and PN preserve the previous scores, $94.3\%$ and $94.5\%$ respectively, because these models are invariant to the order of points.

\subsection{Task Related Experiments}
\label{ssec:task_exp}

\begin{figure}[h!]
    \centering
    \includegraphics[width=.8\linewidth]{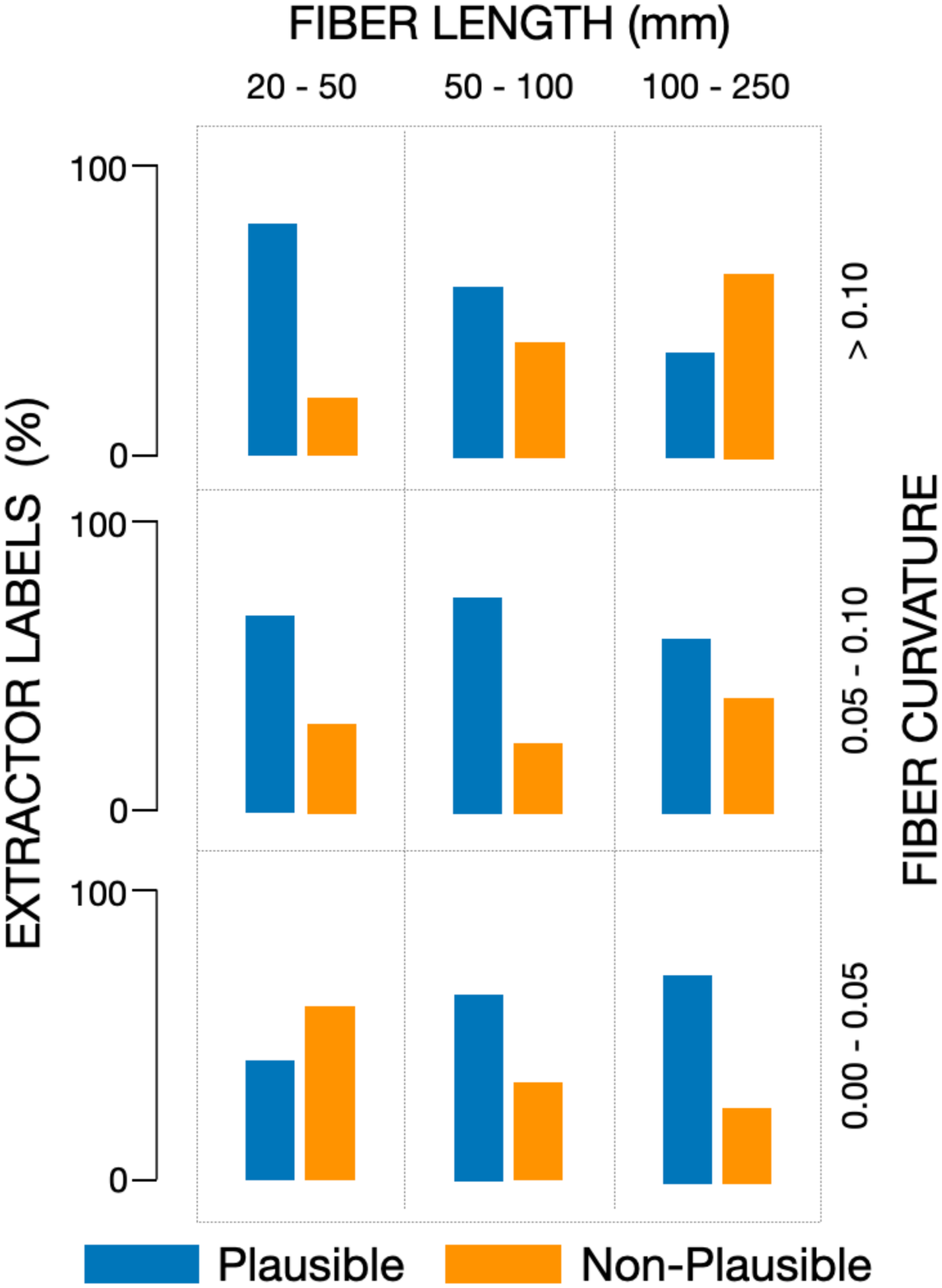}
    \caption{Distribution of Extractor labels per streamline category.}
    \label{fig:streams_categories}
\end{figure}

\begin{figure*}[ht]
    \centering
        \begin{subfigure}[t]{0.4\linewidth}
        \centering
        \includegraphics[width=\textwidth]{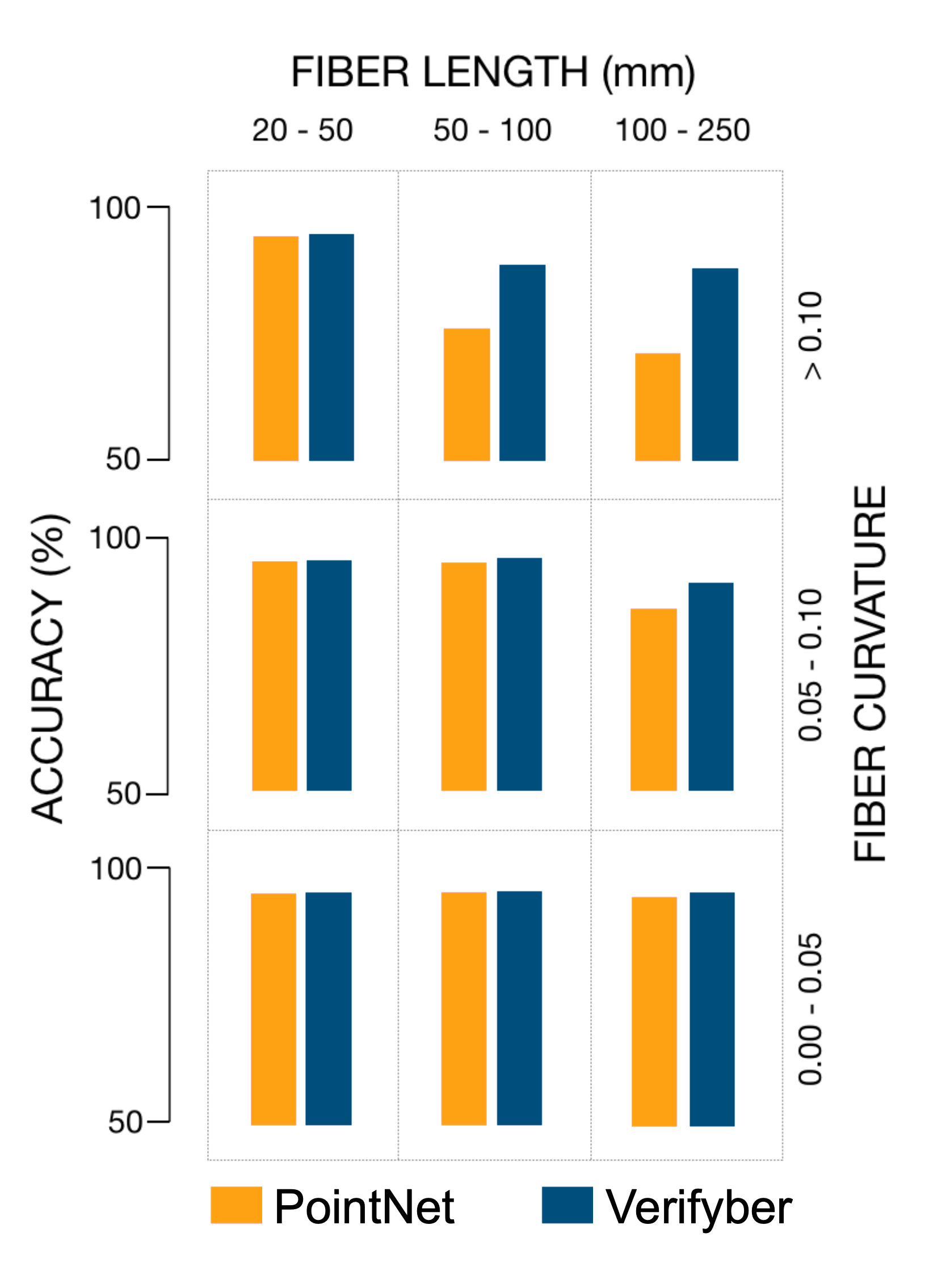}
        \caption{}
        \label{fig:acc_pn_sdec}
    \end{subfigure}
    \begin{subfigure}[t]{0.4\linewidth}
        \centering
        \includegraphics[width=\textwidth]{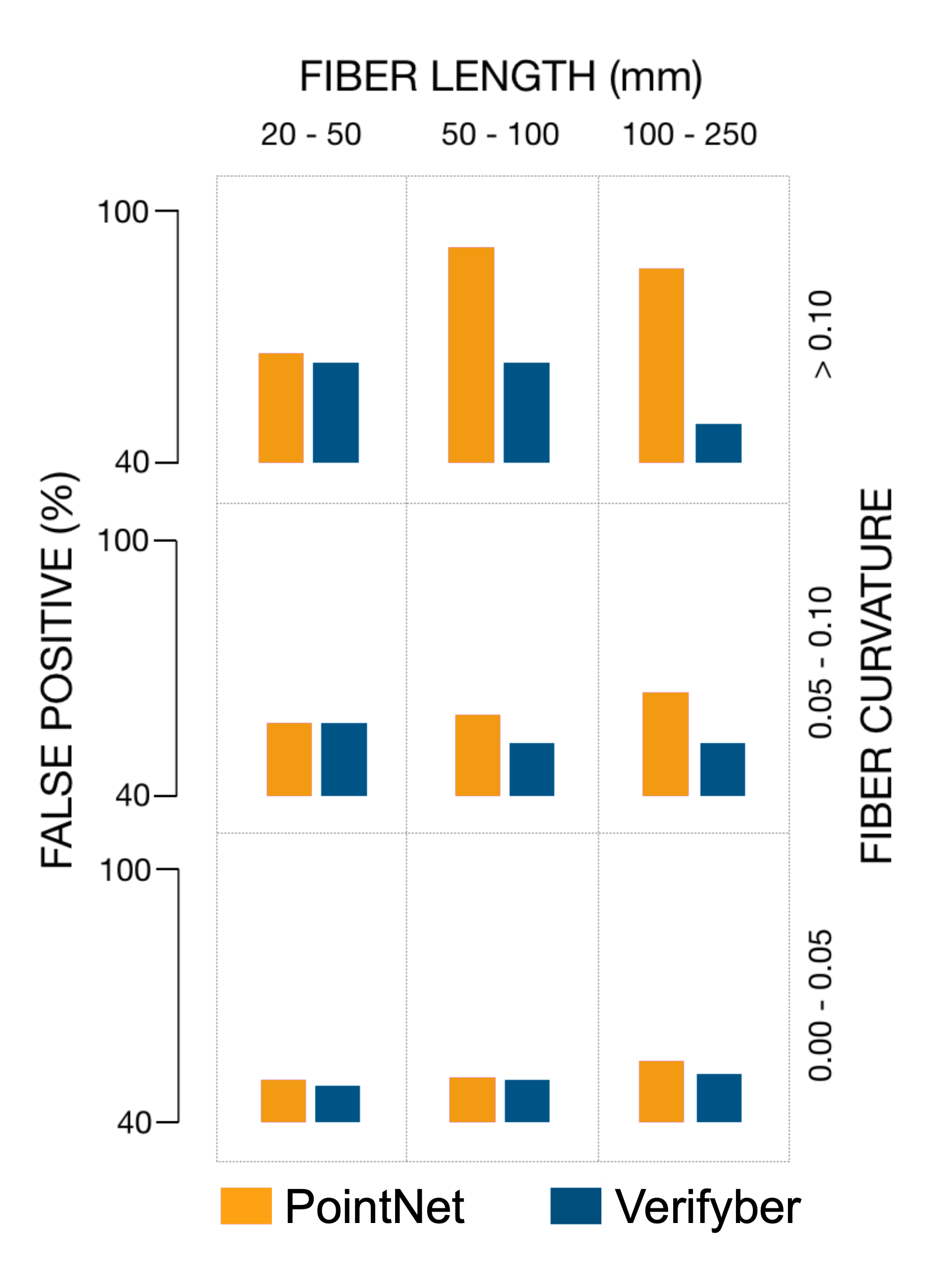}
        \caption{}
        \label{fig:fp_pn_sdec}
    \end{subfigure}
    \caption[Permormance analysis of VF vs. PN]{Comparison between PN and VF with respect to 9 streamlines categories that differ for curvature and length. We report the comparison in terms of accuracy (a) and false positive percentage (b).}
  \label{fig:acc_fp_pn_sdec}
\end{figure*}

\outline{Error analysis for type of streamlines\\}
\paragraph{Misclassification analysis}
% A second purpose of the empirical investigation is to understand how the different models behave with respect to the geometrical and anatomical properties of fibers.
The geometrical properties of fibers might be captured considering two features: the length and the curvature. The combination of these features may represent a good proxy of the anatomical properties. To analyze the misclassification error with respect to these features, We define a partition of fibers, according to their length, into three intervals: short [0, 50] mm, medium [50,100] mm, long [100,300] mm. Similarly, we operate a partition over the mean curvature: straight [0.0,0.05], curved [0.05, 0.10], very curved [0.10, 0.20]. The partitions are designed to have at least $15\%$ of fibers in each interval. Combining the intervals of length and curvature, we obtain 9 groups for fibers in the HCP-EP dataset, as reported in Figure~\ref{fig:streams_categories}.

%\paragraph{Comparison with PointNet~\citep{qi_pointnet_2017}}
We are interested to investigate how the misclassification error differs between Verifyber and PointNet, more precisely the difference in considering the point cloud only with respect to the edges. We proceed by looking at these groups and inspecting where the predictions fail to discriminate between anatomically plausible and non-plausible fibers. Despite a similar score of classification accuracy, the two methods share only $60\%$ of the error while the remaining $40\%$ concerns different fibers. The distribution of the error with respect to these 9 groups is reported in Figure~\ref{fig:acc_pn_sdec}. 

\begin{figure*}[th]
	\begin{tabular}{cc}
	\begin{minipage}{0.47\textwidth} 
		\includegraphics[width=1.0\textwidth]{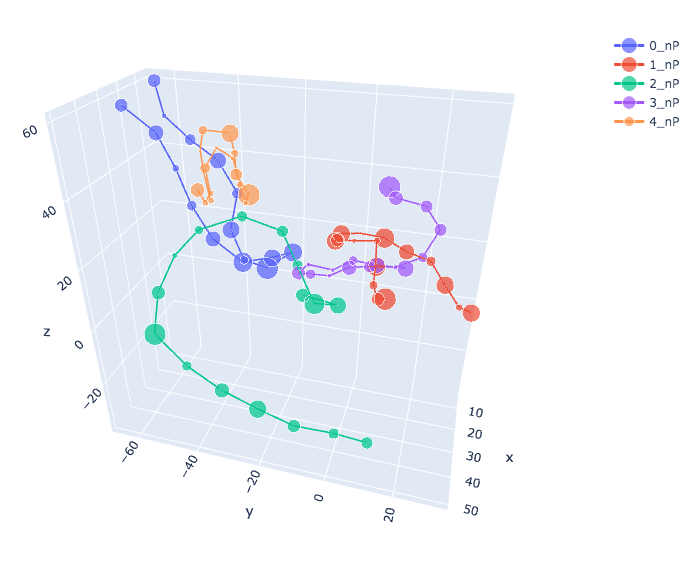} 
		\subcaption{PN}
    \end{minipage}
    &
    \begin{minipage}{0.47\textwidth}
    	\includegraphics[width=1.\textwidth]{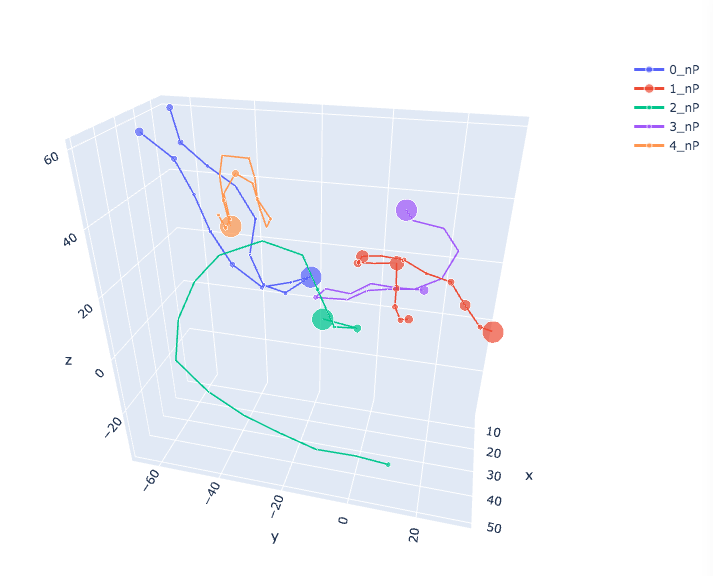}
    	\subcaption{Verifyber}
    	\end{minipage}
    \end{tabular}
  	
\caption[Point-wise max activation of Verifyber and PN]{Given a set of streamlines, we compare the per-point contribution to the classification performed by PN and VF. The reported streamlines are all non-plausible and belong to the category of long and curved where the misclassification error is greater. We analyzed the amount of contribution of each point to the global max pooling present in both PN and VF. The size of points indicate their importance with respect to the single fiber descriptor generated by the pooling. We observe how PN tends to maintain a uniformly distributed importance, while VF seems able to individuate few strategic points for the filtering task.}
\label{fig:max_activations}
%\vspace{-1cm}
\end{figure*}

\outline{Evaluation of VF vs PN\\}
Figure~\ref{fig:acc_pn_sdec} highlights that on longer and more curved fibers, Verifyber outperforms PointNet. An interpretation of the source of such a difference can be obtained by looking at the internal representation of the two models. The relevance weights associated with each fiber point are uniformly distributed in PointNet, whereas in Verifyber the learning process clearly identifies a few more discriminating points. We show visual evidence of this difference for long and more curved fibers in Figure~\ref{fig:max_activations}. 

\begin{figure*}[ht]
  \centering
  \includegraphics[width=.835\textwidth]{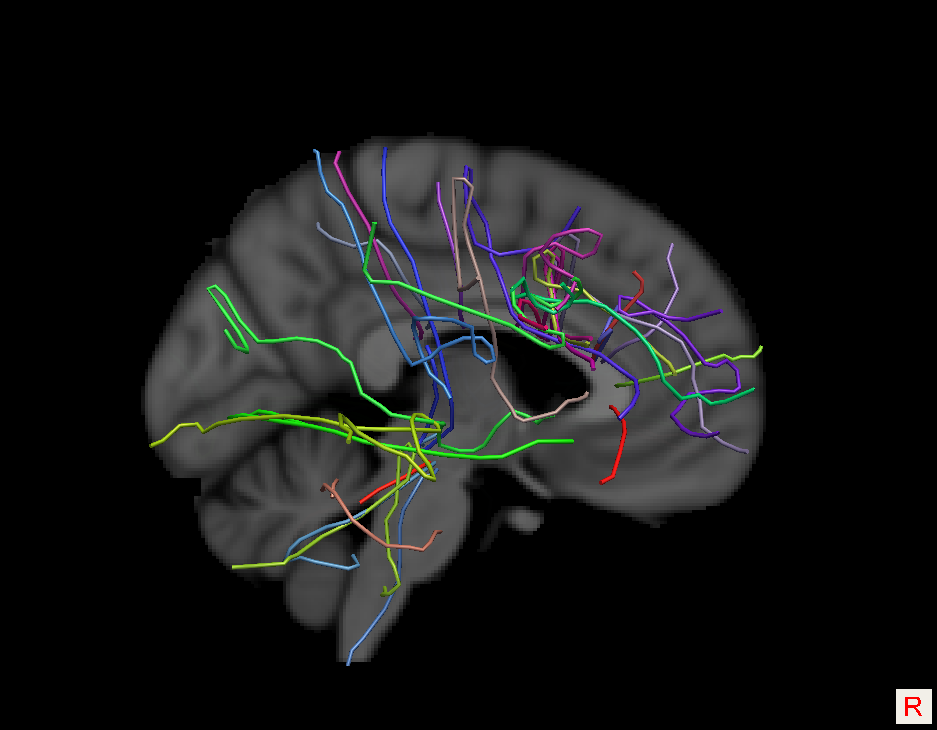}
  \caption{Example of FN i.e., labeled as plausible but classified as non-plausible, produced by VF. The streamlines shown are clearly non-plausible showing the presence of noise in the rule-based labeling, and the capacity of VF of extending the concept learnt by the rules.}
  \label{fig:fn_sdec}
\end{figure*}

\outline{Evaluation of False positive error component: Table/Figure (ref?)\\}
\paragraph{In-depth error characterization} We deepen our analysis by focusing our attention on the false positive rate, i.e., the tendency of misclassifying non-plausible fiber as plausible. We neglect the false negative rate in this analysis because Extractor~\citep{petit_structural_2022} adopts an exclusive policy to label the fibers as anatomically non-plausible, i.e., is more sensitive to false positive error. False negative error is qualitatively investigated later.  In Figure \ref{fig:fp_pn_sdec}, we show how false positive error differs between Verifyber and PointNet. Even in this case, Verifyber behaves better than PointNet, meaning lower false positive rate when fibers are long and curved. PointNet has a clear bias to classify those fibers as anatomically plausible, while Verifyber is more robust and keeps the false positive rate consistently in the range of $50-60\%$ across all the groups of fibers. More in detail, the worst performance of PointNet are for medium length and very curved fibers ($91.4\%$), long and very curved ($86.5\%$),  long and curved ($64.2\%$), where the rates of Verifyber are $64.1\%$, $49.5\%$ and $51.1\%$ respectively. 

\outline{Interpretation of False Negative error component: Table/Figure (ref?)\\}
Considering the conservative approach adopted by exclusive labelings, it might be interesting to inspect the false negative error qualitatively. In this case, the goal is to evaluate whether anatomically plausible fibers misclassified as non-plausible can be considered controversial due to the noisy process of ground truth definition. For this purpose we operate a visual inspection on a random sample of misclassified fibers, as reported in Figure~\ref{fig:fn_sdec}. Although those fibers are labeled as anatomically plausible, --- probably because considered unknown by the exclusive policy --- a manual survey by an expert anatomist confirms the classification of Verifyber as anatomically non-plausible.

\newcolumntype{C}{>{\centering\arraybackslash}p{1.4cm}}
\begin{table*}[h!]
    \centering
    \caption[Incremental results on HCP-IZ]{VF results on incremental learning setting in HCP-IZ dataset. The model is trained always with the same configuration and hyperparameters. The labeling changes incrementally: first row considers the streamlines of Association (A) bundles as plausible and the rest as non-plausible; in the second row the labeling of plausible is incremented considering also the streamlines of Projection (P) bundles; similarly the third and fourth rows add the streamlines of Commissural (Co) and Cerebellar (Ce) bundles. Finally, the last row adds superficial U-shape fibers (Sup) as plausible. This corresponds to considering the whole \citet{zhang_anatomically_2018} atlas as plausible and the rest as non-plausible.
    }
    \label{tab:res_zhang}
    \begin{tabular}{ l | l | C | C | C | C }
        \toprule
        Method & Plausible & Accuracy & Precision & Recall & DSC \\
        \midrule
%        DeepWMA & A+P+Co+Ce & 91.0\tablefootnote{result taken from~\citep{zhang_deep_2020}. The score is obtained using the same dataset split 80/20\% for train and test as in our case, but with a different random sampling of the two sets. In~\citep{zhang_deep_2020} the training is performed multiclass using the 54 classes of bundles plus 1 class of ``other'', while we trained using only the two classes of plausible and non-plausible corresponding to the 54 bundles and ``other'' respectively.}
%         & - & - & -\\
%        \hline 
        \textbf{Verifyber} & A  & 98.8 & 96.4 & 96.0 & 96.2 \\
        & A+P & 98.0 & 96.4 & 95.9 & 96.1 \\
        & A+P+Co & 97.9 & 97.0 & 96.6 & 96.8 \\
        & A+P+Co+Ce & 97.8 & 97.1 & 96.4 & 96.7 \\
        & A+P+Co+Ce+Sup & 97.1 & 97.6 & 98.0 & 97.8 \\
        \bottomrule
    \end{tabular}
\end{table*}

\begin{figure*}[t]
  \centering
  \includegraphics[width=\textwidth]{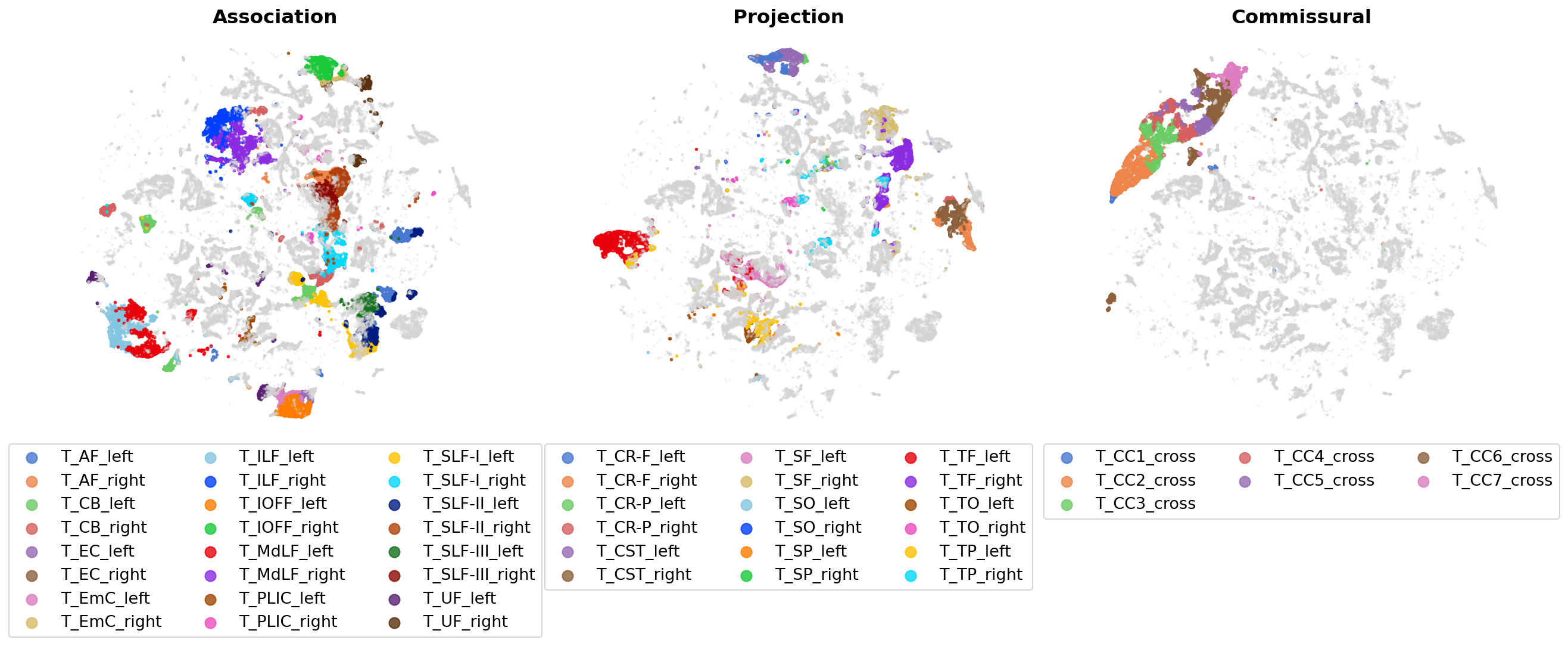}
  \caption[Latent space learned by Verifyber on HCP-IZ]{Latent space learned by \vfiz. Each point corresponds to a streamline in the 1024D space reduced to 2D by means of t-SNE~\citep{van_der_maaten_visualizing_2008}. The three plots show the non-plausible streamlines in light-gray, together with one macro category of bundle, i.e., association, projection, commissural at a time.}
  \label{fig:tsne_emb}
\end{figure*}

\outline{Incremental learning with Zhang atlas\\}
\paragraph{Incremental learning}
The qualitative analysis of false negative fibers highlights the issue of ground truth. Inconsistencies or mistakes in the definition of anatomical plausibility are not only related to the manual labeling process. The debate on human brain anatomy is an ongoing challenge, and the knowledge of white matter pathways is constantly evolving. In the machine learning literature, this circumstance is known as concept drift. For this reason, we need to investigate how Verifyber might be robust when the ground truth is incrementally updated.

We design a simulation where the labeling of fibers is revised at different stages by adding new knowledge following an inclusive policy. For this analysis, we refer to the HCP-IZ dataset and the categories of bundles defined in the related atlas: association, projection, commissural, cerebellar, and superficial. In the first stage, only fibers of association bundles are labeled as anatomically plausible, non-plausible otherwise. In the second, third, and fourth stages, we add the fibers of projection, commissural and cerebellar bundles, respectively. Finally, in the fifth stage, we consider the fibers of all bundles defined in the atlas, i.e., deep and superficial bundles. Even though the HCP-IZ dataset is the result of processing hundred of individuals, it is composed of only a single average tractogram. For this reason, we organize the training set by randomly picking $80\%$ of fibers and the test set with the remaining $20\%$. For all stages, we carry out a training process using the same hyperparameters described above.

\outline{incremetal results on zhang\\}
Table~\ref{tab:res_zhang} shows the results of the incremental learning for the five stages. The scores confirm that Verifyber is considerably stable and does not suffer the drift of anatomical plausibility. However, we may notice a small decrease in accuracy, compensated by an increase of DSC, when new groups of bundles are added to the ground truth. This behavior is explainable by the balance shift between the number of plausible and non-plausible fibers.

\outline{T-SNE analysis on zhang\\}
\paragraph{T-SNE analysis of learned features} A deeper analysis of the results can be carried out by looking at the latent space learned by Verifyber after the training process. In the latent space, each fiber is encoded into a vector of 1024 dimensions. We may visualize this space by projecting all the fibers into a two-dimensional plot by means of t-SNE~\citep{van_der_maaten_visualizing_2008} as reported in Figure \ref{fig:tsne_emb}. Using a color scheme, we highlight the proximity of fibers that belong to the same bundle. It is worth noting that even the proximity among the bundles is preserved, e.g., AF is close to SLF-II and SLF-III, CC[1-7] are almost consecutive, MdLF is close to ILF. Lateralized bundles are well separated from each other, e.g., IFOF left and right. There is consistency in the lateral grouping of similar bundles, i.e., if a left bundle is close to another left bundle, the corresponding right bundles are close too.           

%\outline{Test performance across dataset: train HCP, test GIN\\}
%PENDING
%The results of filtering on the GIN dataset (one subject 1M streamlines) using the model trained on one of the 5-fold split are: bLSTM 91.1, PN 92.0, DGCNN 92.8, VF 93.4.  

\paragraph{Comparison with an unsupervised method}
An interesting open question is investigating how our supervised model behaves compared to a method such as FINTA~\citep{legarreta_filtering_2021}; a state-of-the-art approach for unsupervised tractogram filtering based on deep learning. Specifically, FINTA employs a convolutional autoencoder, which, similarly to our case, is trained directly on the raw streamline structure. We believe that such a comparison might be relevant to clarify the difference between unsupervised and supervised approaches. Unfortunately, neither the code nor the data used in \citet{legarreta_filtering_2021} has been publicly distributed. For this reason, we re-implemented FINTA, following the methodological description provided by the Authors in their article. We publish our FINTA implementation, see Section~\ref{ssec:t_filt_code}.

The tractogram filtering in FINTA is designed as a two steps process: (i) computation of a latent space for streamlines representation; (ii) filtering with a lazy classifier based on a thresholded nearest neighbor rule. Analogously to what is carried out in~\citep{legarreta_filtering_2021}, we train the autoencoder of FINTA with a single average tractogram, the one of HCP-IZ, whose streamlines are randomly split into 80/20 partition for train and test respectively. The training is performed using the hyper-parameters reported by the authors where possible; default values are used otherwise. We trained the autoencoder until convergence (see Appendix). In addition, we performed a qualitative assessment of streamlines that were reconstructed with the trained autoencoder to double-check that the reproduced approach worked as expected (see Appendix). After the computation of the latent space on HCP-IW we proceed with the inference both on HCP-IZ and HCP-EP. For each of the two datasets we tuned the choice of the threshold value for the nearest neighbor rule. The tuning is carried out on a random subsample of streamlines from the validation set.

In Table \ref{tab:res_finta}, we report the values of accuracy, precision, recall, and DSC for the HCP-IZ and the HCP-EP dataset. We observe that on HCP-IZ FINTA achieves the scores in line with the results published in ~\citet{legarreta_filtering_2021}. Nevertheless, the performance of Verifyber is $\sim$6.5\%p  higher in terms of DSC and $\sim$9.1\%p in accuracy. However, the gap becomes even more consistent if we look at the result on HCP-EP, where FINTA has a significant drop in precision.

\begin{table}[h]
\centering
    \caption[Comparison with FINTA]{FINTA performance evaluation. We report the test score of accuracy, precision, recall, and DSC on HCP-IZ all ({\em p} = A+P+Co+Ce+Sup), and HCP-EP. The training of the FINTA autoencoder is performed using the dataset HCP-IZ. Then, we select a portion of plausible ({\em p}) fibers, which are used as anchors for the embedded nearest neighbor step of FINTA. The selection is performed from the training set of the dataset at hand.}
    \label{tab:res_finta}
    \begin{adjustbox}{width=\linewidth}
    \begin{tabular}{ l | c | c | c | c | c }
        \toprule
        Method %& Thr 
        & {\em p/np} label & Acc & Prec & Rec & DSC \\
        \midrule
%        \rowcolor[RGB]{240,240,240}
%        FINTA & 13.6 &\small{\citet{legarreta_filtering_2021}} & 91.0 & 91.0 & 91.0 & 91.0 \\
%        \midrule
        \multicolumn{6}{c}{HCP-IZ all} \\
        \midrule
        FINTA %& 36.9 
        &\multirow{2}*{\small{\citet{zhang_anatomically_2018}}} & 88.0 & 87.3 & 95.8 & 91.3 \\
        \textbf{Verifyber} % & - 
        &  & \textbf{97.1} & \textbf{97.6} & \textbf{98.0} & \textbf{97.8} \\
        \midrule
        \multicolumn{6}{c}{HCP-EP} \\
        \midrule
        FINTA %& 46.0 
        & \multirow{2}*{\small{\citet{petit_structural_2022}}} & 74.3 & 75.1 & 94.6 & 83.8 \\
%        74.3(1.0) & 75.1(1.6) & 94.6(0.3) & 83.8(1.0)
        \textbf{Verifyber}  %& - 
        &  & \textbf{95.2} & \textbf{96.2} & \textbf{96.9} & \textbf{96.6} \\
%        95.2(0.1) & 96.2(0.3) & 96.9(0.1) & 96.6(0.2)
        \bottomrule
    \end{tabular}
    \end{adjustbox}
\end{table}

\begin{figure*}[t]
\centering
  \includegraphics[width=\textwidth]{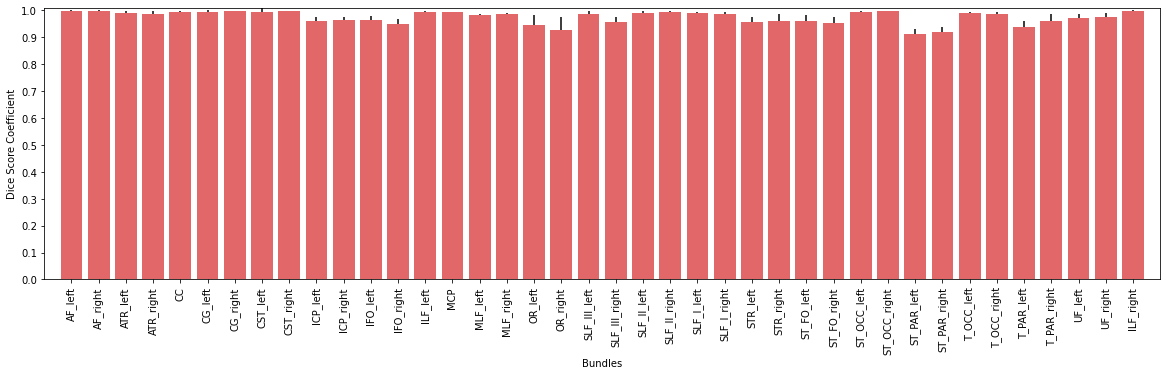}
  \caption[\vfiz quantitative results on HCP-IW]{Volumetric DSC of Tractseg bundles after filtering using \vfiz. Reported mean and standard deviation refers to 23 subjects. The filtering does not impact significantly the shape of the bundle guaranteeing at least 85\% of DSC. In most of the cases the DSC is above 95\%. The mean DSC is $97.5 \pm 2.4$.}
  \label{fig:ts_dsc_sdec_zhang}
\end{figure*}

\outline{Model deployment of VF trained on Zhang and on extractor using tractseg ground truth as evaluation metrics}
%A further question is to evaluate how such a latent space might be dataset specific. 
\paragraph{Generalization across data sources}
In the previous experiments, we trained Verifyber on HCP-EP and HCP-IZ datasets, in which tractography and ground truth policy differ. An open question is to assess how these models behave on unseen tractograms. For this purpose, we design an experiment to perform inference on a new dataset, namely HCP-IW, using the models trained on HCP-EP and HCP-IZ. Since a ground truth is unavailable on HCP-IW, we need to revise the evaluation procedure. The segmented bundles in HCP-IW might be considered fiducial regions, where to focus for quantitative and qualitative analysis. We limit our analysis to 40 most common bundles, those shared with HCP-IZ, out of the 72 available bundles. The list of chosen bundles is shown in Figure \ref{fig:ts_dsc_sdec_zhang}. 

We carry out the inference on the fibers of these bundles, then perform a quantitative and qualitative analysis of potential false negative error since the expected prediction should be only anatomically plausible by design. As a quantitative measure, we compute the volumetric DSC score between the mask of the original bundle and the mask of fibers classified as anatomically plausible. The working assumption is that a moderate false negative error would not affect the estimate of the volumetric region of a bundle. We deepen our evaluation with a qualitative analysis by visually inspecting a sample of fibers classified as anatomically non-plausible, i.e., drawn from the portion of potential false negative fibers. We replicate this procedure both on HCP-IZ and the $23$ individuals distributed as test set from HCP-IW datasets. In addition, we investigate the agreement of the two models trained on two different ground truths, i.e., \vfep and \vfiz.
 
\begin{figure*}[t]
\centering
  \includegraphics[width=\textwidth]{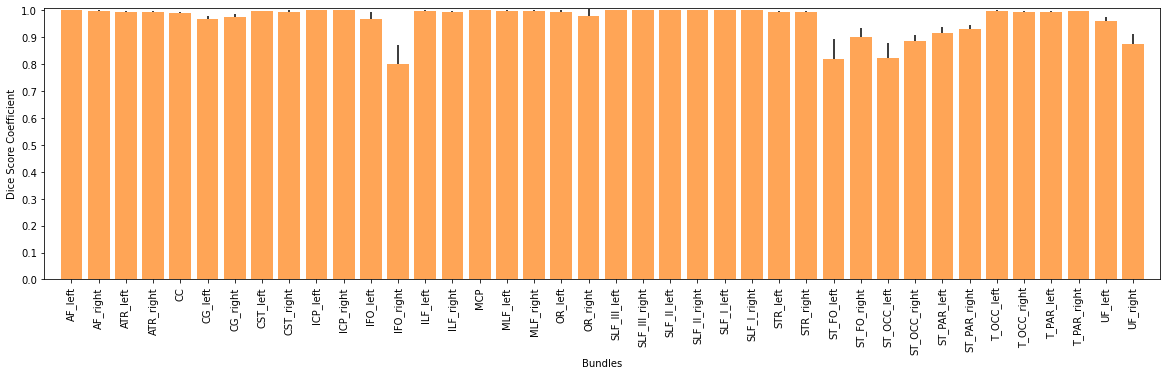}
  \caption[\vfep quantitative results on HCP-IW]{ Volumetric DSC of Tractseg bundles after filtering using \vfep. Reported mean and standard deviation refers to 23 subjects. The filtering does not impact significantly the shape of the bundle guaranteeing at least 80\% of DSC. The mean DSC is $96.8 \pm 5.4$.}
  \label{fig:ts_dsc_sdec_ext}
\end{figure*}

\begin{figure*}[t]
  \centering
  \includegraphics[width=.835\textwidth]{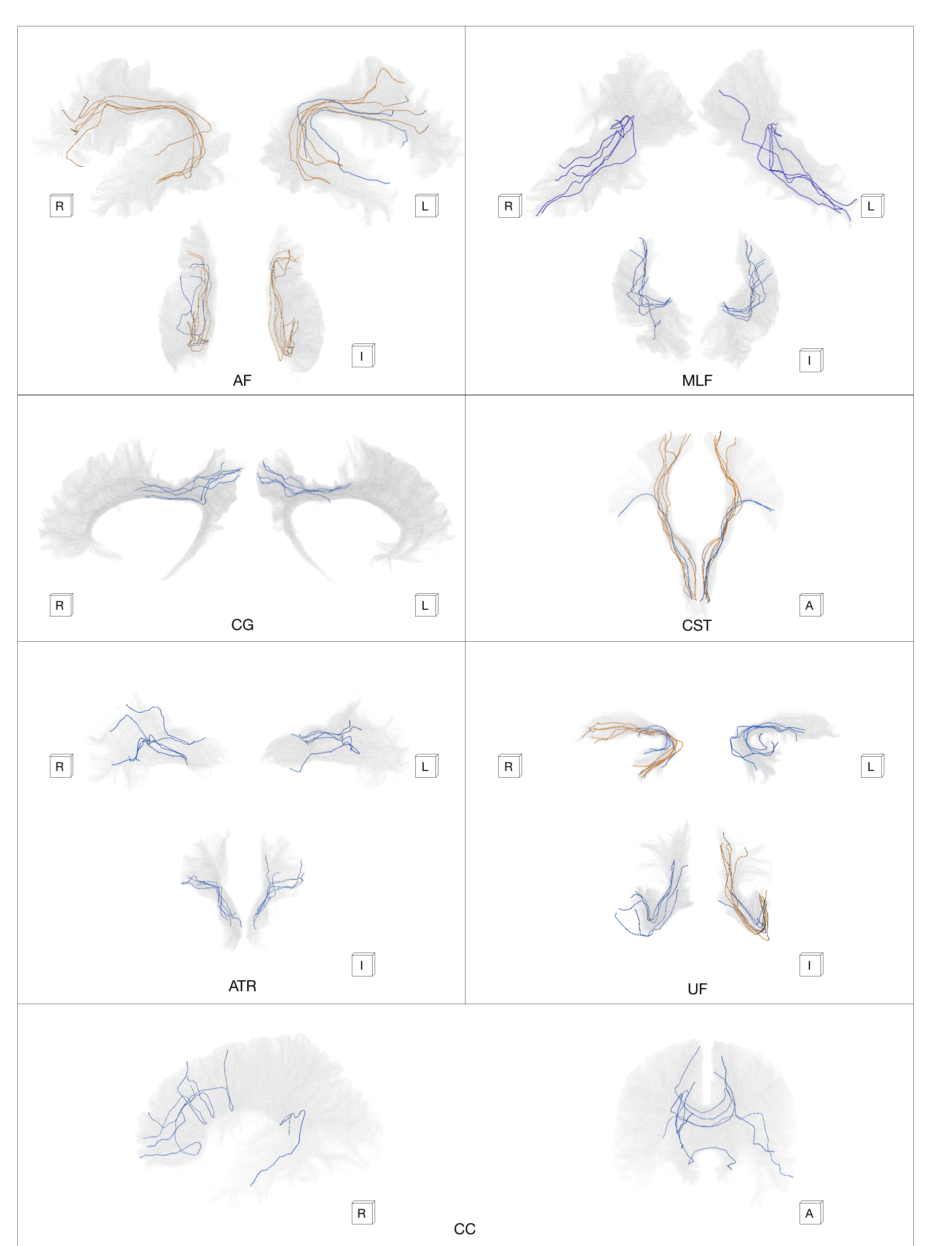}
  \caption[\vfep qualitative results on HCP-IW]{Qualitative example of non-plausible streamlines belonging to Tractseg bundles individuated using \vfep. The figure shows in blue streamlines individuated as non-plausible both by \vfep and \vfiz (shared non-plausible). The orange streamlines are instead exclusive of \vfep. In some bundles there is only a very small number of shared non-plausible, because the amount of non-plausible found by both model on that bundle is globally very low e.g., the AF bundles has 0.001\% of non-plausible fibers found globally both when we use \vfep and when we use \vfiz.}
  \label{fig:ts_qual_extractor}
\end{figure*}

\begin{figure*}[t]
  \centering
  \includegraphics[width=.835\textwidth]{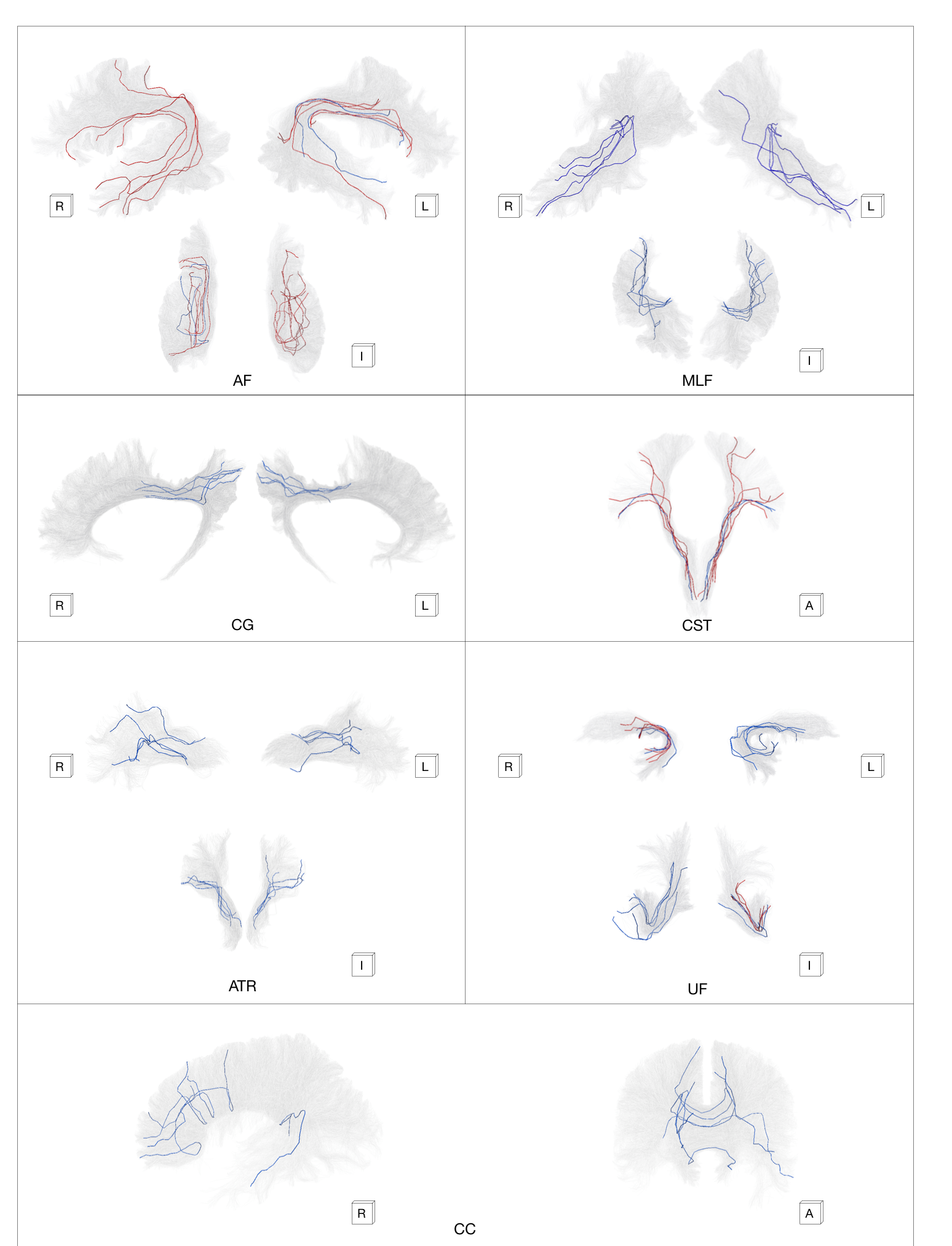}
  \caption[\vfiz qualitative results on HCP-IW]{Qualitative example of non-plausible streamlines belonging to Tractseg bundles individuated using \vfiz. Blue streamlines are shared non-plausible examples, while red streamlines are exclusive non-plausible of \vfiz.}
  \label{fig:ts_qual_zhang}
\end{figure*}

\outline{results tractseg + extractor + zhang}
In Figures~\ref{fig:ts_dsc_sdec_ext} and~\ref{fig:ts_dsc_sdec_zhang}, we report the volumetric DSC score obtained with Verifyber trained on HCP-EP and HCP-IZ respectively. In both the plots, the average value of DSC is above $0.95$ while the minimum does not fall below $0.80$. The qualitative analysis of the filtering is depicted in Figures~\ref{fig:ts_qual_extractor} and~\ref{fig:ts_qual_zhang}. A small sample of fibers misclassified as false negative is selected from a few common bundles. Even though these fibers belong to segmented bundles, the visual inspection by an expert confirms that their pathways are anatomically non-plausible. It is worth noting that even if a fiber falls in the volumetric mask of a bundle not necessarily the related pathway is anatomically plausible. The color scheme highlights those fibers that are classified as anatomically non-plausible from both models.
 
\FloatBarrier

\balance

\outline{Model deployment VF on APSS\\}
\paragraph{Model deployment on clinical data}
As an additional evaluation, we investigate how the VF learnt model behaves when inference is carried out on clinical data. This experiment considers the patients with tumors of the APSS-IS dataset, and we filter their tractograms with \vfep. Note that the choice of filtering with \vfep rather than \vfiz is driven by the more conservative approach of the underneath labeling. Similar to the experiments involving the HCP-IW dataset, we refer to the segmented bundles as a proxy evaluator of plausible streamlines. However, differently from HCP-IW, the quality of bundles in clinical data is lower since the time restrictions on DWI sequence acquisition. The shape of the segmented bundles is more sensitive to the removal of hypothetical non-plausible streamline, therefore we cannot carry out the quantitative analysis previously adopted for HCP-IW. Hence, we proceed with a qualitative analysis by visual inspection of three different bundles from the healthy hemisphere and one bundle from the tumored hemisphere for each of the 5 subjects.

\begin{figure*}[t]
  \centering
  \includegraphics[height=\textheight]{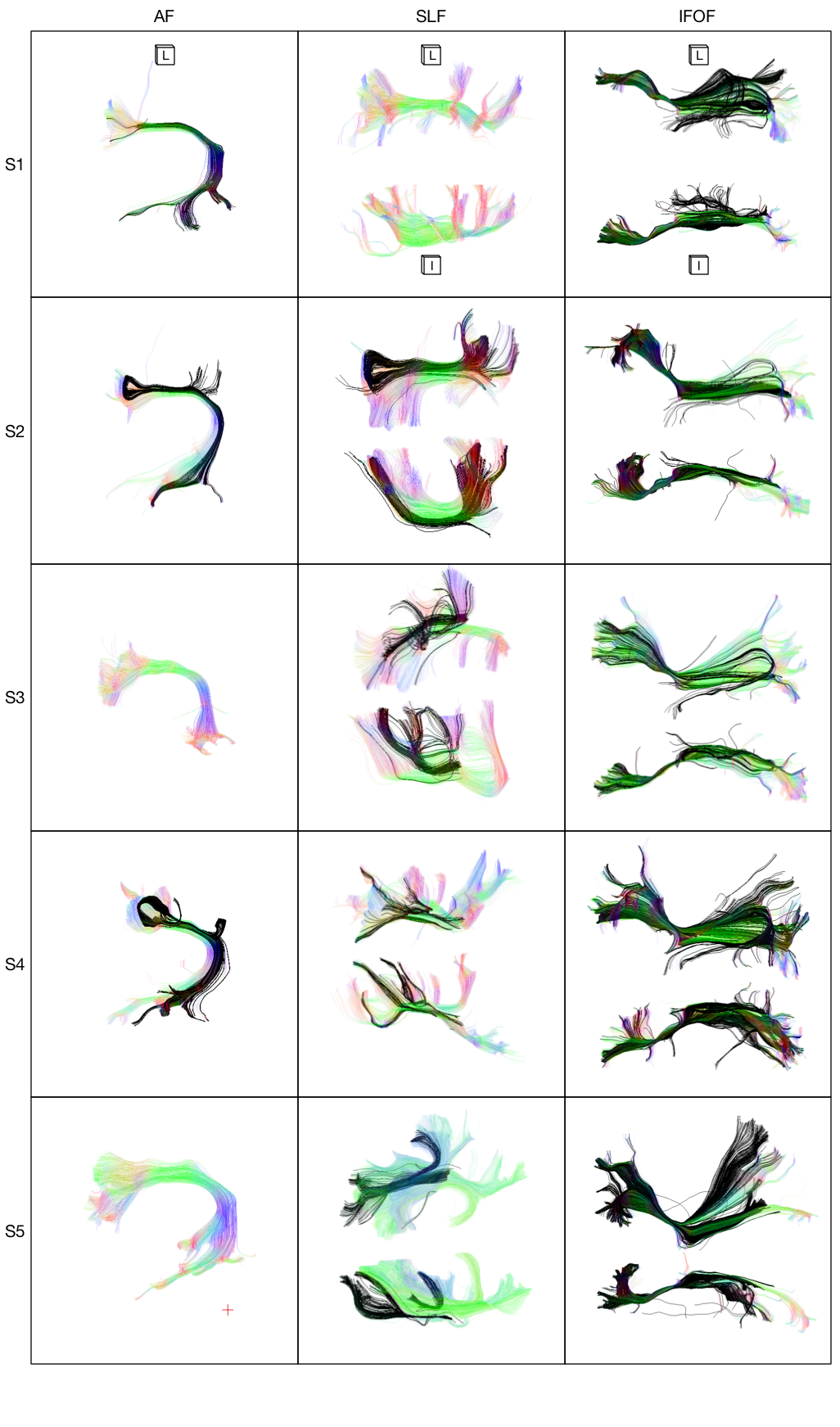}
  \vspace{-1cm}
  \caption[Qualitative results on clinical data. Healthy hemisphere]{Qualitative results after filtering APSS clinical dataset using \vfep. Black streamlines are classified as non-plausible. Bundles reported in this figure belong to the healthy hemisphere of patients.}
  \label{fig:apss_qual_extractor}
\end{figure*}

\begin{figure*}[t]
  \centering
  \vspace{-1cm}
  \includegraphics[height=\textheight]{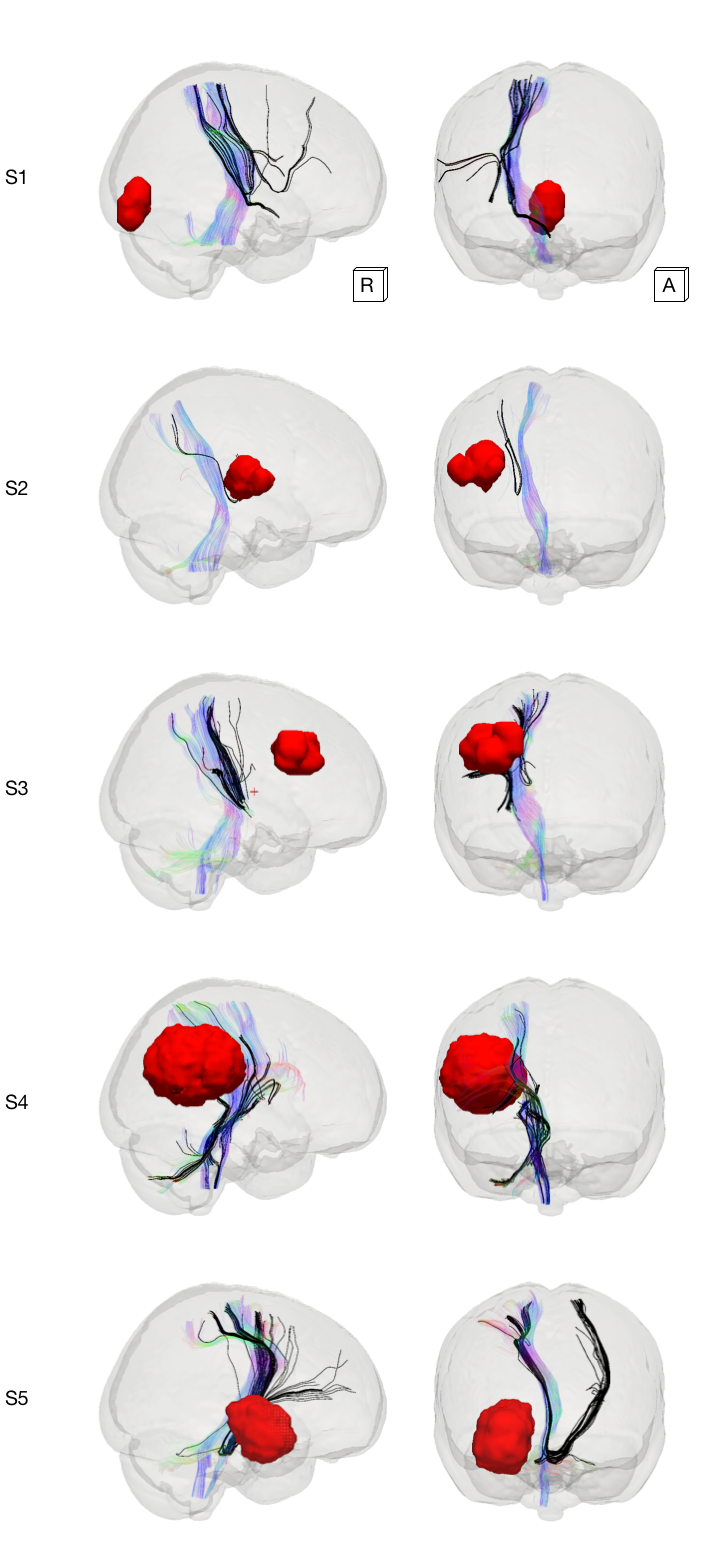}
  \vspace{-0.5cm}
  \caption[Qualitative results on clinical results. Tumored hemisphere]{Qualitative results of the tumored hemisphere after filtering APSS clinical dataset using \vfep. The bundle shown is the pyramidal tract (PT). Black streamlines are classified as non-plausible. The red ROI is the segmented tumor.}
  \label{fig:apss_tumored_qual_extractor}
\end{figure*}

\outline{results apss + extractor\\}
To illustrate the results of the qualitative analysis we report two figures. Figure~\ref{fig:apss_qual_extractor} depicts the filtering of bundles segmented from the healthy hemisphere of the subjects, while Figure \ref{fig:apss_tumored_qual_extractor} shows one bundle segmented from the tumored hemisphere. In both cases, the individuated non-plausible streamlines, colored in black, appear to be clearly artifactual and poorly compliant with the expected shape of the bundle. In many cases, such streamlines either do a strict U-turn like in the AF of subject S2 or are truncated like in the PT of S3.  

\subsection{Code and reproducibility}
\label{ssec:t_filt_code}
Verifyber has been implemented using PyTorch~\citep{paszke_pytorch_2019} and the extension for geometric deep learning, PyG~\citep{fey_fast_2019}. Our implementation, along with the trained models used in the experiments, is available at \url{https://github.com/FBK-NILab/verifyber}. The Github repository also contains our FINTA~\citep{legarreta_filtering_2021} implementation. In addition, to simplify the deployment of \vfep and \vfiz, we published an App on the BrainLife platform~\citep{avesani_open_2019}, accessible at \url{https://doi.org/10.25663/brainlife.app.390}.

%%% Local Variables: 
%%% mode: latex
%%% TeX-master: "tractogram_filtering_main"
%%% End:

%\input{results}
\section{Discussion and Conclusions}%
\label{sec:discussion}

\subsection{Verifyber Performance}

\outline{CLAIM 1: VF effective for tractogram filtering\\}
\paragraph{Verifyber is effective for tractogram filtering} The proposed method, Verifyber, resulted successful for the tractogram filtering task. It can learn from an external supervision, e.g. Extractor labeling (see Figure \ref{fig:training_ext}), and then preserving the high performance reached during training also for inference on unseen data, see Table \ref{tab:res_extractor}. In all the reported metrics the high scores are coupled with a low standard deviation in the cross-validation setting, e.g., accuracy $95.2\%\pm 0.01$ and DSC $96.6\%\pm 0.01$, an empirical evidence of the robustness with respect to the training and testing partition. The higher recall compared to the precision measure suggests that the model operate a more conservative filtering, in agreement with the bias of the {\em exclusive} labelling policy. Additionally, VF revealed fast inference at test time, classifying 1M streamlines in less than a minute (46.2 sec using a GPU NVIDIA Titan XP 12GB), a two order of magnitude less than more common filtering approaches, as signal-based and rule-based methods.

\outline{CLAIM 2: Overcome state of the art GDL\\}
\paragraph{Sequence edge convolution benefits} Performing tractogram filtering using a state of the art GDL model, e.g., PN or DGCNN, allows the use of raw streamline representation, i.e., varying number of points without forcing an arbitrary orientation, at the cost to be invariant to the permutation of streamline points. In VF we overcome such a drawback as confirmed by results of ablation study reported in Table \ref{tab:res_permutation} while remaining invariant to the fiber orientation. Furthermore, we show that considering the edges of the streamlines matters and enables VF to achieve a competitive advantage over PN when streamlines are longer and curved. In these groups of streamlines, VF has only a small drop in accuracy while significantly outperforming PN: $89\%$ vs. $76\%$ when long and very curved, and $88\%$ vs. $71\%$ when long and middle curved (see Figure \ref{fig:acc_fp_pn_sdec}). This gap can be explained by the different property of representation learning in the two models. Sequence edge convolutional layers of VF is more effective to capture the long range spatial information into the learned embedding of fibers. VF successfully encodes the most salient point in complex pathways while PN struggles to identify the salient points because the learned global embedding is not informative for local patterns (see Figure~\ref{fig:max_activations}).

\outline{CLAIM 3: Error analysis\\}
\paragraph{Error analysis} In {\em exclusive} labeling policies, the false positive error (FP) is more relevant than the false negative. Observing the FP analysis in Figure \ref{fig:acc_fp_pn_sdec}, VF disclosed robustness despite the uneven distribution of the plausible and non-plausible labels, as observable in Figure \ref{fig:streams_categories}. Compared to PN, which seems to be influenced by the higher number of plausible streamlines, the FP rate of VF is always lower or equal. Again, the difference is greater for curve and long streamlines where considering the edges helps the classification. In the false negative analysis, the presence of artifactual fibers (see Figure \ref{fig:fn_sdec}) labelled as plausible proves to some extent the ability of the model to generalize the rules beyond the labeling. Although some noise in the annotation process that brings to overestimate the plausibility of some streamlines, the large number of streamlines given in input to the training (around 10M) provides a good generalization ability of VF. We may remark the importance of this property as manual labeling is intrinsically error-prone and always leads to having some noisy labels. 

%\outline{CLAIM 4: Generalization across dataset\\}
%\todo{- successful on unseen dataset same ground truth (Table ?) 
%- remark different scan and tracking}

\outline{CLAIM 4: Open Learning\\}
\paragraph{Incremental learning} The lack of a complete WM knowledge and the constant effort of neuroanatomy community lead to a continuous evolution of the notion of anatomical plausibility, which reflects on tractogram labelings, a phenomenon known as {\em concept drift}. New bundles are added to the definition of plausibility as soon as the neuroanatomist consolidates their definitions. We emulated such a real-world scenario in the incremental learning experiments, where VF obtained convincing results, (see Table \ref{tab:res_zhang}). The iterative addition of new categories of bundles, e.g., adding projection bundles to association bundles, has not affected the performance of VF. VF learns a meaningful latent space where fibers of a bundle are grouped, and similar bundles are close (see Figure \ref{fig:tsne_emb}). This result suggests that VF effectively captures the trajectory of fibers. Therefore, we may expect better robustness when an inclusive policy of labelling is incrementally adopted in a real world setting. Moreover, in the same experiment, we unveiled the possibility to learn the filtering by training VF with a single averaged tractogram. Single-subject training works because our method scales with the number of streamlines and not with the number of subjects. Note that a single-subject training might lead to low inter-individual generalization, but this is unlikely to be the case in HCP-IZ as the tractogram already contains information of 100 subjects.
% The obtained results outperform significantly, circa 6 percent points, the reference result of deepWMA~\citep{zhang_deep_2020} taken from their paper. We sustain that the small differences in the training setting of deepWMA and VF i.e., different random split of 80/20 train/test sets and multiclass classification instead of binary classification, do not invalid the comparison. Despite a multiclass task like bundle segmentation could be tackled by our method, it is not our intention to show performance on such a task as the primary interest of this study is to distinguish plausible and non-plausible streamlines. % The use of an average of (100) tractograms rather than a single subject tractogram, is more representative of the data distribution.

\outline{CLAIM 5: Supervised approach more flexible to different plausibility definitions}
\paragraph{Supervised vs. unsupervised} One of our working hypotheses is the use of a supervised deep learning approach, in contrast with unsupervised, to guarantee higher flexibility to different definitions of anatomical plausibility. The comparison with FINTA, chosen as representative of unsupervised methods, in Table~\ref{tab:res_finta} highlights a significant difference between the two types of approaches. The performance of Verifyber is high and stable on both the {\em inclusive} and {\em exclusive} labeling policies, i.e., only $2$\%p of accuracy and $1$\%p of DSC difference. On the contrary, FINTA is not robust to the two labeling policies, achieving $14$\%p of accuracy and $7$\%p of DSC difference. We remark that such a difference occurs although we tuned the filtering threshold of FINTA specifically for each dataset. In our experience, adopting the same threshold across different datasets highly worsen the results.        
% - DL supervised vs unsupervised
% - unsupervised not robust across labeling
% - Supervised/Verifyber robust if labeling change/evolve
% - drop of FINTA even though label-specific fine tuning parameters (threshold)

An additional observation concerning FINTA is that it performs considerably better on the inclusive labeling such as HCP-IZ dataset. This labeling policy discriminates between plausible and non-plausible using the concept of bundle or cluster. Non-plausible fibers are either unknown bundle or outliers manually removed from clusters. Both the situations are favorable for solutions such as FINTA based on autoencoders, as they are known to perform well in the task of anomaly/outlier detection~\citep{pang_deep_2021}. The low dimensional bottleneck of the autoencoder combined with the reconstruction loss acts as a regularizer of the latent space and promotes a {\em denoising} of the input fiber trajectory~\citep{vincent_stacked_2010}. As a result, in FINTA, the loss based on fiber reconstruction is effective for detecting non-plausible fibers, considered outliers. However, the denoising action, especially in cases of local geometrical distortion of the pathway (see Appendix), lead to producing false positives during filtering, as highlighted by lower precision {\em w.r.t.} recall in Table~\ref{tab:res_finta}.  
               
% - FINTA robust with Inclusive but drop with exclusion
% - in inclusive non-plausible = outliers
% - loss FINTA based on fiber reconstruction (noise reduction)  is effective to detect outliers 
% - according to literature autoencoder are effective for anomaly detection

When we move from inclusive to exclusive labelings, the hypothesis of non-plausible as outliers bundles no more holds. In exclusive labelings, the anatomical plausibility concerns the artifactual geometry of fibers and the anatomical regions in which fibers pass through or terminate. However, such characteristics might be altered by the denoising effect of AEs (see Appendix), e.g., a non-plausible sharp turn within a fiber is deleted by denoising. The excessive denoising may explain the drop of performance of FINTA on HCP-EP, $14$\%p lower accuracy. In general, we may state that unsupervised filtering approaches are accurate when the premises of their loss holds in the labeling at hand. Instead, supervised approaches like Verifyber are more flexible to different labeling policies as the classification loss directly optimizes the labeling at hand. For this reason, VF substantially outperforms FINTA, +$20$\%p accuracy and +$13$\%p DSC on the exclusive labeling.

% - loss Verifyber more robust when hypo non-plausible=outliers doesn’t hold, i.e. exclusion
% - recap: labeling can hold different hypo, loss supervised more flexible
% - unsupervised good if premises on loss hold on labeling of interest

\begin{figure*}[t]
  \centering
  \includegraphics[width=.835\textwidth]{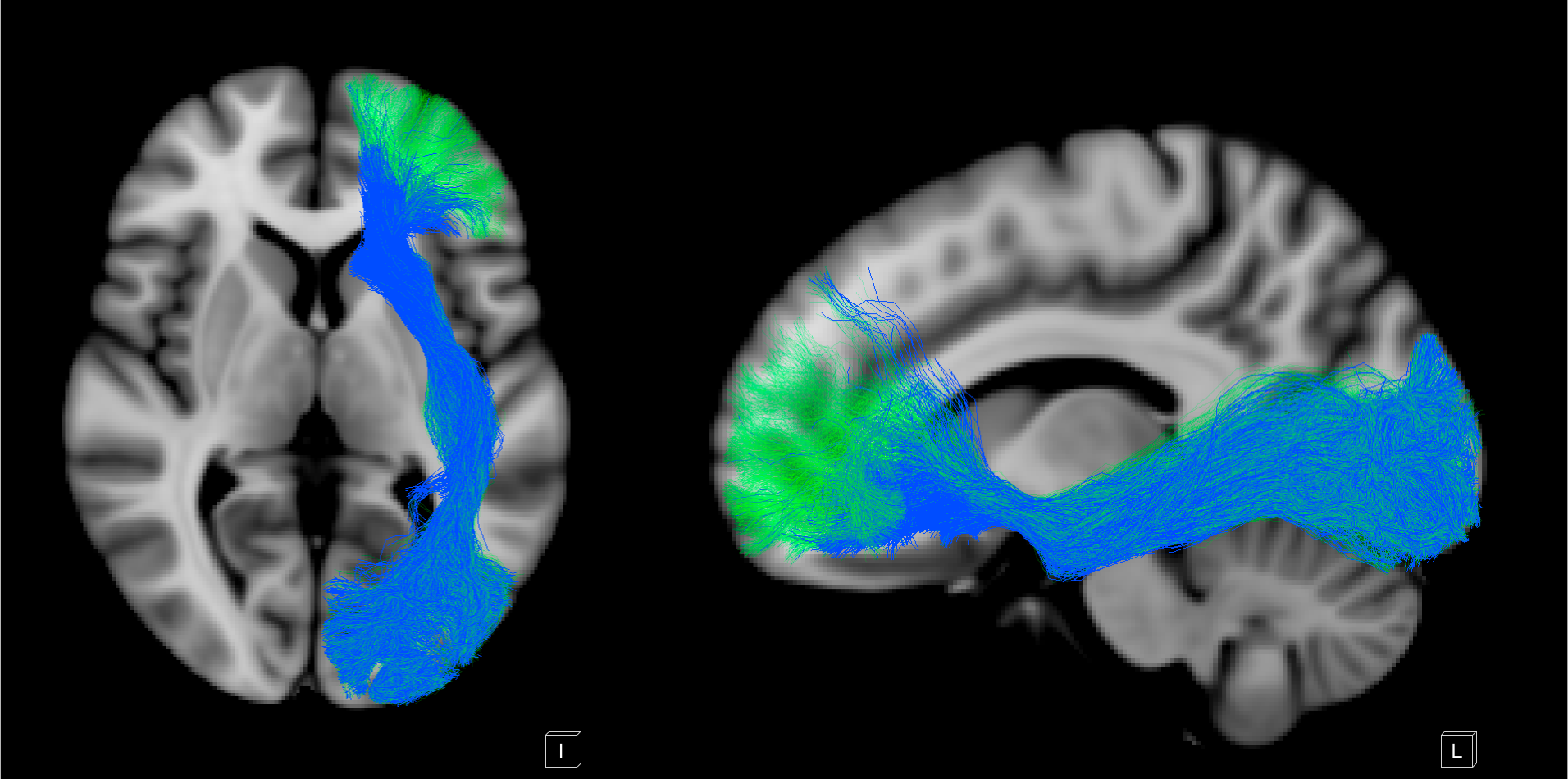}
  \caption[Example of IFOF filtered with \vfiz]{Filtering of Tractseg IFOF using \vfiz. Blue streamlines are classified as non-plausible, while transparent green are classified as plausible. }
  \label{fig:sdec_zhang_ifof}
\end{figure*}

\subsection{Model Deployment}
\outline{CLAIM 6: Model deployment\\}
\paragraph{Generalization across data sources} The reproducibility analysis aims to assess the behavior of a model on different datasets with respect to those used for training. For this purpose, we investigated the inference of VF on HCP-IW dataset, a large collection of high quality data. We considered both models trained with the two labelling policies, \vfep and \vfiz, {\em exclusive} and {\em inclusive} respectively. It is worthwhile to remark that for HCP-IW we do not have a labeling for anatomical plausibility but only the segmentation of major bundles, which we use as a proxy for plausible fibers. From the results reported in Figure~\ref{fig:ts_dsc_sdec_ext} and~\ref{fig:ts_dsc_sdec_zhang}, we notice that in both cases there is no meaningful alteration of volumetric bundle masks before and after the filtering despite the removal of around $20\%$ of the streamlines. In some specific bundles such as the IFOF, our method detected a higher percentage of non-plausible streamlines, around $50-60\%$. However, given the absence of a ground truth labeling to refer to, such high percentages of non-plausible fibers could be an error of our model. For this reason, we needed to operate an additional visual check to assess the source of the error, as illustrated in Figure~\ref{fig:sdec_zhang_ifof}. The assessment revealed a premature termination of the fiber in the frontal lobe. This anomaly is probably due to the lack of anatomical constraints (ACT) in the tracking procedure that ~\citet{wasserthal_tractseg_2018} adopted for the tracking of IFOF. For all the other bundles, an expert visually investigated a portion of the predicted non-plausible fibers confirming that such fibers were having a non-plausible pathway, see Figure~\ref{fig:ts_qual_extractor} and~\ref{fig:ts_qual_zhang}. However, the partial visual check is not enough to guarantee a $100\%$ correct filtering --- which we do not expect ---, but the preserved bundles masks reveal a conservative filtering of VF.  Testing the learnt models, \vfep and \vfiz, on the same third party test dataset allows us to estimate the labeling policies overlaps computing the agreement of the predictions. The result is $84\%$ of DSC of plausible streamlines and $37\%$ of DSC of non-plausible streamlines. The low overlap of non-plausible is explainable by the different approaches of \citet{zhang_anatomically_2018} and Extractor, i.e., inclusive vs. exclusive.

\outline{CLAIM 7: Suitable for clinic data\\}
\paragraph{Clinical application} The last validation of VF on clinical data, i.e., APSS dataset, shows that the method might be effective  even in the context of clinical data. Although the completely different data quality, tracking pipeline, and the presence of alteration caused by tumors, our method filters streamlines that after visual inspection confirmed to be truly non-plausible, see \ref{fig:apss_qual_extractor} and \ref{fig:apss_tumored_qual_extractor}. The characteristics of the detected non-plausible streamlines agree with the principles of some of the Extractor rules, e.g., truncation, loop and strict U-turn. In most cases, it is straightforward to notice the disagreement of filtered streamlines compared to the retained streamlines of the respective bundles. One could argue that a similar strong disagreement may happen when plausible streamlines belonging to other bundles are wrongly present in the APSS dataset segmentations. To answer, we can assert that the inspection shows, on the contrary, that such streamlines are truly non-plausible. Given the clinical circumstances, we may claim that filtering such non-plausible streamlines could simplify and speed up clinicians' manual process of bundle extraction. With fewer false positives, they do not have to draw several ROIs of exclusion.

\subsection{Possible weaknesses and limitation}
\paragraph{Biases of  labelling policies} Some cautions must be taken into account when deploying Verifyber to new data, particularly considering the choice of \vfep versus \vfiz. In the case of \vfep, the filtering is highly dependent on a proper non-linear co-registration of the target brain to standard MNI as many rules of Extractor concern the anatomy of that standard. The same strict requirement is not present for \vfiz, as the HCP-IZ training dataset contains fibers that are only linearly registered to a shared anatomical reference. However, \vfiz has other limitations. First, it is biased towards the bundle definitions followed by \citeauthor{zhang_anatomically_2018}. Secondly, the averaged tractogram of the atlas does not contain spurious streamlines that terminate in the WM or contain loops, and thus one cannot expect the removal of these in a new tractogram. 

\paragraph{Verifyber is not designed for signal-based filtering} In case one is interested in re-training VF using new labelings, it is important to remind that VF is not meant to learn signal-based filtering. The reason is that signal-based criteria are based on spatial regularization of fibers, i.e., fiber density map, and not on fiber anatomy. For example, considering SIFT2~\citep{smith_sift2_2015} weights, the anatomy of high-weight fibers is often non-plausible. For the sake of completeness, in additional experiments, we explored using VF to learn the SIFT2 filtering (see Appendix).

\paragraph{Quantitative generalization experiments are missing} Despite the extensive set of experiments carried out to test Verifyber, we miss a second labeled dataset where to quantitatively confirm the generalization performance of VF. The availability of labeled tractograms is still limited, and in our case, we had available only a single additional non-HCP tractogram labeled with Extractor. Despite that tractogram contains more than 1.5M fibers, we considered a single-subject test not statistically significant. We attach the result on that dataset in Appendix.

%%% Local Variables: 
%%% mode: latex
%%% TeX-master: "tractogram_filtering_main"
%%% End:
%\input{code}
\section*{CRediT}

%term list: \url{https://www.elsevier.com/authors/policies-and-guidelines/credit-author-statement}
\noindent
\textbf{Pietro Astolfi:} Conceptualization, Methodology, Software, Validation, Formal analysis, Investigation, Data Curation, Writing - Original Draft, Writing - Review \& Editing, Visualization, Project administration.
\textbf{Ruben Verhagen:} Software, Formal analysis, Investigation, Writing - Review \& Editing.
\textbf{Laurent Petit:} Resources, Validation, Data Curation, Writing - Review \& Editing.
\textbf{Emanuele Olivetti:} Software, Writing - Review \& Editing.
\textbf{Silvio Sarubbo:} Resources, Data Curation, Writing - Review \& Editing.
\textbf{Jonathan Masci:} Methodology, Writing - Review \& Editing.
\textbf{Davide Boscaini:} Conceptualization, Methodology, Writing - Review \& Editing, Visualization.
\textbf{Paolo Avesani:} Conceptualization, Methodology, Writing - Original Draft, Writing - Review \& Editing, Visualization, Supervision, Project administration, Funding acquisition.

\section*{Acknowledgements}
This work was partially supported by the grant PAT Reg. n. 764/2021 NeuSurPlan.
Finally, we gratefully acknowledge the support of NVIDIA Corporation with the donation of the Titan Xp GPU used for this research.

%\section*{References}
%%Harvard
{\footnotesize
\bibliographystyle{model2-names.bst}\biboptions{authoryear}
\bibliography{mia_2021_filtering}

\begin{thebibliography}{100}
\expandafter\ifx\csname natexlab\endcsname\relax\def\natexlab#1{#1}\fi
\providecommand{\url}[1]{\texttt{#1}}
\providecommand{\href}[2]{#2}
\providecommand{\path}[1]{#1}
\providecommand{\DOIprefix}{doi:}
\providecommand{\ArXivprefix}{arXiv:}
\providecommand{\URLprefix}{URL: }
\providecommand{\Pubmedprefix}{pmid:}
\providecommand{\doi}[1]{\href{http://dx.doi.org/#1}{\path{#1}}}
\providecommand{\Pubmed}[1]{\href{pmid:#1}{\path{#1}}}
\providecommand{\bibinfo}[2]{#2}
\ifx\xfnm\relax \def\xfnm[#1]{\unskip,\space#1}\fi
%Type = Inproceedings
\bibitem[{Astolfi et~al.(2020)Astolfi, Verhagen, Petit, Olivetti, Masci,
  Boscaini and Avesani}]{astolfi_tractogram_2020}
\bibinfo{author}{Astolfi, P.}, \bibinfo{author}{Verhagen, R.},
  \bibinfo{author}{Petit, L.}, \bibinfo{author}{Olivetti, E.},
  \bibinfo{author}{Masci, J.}, \bibinfo{author}{Boscaini, D.},
  \bibinfo{author}{Avesani, P.}, \bibinfo{year}{2020}.
\newblock \bibinfo{title}{Tractogram {{Filtering}} of {{Anatomically
  Non-plausible Fibers}} with {{Geometric Deep Learning}}}, in:
  \bibinfo{booktitle}{Medical {{Image Computing}} and {{Computer Assisted
  Intervention}} \textendash{} {{MICCAI}} 2020}, pp. \bibinfo{pages}{291--301}.
\newblock \DOIprefix\doi{10.1007/978-3-030-59728-3_29}.
%Type = Article
\bibitem[{Avesani et~al.(2019)Avesani, McPherson, Hayashi, Caiafa, Henschel,
  Garyfallidis, Kitchell, Bullock, Patterson, Olivetti, Sporns, Saykin, Wang,
  Dinov, Hancock, Caron, Qian and Pestilli}]{avesani_open_2019}
\bibinfo{author}{Avesani, P.}, \bibinfo{author}{McPherson, B.},
  \bibinfo{author}{Hayashi, S.}, \bibinfo{author}{Caiafa, C.F.},
  \bibinfo{author}{Henschel, R.}, \bibinfo{author}{Garyfallidis, E.},
  \bibinfo{author}{Kitchell, L.}, \bibinfo{author}{Bullock, D.},
  \bibinfo{author}{Patterson, A.}, \bibinfo{author}{Olivetti, E.},
  \bibinfo{author}{Sporns, O.}, \bibinfo{author}{Saykin, A.J.},
  \bibinfo{author}{Wang, L.}, \bibinfo{author}{Dinov, I.},
  \bibinfo{author}{Hancock, D.}, \bibinfo{author}{Caron, B.},
  \bibinfo{author}{Qian, Y.}, \bibinfo{author}{Pestilli, F.},
  \bibinfo{year}{2019}.
\newblock \bibinfo{title}{The open diffusion data derivatives, brain data
  upcycling via integrated publishing of derivatives and reproducible open
  cloud services}.
\newblock \bibinfo{journal}{Scientific Data} \bibinfo{volume}{6},
  \bibinfo{pages}{1--13}.
\newblock \DOIprefix\doi{10.1038/s41597-019-0073-y}.
%Type = Inproceedings
\bibitem[{Aydogan and Shi(2015)}]{aydogan_track_2015}
\bibinfo{author}{Aydogan, D.B.}, \bibinfo{author}{Shi, Y.},
  \bibinfo{year}{2015}.
\newblock \bibinfo{title}{Track {{Filtering}} via {{Iterative Correction}} of
  {{TDI Topology}}}, in: \bibinfo{booktitle}{Medical {{Image Computing}} and
  {{Computer-Assisted Intervention}} -- {{MICCAI}} 2015}, pp.
  \bibinfo{pages}{20--27}.
\newblock \DOIprefix\doi{10.1007/978-3-319-24553-9_3}.
%Type = Article
\bibitem[{Aydogan and Shi(2018)}]{aydogan_tracking_2018}
\bibinfo{author}{Aydogan, D.B.}, \bibinfo{author}{Shi, Y.},
  \bibinfo{year}{2018}.
\newblock \bibinfo{title}{Tracking and validation techniques for
  topographically organized tractography}.
\newblock \bibinfo{journal}{NeuroImage} \bibinfo{volume}{181},
  \bibinfo{pages}{64--84}.
\newblock \DOIprefix\doi{10.1016/j.neuroimage.2018.06.071}.
%Type = Article
\bibitem[{Basser et~al.(2000)Basser, Pajevic, Pierpaoli, Duda and
  Aldroubi}]{basser_vivo_2000}
\bibinfo{author}{Basser, P.J.}, \bibinfo{author}{Pajevic, S.},
  \bibinfo{author}{Pierpaoli, C.}, \bibinfo{author}{Duda, J.},
  \bibinfo{author}{Aldroubi, A.}, \bibinfo{year}{2000}.
\newblock \bibinfo{title}{In vivo fiber tractography using {{DT-MRI}} data}.
\newblock \bibinfo{journal}{Magnetic Resonance in Medicine}
  \bibinfo{volume}{44}, \bibinfo{pages}{625--632}.
%Type = Article
\bibitem[{Bastiani et~al.(2012)Bastiani, Shah, Goebel and
  Roebroeck}]{bastiani_human_2012}
\bibinfo{author}{Bastiani, M.}, \bibinfo{author}{Shah, N.J.},
  \bibinfo{author}{Goebel, R.}, \bibinfo{author}{Roebroeck, A.},
  \bibinfo{year}{2012}.
\newblock \bibinfo{title}{Human cortical connectome reconstruction from
  diffusion weighted {{MRI}}: The effect of tractography algorithm}.
\newblock \bibinfo{journal}{NeuroImage} \bibinfo{volume}{62},
  \bibinfo{pages}{1732--1749}.
%Type = Article
\bibitem[{Bert{\`o} et~al.(2021)Bert{\`o}, Bullock, Astolfi, Hayashi, Zigiotto,
  Annicchiarico, Corsini, De~Benedictis, Sarubbo, Pestilli, Avesani and
  Olivetti}]{berto_classifyber_2021}
\bibinfo{author}{Bert{\`o}, G.}, \bibinfo{author}{Bullock, D.},
  \bibinfo{author}{Astolfi, P.}, \bibinfo{author}{Hayashi, S.},
  \bibinfo{author}{Zigiotto, L.}, \bibinfo{author}{Annicchiarico, L.},
  \bibinfo{author}{Corsini, F.}, \bibinfo{author}{De~Benedictis, A.},
  \bibinfo{author}{Sarubbo, S.}, \bibinfo{author}{Pestilli, F.},
  \bibinfo{author}{Avesani, P.}, \bibinfo{author}{Olivetti, E.},
  \bibinfo{year}{2021}.
\newblock \bibinfo{title}{Classifyber, a robust streamline-based linear
  classifier for white matter bundle segmentation}.
\newblock \bibinfo{journal}{NeuroImage} \bibinfo{volume}{224},
  \bibinfo{pages}{117402}.
\newblock \DOIprefix\doi{10.1016/j.neuroimage.2020.117402}.
%Type = Article
\bibitem[{Bronstein et~al.(2017)Bronstein, Bruna, LeCun, Szlam and
  Vandergheynst}]{bronstein_geometric_2017}
\bibinfo{author}{Bronstein, M.M.}, \bibinfo{author}{Bruna, J.},
  \bibinfo{author}{LeCun, Y.}, \bibinfo{author}{Szlam, A.},
  \bibinfo{author}{Vandergheynst, P.}, \bibinfo{year}{2017}.
\newblock \bibinfo{title}{Geometric deep learning: Going beyond euclidean
  data}.
\newblock \bibinfo{journal}{IEEE Signal Processing Magazine}
  \bibinfo{volume}{34}, \bibinfo{pages}{18--42}.
%Type = Article
\bibitem[{Buchanan et~al.(2014)Buchanan, Pernet, Gorgolewski, Storkey and
  Bastin}]{buchanan_testretest_2014}
\bibinfo{author}{Buchanan, C.R.}, \bibinfo{author}{Pernet, C.R.},
  \bibinfo{author}{Gorgolewski, K.J.}, \bibinfo{author}{Storkey, A.J.},
  \bibinfo{author}{Bastin, M.E.}, \bibinfo{year}{2014}.
\newblock \bibinfo{title}{Test\textendash retest reliability of structural
  brain networks from diffusion {{MRI}}}.
\newblock \bibinfo{journal}{NeuroImage} \bibinfo{volume}{86},
  \bibinfo{pages}{231--243}.
\newblock \DOIprefix\doi{10.1016/j.neuroimage.2013.09.054}.
%Type = Article
\bibitem[{Chamberland et~al.(2017)Chamberland, Scherrer, Prabhu, Madsen,
  Fortin, Whittingstall, Descoteaux and Warfield}]{chamberland_active_2017}
\bibinfo{author}{Chamberland, M.}, \bibinfo{author}{Scherrer, B.},
  \bibinfo{author}{Prabhu, S.P.}, \bibinfo{author}{Madsen, J.},
  \bibinfo{author}{Fortin, D.}, \bibinfo{author}{Whittingstall, K.},
  \bibinfo{author}{Descoteaux, M.}, \bibinfo{author}{Warfield, S.K.},
  \bibinfo{year}{2017}.
\newblock \bibinfo{title}{Active delineation of {{Meyer}}'s loop using oriented
  priors through {{MAGNEtic}} tractography ({{MAGNET}})}.
\newblock \bibinfo{journal}{Human Brain Mapping} \bibinfo{volume}{38},
  \bibinfo{pages}{509--527}.
%Type = Inproceedings
\bibitem[{Chandio et~al.(2022)Chandio, Chattopadhyay, {Owens-Walton}, Reina,
  Nabulsi, Thomopoulos, Garyfallidis and Thompson}]{chandio_fiberneat_2022}
\bibinfo{author}{Chandio, B.Q.}, \bibinfo{author}{Chattopadhyay, T.},
  \bibinfo{author}{{Owens-Walton}, C.}, \bibinfo{author}{Reina, J.E.V.},
  \bibinfo{author}{Nabulsi, L.}, \bibinfo{author}{Thomopoulos, S.I.},
  \bibinfo{author}{Garyfallidis, E.}, \bibinfo{author}{Thompson, P.M.},
  \bibinfo{year}{2022}.
\newblock \bibinfo{title}{{{FiberNeat}}: {{Unsupervised White Matter Tract
  Filtering}}}, in: \bibinfo{booktitle}{2022 44th {{Annual International
  Conference}} of the {{IEEE Engineering}} in {{Medicine}} {$\&$} {{Biology
  Society}} ({{EMBC}})}, pp. \bibinfo{pages}{5055--5061}.
\newblock \DOIprefix\doi{10.1109/EMBC48229.2022.9870877}.
%Type = Article
\bibitem[{C{\^o}t{\'e} et~al.(2015)C{\^o}t{\'e}, Garyfallidis, Larochelle and
  Descoteaux}]{cote_cleaning_2015}
\bibinfo{author}{C{\^o}t{\'e}, M.A.}, \bibinfo{author}{Garyfallidis, E.},
  \bibinfo{author}{Larochelle, H.}, \bibinfo{author}{Descoteaux, M.},
  \bibinfo{year}{2015}.
\newblock \bibinfo{title}{Cleaning up the mess: Tractography outlier removal
  using hierarchical {{QuickBundles}} clustering}.
\newblock \bibinfo{journal}{Proceedings of International Society of Magnetic
  Resonance in Medicine (ISMRM)} .
%Type = Article
\bibitem[{C{\^o}t{\'e} et~al.(2013)C{\^o}t{\'e}, Girard, Bor{\'e},
  Garyfallidis, Houde and Descoteaux}]{cote_tractometer_2013}
\bibinfo{author}{C{\^o}t{\'e}, M.A.}, \bibinfo{author}{Girard, G.},
  \bibinfo{author}{Bor{\'e}, A.}, \bibinfo{author}{Garyfallidis, E.},
  \bibinfo{author}{Houde, J.C.}, \bibinfo{author}{Descoteaux, M.},
  \bibinfo{year}{2013}.
\newblock \bibinfo{title}{Tractometer: {{Towards}} validation of tractography
  pipelines}.
\newblock \bibinfo{journal}{Medical Image Analysis} \bibinfo{volume}{17},
  \bibinfo{pages}{844--857}.
\newblock \DOIprefix\doi{10.1016/j.media.2013.03.009}.
%Type = Article
\bibitem[{Daducci et~al.(2015)Daducci, Dal~Palu, Lemkaddem and
  Thiran}]{daducci_commit_2015}
\bibinfo{author}{Daducci, A.}, \bibinfo{author}{Dal~Palu, A.},
  \bibinfo{author}{Lemkaddem, A.}, \bibinfo{author}{Thiran, J.P.},
  \bibinfo{year}{2015}.
\newblock \bibinfo{title}{{{COMMIT}}: {{Convex Optimization Modeling}} for
  {{Microstructure Informed Tractography}}}.
\newblock \bibinfo{journal}{IEEE Transactions on Medical Imaging}
  \bibinfo{volume}{34}, \bibinfo{pages}{246--257}.
\newblock \DOIprefix\doi{10.1109/TMI.2014.2352414}.
%Type = Article
\bibitem[{De~Benedictis et~al.(2016)De~Benedictis, Petit, Descoteaux, Marras,
  Barbareschi, Corsini, Dallabona, Chioffi and
  Sarubbo}]{de_benedictis_new_2016}
\bibinfo{author}{De~Benedictis, A.}, \bibinfo{author}{Petit, L.},
  \bibinfo{author}{Descoteaux, M.}, \bibinfo{author}{Marras, C.E.},
  \bibinfo{author}{Barbareschi, M.}, \bibinfo{author}{Corsini, F.},
  \bibinfo{author}{Dallabona, M.}, \bibinfo{author}{Chioffi, F.},
  \bibinfo{author}{Sarubbo, S.}, \bibinfo{year}{2016}.
\newblock \bibinfo{title}{New insights in the homotopic and heterotopic
  connectivity of the frontal portion of the human corpus callosum revealed by
  microdissection and diffusion tractography}.
\newblock \bibinfo{journal}{Human Brain Mapping} \bibinfo{volume}{37},
  \bibinfo{pages}{4718--4735}.
\newblock \DOIprefix\doi{10.1002/hbm.23339}.
%Type = Incollection
\bibitem[{Descoteaux(2015)}]{descoteaux_high_2015}
\bibinfo{author}{Descoteaux, M.}, \bibinfo{year}{2015}.
\newblock \bibinfo{title}{High {{Angular Resolution Diffusion Imaging}}
  ({{HARDI}})}, in: \bibinfo{booktitle}{Wiley {{Encyclopedia}} of
  {{Electrical}} and {{Electronics Engineering}}}, pp. \bibinfo{pages}{1--25}.
\newblock \DOIprefix\doi{10.1002/047134608X.W8258}.
%Type = Article
\bibitem[{Descoteaux et~al.(2007)Descoteaux, Angelino, Fitzgibbons and
  Deriche}]{descoteaux_regularized_2007}
\bibinfo{author}{Descoteaux, M.}, \bibinfo{author}{Angelino, E.},
  \bibinfo{author}{Fitzgibbons, S.}, \bibinfo{author}{Deriche, R.},
  \bibinfo{year}{2007}.
\newblock \bibinfo{title}{Regularized, fast, and robust analytical {{Q-ball}}
  imaging}.
\newblock \bibinfo{journal}{Magnetic Resonance in Medicine}
  \bibinfo{volume}{58}, \bibinfo{pages}{497--510}.
\newblock \DOIprefix\doi{10.1002/mrm.21277}.
%Type = Article
\bibitem[{Essayed et~al.(2017)Essayed, Zhang, Unadkat, Cosgrove, Golby and
  O'Donnell}]{essayed_white_2017}
\bibinfo{author}{Essayed, W.I.}, \bibinfo{author}{Zhang, F.},
  \bibinfo{author}{Unadkat, P.}, \bibinfo{author}{Cosgrove, G.R.},
  \bibinfo{author}{Golby, A.J.}, \bibinfo{author}{O'Donnell, L.J.},
  \bibinfo{year}{2017}.
\newblock \bibinfo{title}{White matter tractography for neurosurgical planning:
  {{A}} topography-based review of the current state of the art}.
\newblock \bibinfo{journal}{NeuroImage: Clinical} \bibinfo{volume}{15},
  \bibinfo{pages}{659--672}.
\newblock \DOIprefix\doi{10.1016/j.nicl.2017.06.011}.
%Type = Inproceedings
\bibitem[{Fey and Lenssen(2019)}]{fey_fast_2019}
\bibinfo{author}{Fey, M.}, \bibinfo{author}{Lenssen, J.E.},
  \bibinfo{year}{2019}.
\newblock \bibinfo{title}{Fast {{Graph Representation Learning}} with {{PyTorch
  Geometric}}}, in: \bibinfo{booktitle}{{{ICLR Workshop}} on {{Representation
  Learning}} on {{Graphs}} and {{Manifolds}}}.
%Type = Article
\bibitem[{Fillard et~al.(2011)Fillard, Descoteaux, Goh, Gouttard, Jeurissen,
  Malcolm, {Ramirez-Manzanares}, Reisert, Sakaie, Tensaouti, Yo, Mangin and
  Poupon}]{fillard_quantitative_2011}
\bibinfo{author}{Fillard, P.}, \bibinfo{author}{Descoteaux, M.},
  \bibinfo{author}{Goh, A.}, \bibinfo{author}{Gouttard, S.},
  \bibinfo{author}{Jeurissen, B.}, \bibinfo{author}{Malcolm, J.},
  \bibinfo{author}{{Ramirez-Manzanares}, A.}, \bibinfo{author}{Reisert, M.},
  \bibinfo{author}{Sakaie, K.}, \bibinfo{author}{Tensaouti, F.},
  \bibinfo{author}{Yo, T.}, \bibinfo{author}{Mangin, J.F.},
  \bibinfo{author}{Poupon, C.}, \bibinfo{year}{2011}.
\newblock \bibinfo{title}{Quantitative evaluation of 10 tractography algorithms
  on a realistic diffusion {{MR}} phantom}.
\newblock \bibinfo{journal}{NeuroImage} \bibinfo{volume}{56},
  \bibinfo{pages}{220--234}.
\newblock \DOIprefix\doi{10.1016/j.neuroimage.2011.01.032}.
%Type = Article
\bibitem[{Fischl(2012)}]{fischl_freesurfer_2012}
\bibinfo{author}{Fischl, B.}, \bibinfo{year}{2012}.
\newblock \bibinfo{title}{{{FreeSurfer}}}.
\newblock \bibinfo{journal}{NeuroImage} \bibinfo{volume}{62},
  \bibinfo{pages}{774--781}.
\newblock \DOIprefix\doi{10.1016/j.neuroimage.2012.01.021}.
%Type = Article
\bibitem[{Fonov et~al.(2011)Fonov, Evans, Botteron, Almli, McKinstry, Collins
  and {Brain Development Cooperative Group}}]{fonov_unbiased_2011}
\bibinfo{author}{Fonov, V.}, \bibinfo{author}{Evans, A.C.},
  \bibinfo{author}{Botteron, K.}, \bibinfo{author}{Almli, C.R.},
  \bibinfo{author}{McKinstry, R.C.}, \bibinfo{author}{Collins, D.L.},
  \bibinfo{author}{{Brain Development Cooperative Group}},
  \bibinfo{year}{2011}.
\newblock \bibinfo{title}{Unbiased average age-appropriate atlases for
  pediatric studies}.
\newblock \bibinfo{journal}{NeuroImage} \bibinfo{volume}{54},
  \bibinfo{pages}{313--327}.
\newblock \DOIprefix\doi{10.1016/j.neuroimage.2010.07.033}.
%Type = Article
\bibitem[{Frigo et~al.(2020)Frigo, {Deslauriers-Gauthier}, Parker, Aziz
  Ould~Ismail, John~Kim, Verma and Deriche}]{frigo_diffusion_2020}
\bibinfo{author}{Frigo, M.}, \bibinfo{author}{{Deslauriers-Gauthier}, S.},
  \bibinfo{author}{Parker, D.}, \bibinfo{author}{Aziz Ould~Ismail, A.},
  \bibinfo{author}{John~Kim, J.}, \bibinfo{author}{Verma, R.},
  \bibinfo{author}{Deriche, R.}, \bibinfo{year}{2020}.
\newblock \bibinfo{title}{Diffusion {{MRI}} tractography filtering techniques
  change the topology of structural connectomes}.
\newblock \bibinfo{journal}{Journal of Neural Engineering}
  \bibinfo{volume}{17}, \bibinfo{pages}{065002}.
\newblock \DOIprefix\doi{10.1088/1741-2552/abc29b}.
%Type = Article
\bibitem[{Garyfallidis et~al.(2014)Garyfallidis, Brett, Amirbekian, Rokem, {van
  der Walt}, Descoteaux, {Nimmo-Smith} and
  Contributors}]{garyfallidis_dipy_2014}
\bibinfo{author}{Garyfallidis, E.}, \bibinfo{author}{Brett, M.},
  \bibinfo{author}{Amirbekian, B.}, \bibinfo{author}{Rokem, A.},
  \bibinfo{author}{{van der Walt}, S.}, \bibinfo{author}{Descoteaux, M.},
  \bibinfo{author}{{Nimmo-Smith}, I.}, \bibinfo{author}{Contributors, D.},
  \bibinfo{year}{2014}.
\newblock \bibinfo{title}{Dipy, a library for the analysis of diffusion {{MRI}}
  data}.
\newblock \bibinfo{journal}{Frontiers in Neuroinformatics} \bibinfo{volume}{8},
  \bibinfo{pages}{1+}.
%Type = Article
\bibitem[{Garyfallidis et~al.(2012)Garyfallidis, Brett, Correia, Williams and
  {Nimmo-Smith}}]{garyfallidis_quickbundles_2012}
\bibinfo{author}{Garyfallidis, E.}, \bibinfo{author}{Brett, M.},
  \bibinfo{author}{Correia, M.M.}, \bibinfo{author}{Williams, G.B.},
  \bibinfo{author}{{Nimmo-Smith}, I.}, \bibinfo{year}{2012}.
\newblock \bibinfo{title}{{{QuickBundles}}, a {{Method}} for {{Tractography
  Simplification}}.}
\newblock \bibinfo{journal}{Frontiers in Neuroscience} \bibinfo{volume}{6}.
\newblock \DOIprefix\doi{10.3389/fnins.2012.00175}.
%Type = Article
\bibitem[{Garyfallidis et~al.(2018)Garyfallidis, C{\^o}t{\'e}, Rheault, Sidhu,
  Hau, Petit, Fortin, Cunanne and Descoteaux}]{garyfallidis_recognition_2018}
\bibinfo{author}{Garyfallidis, E.}, \bibinfo{author}{C{\^o}t{\'e}, M.A.},
  \bibinfo{author}{Rheault, F.}, \bibinfo{author}{Sidhu, J.},
  \bibinfo{author}{Hau, J.}, \bibinfo{author}{Petit, L.},
  \bibinfo{author}{Fortin, D.}, \bibinfo{author}{Cunanne, S.},
  \bibinfo{author}{Descoteaux, M.}, \bibinfo{year}{2018}.
\newblock \bibinfo{title}{Recognition of white matter bundles using local and
  global streamline-based registration and clustering}.
\newblock \bibinfo{journal}{NeuroImage} \bibinfo{volume}{170},
  \bibinfo{pages}{283--295}.
\newblock \DOIprefix\doi{10.1016/j.neuroimage.2017.07.015}.
%Type = Article
\bibitem[{Girard et~al.(2020)Girard, Caminiti, {Battaglia-Mayer}, {St-Onge},
  Ambrosen, Eskildsen, Krug, Dyrby, Descoteaux, Thiran and
  Innocenti}]{girard_cortical_2020}
\bibinfo{author}{Girard, G.}, \bibinfo{author}{Caminiti, R.},
  \bibinfo{author}{{Battaglia-Mayer}, A.}, \bibinfo{author}{{St-Onge}, E.},
  \bibinfo{author}{Ambrosen, K.S.}, \bibinfo{author}{Eskildsen, S.F.},
  \bibinfo{author}{Krug, K.}, \bibinfo{author}{Dyrby, T.B.},
  \bibinfo{author}{Descoteaux, M.}, \bibinfo{author}{Thiran, J.P.},
  \bibinfo{author}{Innocenti, G.M.}, \bibinfo{year}{2020}.
\newblock \bibinfo{title}{On the cortical connectivity in the macaque brain:
  {{A}} comparison of diffusion tractography and histological tracing data}.
\newblock \bibinfo{journal}{NeuroImage} \bibinfo{volume}{221},
  \bibinfo{pages}{117201}.
\newblock \DOIprefix\doi{10.1016/j.neuroimage.2020.117201}.
%Type = Article
\bibitem[{Girard et~al.(2014)Girard, Whittingstall, Deriche and
  Descoteaux}]{girard_towards_2014}
\bibinfo{author}{Girard, G.}, \bibinfo{author}{Whittingstall, K.},
  \bibinfo{author}{Deriche, R.}, \bibinfo{author}{Descoteaux, M.},
  \bibinfo{year}{2014}.
\newblock \bibinfo{title}{Towards quantitative connectivity analysis: Reducing
  tractography biases}.
\newblock \bibinfo{journal}{NeuroImage} \bibinfo{volume}{98},
  \bibinfo{pages}{266--278}.
\newblock \DOIprefix\doi{10.1016/j.neuroimage.2014.04.074}.
%Type = Article
\bibitem[{Glasser et~al.(2013)Glasser, Sotiropoulos, Wilson, Coalson, Fischl,
  Andersson, Xu, Jbabdi, Webster, Polimeni, Van~Essen and
  Jenkinson}]{glasser_minimal_2013}
\bibinfo{author}{Glasser, M.F.}, \bibinfo{author}{Sotiropoulos, S.N.},
  \bibinfo{author}{Wilson, J.A.}, \bibinfo{author}{Coalson, T.S.},
  \bibinfo{author}{Fischl, B.}, \bibinfo{author}{Andersson, J.L.},
  \bibinfo{author}{Xu, J.}, \bibinfo{author}{Jbabdi, S.},
  \bibinfo{author}{Webster, M.}, \bibinfo{author}{Polimeni, J.R.},
  \bibinfo{author}{Van~Essen, D.C.}, \bibinfo{author}{Jenkinson, M.},
  \bibinfo{year}{2013}.
\newblock \bibinfo{title}{The minimal preprocessing pipelines for the {{Human
  Connectome Project}}}.
\newblock \bibinfo{journal}{NeuroImage} \bibinfo{volume}{80},
  \bibinfo{pages}{105--124}.
\newblock \DOIprefix\doi{10.1016/j.neuroimage.2013.04.127}.
%Type = Article
\bibitem[{Graves and Schmidhuber(2005)}]{graves_framewise_2005}
\bibinfo{author}{Graves, A.}, \bibinfo{author}{Schmidhuber, J.},
  \bibinfo{year}{2005}.
\newblock \bibinfo{title}{Framewise phoneme classification with bidirectional
  {{LSTM}} and other neural network architectures}.
\newblock \bibinfo{journal}{Neural Networks} \bibinfo{volume}{18},
  \bibinfo{pages}{602--610}.
%Type = Inproceedings
\bibitem[{Gupta et~al.(2018)Gupta, Thomopoulos, Corbin, Rashid and
  Thompson}]{gupta_fibernet_2018}
\bibinfo{author}{Gupta, V.}, \bibinfo{author}{Thomopoulos, S.I.},
  \bibinfo{author}{Corbin, C.K.}, \bibinfo{author}{Rashid, F.},
  \bibinfo{author}{Thompson, P.M.}, \bibinfo{year}{2018}.
\newblock \bibinfo{title}{{{FIBERNET}} 2.0: {{An}} automatic neural network
  based tool for clustering white matter fibers in the brain}, in:
  \bibinfo{booktitle}{2018 {{IEEE}} 15th {{International Symposium}} on
  {{Biomedical Imaging}} ({{ISBI}} 2018)}, pp. \bibinfo{pages}{708--711}.
\newblock \DOIprefix\doi{10.1109/ISBI.2018.8363672}.
%Type = Inproceedings
\bibitem[{Gupta et~al.(2017)Gupta, Thomopoulos, Rashid and
  Thompson}]{gupta_fibernet_2017}
\bibinfo{author}{Gupta, V.}, \bibinfo{author}{Thomopoulos, S.I.},
  \bibinfo{author}{Rashid, F.M.}, \bibinfo{author}{Thompson, P.M.},
  \bibinfo{year}{2017}.
\newblock \bibinfo{title}{{{FiberNET}}: {{An Ensemble Deep Learning Framework}}
  for {{Clustering White Matter Fibers}}}, in: \bibinfo{booktitle}{Medical
  {{Image Computing}} and {{Computer Assisted Intervention}} {$-$} {{MICCAI}}
  2017}, pp. \bibinfo{pages}{548--555}.
\newblock \DOIprefix\doi{10.1007/978-3-319-66182-7_63}.
%Type = Article
\bibitem[{Hau et~al.(2017)Hau, Sarubbo, Houde, Corsini, Girard, Deledalle,
  Crivello, Zago, Mellet, Jobard, Joliot, Mazoyer, {Tzourio-Mazoyer},
  Descoteaux and Petit}]{hau_revisiting_2017}
\bibinfo{author}{Hau, J.}, \bibinfo{author}{Sarubbo, S.},
  \bibinfo{author}{Houde, J.}, \bibinfo{author}{Corsini, F.},
  \bibinfo{author}{Girard, G.}, \bibinfo{author}{Deledalle, C.},
  \bibinfo{author}{Crivello, F.}, \bibinfo{author}{Zago, L.},
  \bibinfo{author}{Mellet, E.}, \bibinfo{author}{Jobard, G.},
  \bibinfo{author}{Joliot, M.}, \bibinfo{author}{Mazoyer, B.},
  \bibinfo{author}{{Tzourio-Mazoyer}, N.}, \bibinfo{author}{Descoteaux, M.},
  \bibinfo{author}{Petit, L.}, \bibinfo{year}{2017}.
\newblock \bibinfo{title}{Revisiting the human uncinate fasciculus, its
  subcomponents and asymmetries with stem-based tractography and
  microdissection validation}.
\newblock \bibinfo{journal}{Brain Structure and Function} ,
  \bibinfo{pages}{1--18}\DOIprefix\doi{10.1007/s00429-016-1298-6}.
%Type = Inproceedings
\bibitem[{He et~al.(2016)He, Zhang, Ren and Sun}]{he_deep_2016}
\bibinfo{author}{He, K.}, \bibinfo{author}{Zhang, X.}, \bibinfo{author}{Ren,
  S.}, \bibinfo{author}{Sun, J.}, \bibinfo{year}{2016}.
\newblock \bibinfo{title}{Deep residual learning for image recognition}, in:
  \bibinfo{booktitle}{Proceedings of the {{IEEE Conference}} on {{Computer
  Vision}} and {{Pattern Recognition}}}, pp. \bibinfo{pages}{770--778}.
%Type = Article
\bibitem[{Henderson et~al.(2020)Henderson, Abdullah, Verma and
  Brem}]{henderson_tractography_2020}
\bibinfo{author}{Henderson, F.}, \bibinfo{author}{Abdullah, K.G.},
  \bibinfo{author}{Verma, R.}, \bibinfo{author}{Brem, S.},
  \bibinfo{year}{2020}.
\newblock \bibinfo{title}{Tractography and the connectome in neurosurgical
  treatment of gliomas: The premise, the progress, and the potential}.
\newblock \bibinfo{journal}{Neurosurgical Focus} \bibinfo{volume}{48},
  \bibinfo{pages}{E6}.
\newblock \DOIprefix\doi{10.3171/2019.11.FOCUS19785}.
%Type = Article
\bibitem[{Hochreiter and Schmidhuber(1997)}]{hochreiter_long_1997}
\bibinfo{author}{Hochreiter, S.}, \bibinfo{author}{Schmidhuber, J.},
  \bibinfo{year}{1997}.
\newblock \bibinfo{title}{Long short-term memory}.
\newblock \bibinfo{journal}{Neural Computation} \bibinfo{volume}{9},
  \bibinfo{pages}{1735--1780}.
%Type = Article
\bibitem[{Huang et~al.(2015)Huang, Xu and Yu}]{huang_bidirectional_2015}
\bibinfo{author}{Huang, Z.}, \bibinfo{author}{Xu, W.}, \bibinfo{author}{Yu,
  K.}, \bibinfo{year}{2015}.
\newblock \bibinfo{title}{Bidirectional {{LSTM-CRF}} models for sequence
  tagging}.
\newblock \bibinfo{journal}{arXiv preprint arXiv:1508.01991}
  \href{http://arxiv.org/abs/1508.01991}{\tt arXiv:1508.01991}.
%Type = Article
\bibitem[{Jenkinson et~al.(2012)Jenkinson, Beckmann, Behrens, Woolrich and
  Smith}]{jenkinson_fsl_2012}
\bibinfo{author}{Jenkinson, M.}, \bibinfo{author}{Beckmann, C.F.},
  \bibinfo{author}{Behrens, T.E.J.}, \bibinfo{author}{Woolrich, M.W.},
  \bibinfo{author}{Smith, S.M.}, \bibinfo{year}{2012}.
\newblock \bibinfo{title}{{{FSL}}}.
\newblock \bibinfo{journal}{NeuroImage} \bibinfo{volume}{62},
  \bibinfo{pages}{782--790}.
\newblock \DOIprefix\doi{http://dx.doi.org/10.1016/j.neuroimage.2011.09.015}.
%Type = Article
\bibitem[{Jeurissen et~al.(2017)Jeurissen, Descoteaux, Mori and
  Leemans}]{jeurissen_diffusion_2017}
\bibinfo{author}{Jeurissen, B.}, \bibinfo{author}{Descoteaux, M.},
  \bibinfo{author}{Mori, S.}, \bibinfo{author}{Leemans, A.},
  \bibinfo{year}{2017}.
\newblock \bibinfo{title}{Diffusion {{MRI}} fiber tractography of the brain}.
\newblock \bibinfo{journal}{NMR in Biomedicine} \bibinfo{volume}{32},
  \bibinfo{pages}{e3785}.
\newblock \DOIprefix\doi{10.1002/nbm.3785}.
%Type = Inproceedings
\bibitem[{J{\"o}rgens et~al.(2021)J{\"o}rgens, Descoteaux and
  Moreno}]{jorgens_challenges_2021}
\bibinfo{author}{J{\"o}rgens, D.}, \bibinfo{author}{Descoteaux, M.},
  \bibinfo{author}{Moreno, R.}, \bibinfo{year}{2021}.
\newblock \bibinfo{title}{Challenges for {{Tractogram Filtering}}}, in:
  \bibinfo{booktitle}{Anisotropy {{Across Fields}} and {{Scales}}}, pp.
  \bibinfo{pages}{149--168}.
\newblock \DOIprefix\doi{10.1007/978-3-030-56215-1_7}.
%Type = Article
\bibitem[{Krizhevsky et~al.(2012)Krizhevsky, Sutskever and
  Hinton}]{krizhevsky_imagenet_2012}
\bibinfo{author}{Krizhevsky, A.}, \bibinfo{author}{Sutskever, I.},
  \bibinfo{author}{Hinton, G.E.}, \bibinfo{year}{2012}.
\newblock \bibinfo{title}{Imagenet classification with deep convolutional
  neural networks}.
\newblock \bibinfo{journal}{Advances in Neural Information Processing Systems}
  \bibinfo{volume}{25}, \bibinfo{pages}{1097--1105}.
%Type = Article
\bibitem[{Lee et~al.(2020)Lee, O'Hara, Sonoda, Kuroda, Juhasz, Asano, Dong and
  Jeong}]{lee_novel_2020}
\bibinfo{author}{Lee, M.H.}, \bibinfo{author}{O'Hara, N.},
  \bibinfo{author}{Sonoda, M.}, \bibinfo{author}{Kuroda, N.},
  \bibinfo{author}{Juhasz, C.}, \bibinfo{author}{Asano, E.},
  \bibinfo{author}{Dong, M.}, \bibinfo{author}{Jeong, J.W.},
  \bibinfo{year}{2020}.
\newblock \bibinfo{title}{Novel {{Deep Learning Network Analysis}} of
  {{Electrical Stimulation Mapping-Driven Diffusion MRI Tractography}} to
  {{Improve Preoperative Evaluation}} of {{Pediatric Epilepsy}}}.
\newblock \bibinfo{journal}{IEEE Transactions on Biomedical Engineering}
  \bibinfo{volume}{67}, \bibinfo{pages}{3151--3162}.
\newblock \DOIprefix\doi{10.1109/TBME.2020.2977531}.
%Type = Article
\bibitem[{Legarreta et~al.(2021)Legarreta, Petit, Rheault, Theaud, Lemaire,
  Descoteaux and Jodoin}]{legarreta_filtering_2021}
\bibinfo{author}{Legarreta, J.H.}, \bibinfo{author}{Petit, L.},
  \bibinfo{author}{Rheault, F.}, \bibinfo{author}{Theaud, G.},
  \bibinfo{author}{Lemaire, C.}, \bibinfo{author}{Descoteaux, M.},
  \bibinfo{author}{Jodoin, P.M.}, \bibinfo{year}{2021}.
\newblock \bibinfo{title}{Filtering in {{Tractography}} using {{Autoencoders}}
  ({{FINTA}})}.
\newblock \bibinfo{journal}{Medical Image Analysis} ,
  \bibinfo{pages}{102126}\DOIprefix\doi{10.1016/j.media.2021.102126}.
%Type = Article
\bibitem[{Lemkaddem et~al.(2014)Lemkaddem, Ski{\"o}ldebrand, Dal~Pal{\'u},
  Thiran and Daducci}]{lemkaddem_global_2014}
\bibinfo{author}{Lemkaddem, A.}, \bibinfo{author}{Ski{\"o}ldebrand, D.},
  \bibinfo{author}{Dal~Pal{\'u}, A.}, \bibinfo{author}{Thiran, J.P.},
  \bibinfo{author}{Daducci, A.}, \bibinfo{year}{2014}.
\newblock \bibinfo{title}{Global tractography with embedded anatomical priors
  for quantitative connectivity analysis}.
\newblock \bibinfo{journal}{Frontiers in Neurology} \bibinfo{volume}{5},
  \bibinfo{pages}{232}.
\newblock \DOIprefix\doi{10.3389/fneur.2014.00232}.
%Type = Article
\bibitem[{Maffei et~al.(2018)Maffei, Jovicich, De~Benedictis, Corsini,
  Barbareschi, Chioffi and Sarubbo}]{maffei_topography_2018}
\bibinfo{author}{Maffei, C.}, \bibinfo{author}{Jovicich, J.},
  \bibinfo{author}{De~Benedictis, A.}, \bibinfo{author}{Corsini, F.},
  \bibinfo{author}{Barbareschi, M.}, \bibinfo{author}{Chioffi, F.},
  \bibinfo{author}{Sarubbo, S.}, \bibinfo{year}{2018}.
\newblock \bibinfo{title}{Topography of the human acoustic radiation as
  revealed by ex vivo fibers micro-dissection and in vivo diffusion-based
  tractography}.
\newblock \bibinfo{journal}{Brain Structure and Function}
  \bibinfo{volume}{223}, \bibinfo{pages}{449--459}.
\newblock \DOIprefix\doi{10.1007/s00429-017-1471-6}.
%Type = Article
\bibitem[{{Maier-Hein} et~al.(2017){Maier-Hein}, Neher, Houde, C{\^o}t{\'e},
  Garyfallidis, Zhong, Chamberland, Yeh, Lin, Ji, Reddick, Glass, Chen, Feng,
  Gao, Wu, Ma, He, Li, Westin, {Deslauriers-Gauthier}, Gonz{\'a}lez, Paquette,
  {St-Jean}, Girard, Rheault, Sidhu, Tax, Guo, Mesri, D{\'a}vid, Froeling,
  Heemskerk, Leemans, Bor{\'e}, Pinsard, Bedetti, Desrosiers, Brambati, Doyon,
  Sarica, Vasta, Cerasa, Quattrone, Yeatman, Khan, Hodges, Alexander,
  Romascano, Barakovic, Aur{\'i}a, Esteban, Lemkaddem, Thiran, Cetingul, Odry,
  Mailhe, Nadar, Pizzagalli, Prasad, {Villalon-Reina}, Galvis, Thompson,
  Requejo, Laguna, Lacerda, Barrett, Dell'Acqua, Catani, Petit, Caruyer,
  Daducci, Dyrby, {Holland-Letz}, Hilgetag, Stieltjes and
  Descoteaux}]{maier-hein_challenge_2017}
\bibinfo{author}{{Maier-Hein}, K.H.}, \bibinfo{author}{Neher, P.F.},
  \bibinfo{author}{Houde, J.C.}, \bibinfo{author}{C{\^o}t{\'e}, M.A.},
  \bibinfo{author}{Garyfallidis, E.}, \bibinfo{author}{Zhong, J.},
  \bibinfo{author}{Chamberland, M.}, \bibinfo{author}{Yeh, F.C.},
  \bibinfo{author}{Lin, Y.C.}, \bibinfo{author}{Ji, Q.},
  \bibinfo{author}{Reddick, W.E.}, \bibinfo{author}{Glass, J.O.},
  \bibinfo{author}{Chen, D.Q.}, \bibinfo{author}{Feng, Y.},
  \bibinfo{author}{Gao, C.}, \bibinfo{author}{Wu, Y.}, \bibinfo{author}{Ma,
  J.}, \bibinfo{author}{He, R.}, \bibinfo{author}{Li, Q.},
  \bibinfo{author}{Westin, C.F.}, \bibinfo{author}{{Deslauriers-Gauthier}, S.},
  \bibinfo{author}{Gonz{\'a}lez, J.O.O.}, \bibinfo{author}{Paquette, M.},
  \bibinfo{author}{{St-Jean}, S.}, \bibinfo{author}{Girard, G.},
  \bibinfo{author}{Rheault, F.}, \bibinfo{author}{Sidhu, J.},
  \bibinfo{author}{Tax, C.M.W.}, \bibinfo{author}{Guo, F.},
  \bibinfo{author}{Mesri, H.Y.}, \bibinfo{author}{D{\'a}vid, S.},
  \bibinfo{author}{Froeling, M.}, \bibinfo{author}{Heemskerk, A.M.},
  \bibinfo{author}{Leemans, A.}, \bibinfo{author}{Bor{\'e}, A.},
  \bibinfo{author}{Pinsard, B.}, \bibinfo{author}{Bedetti, C.},
  \bibinfo{author}{Desrosiers, M.}, \bibinfo{author}{Brambati, S.},
  \bibinfo{author}{Doyon, J.}, \bibinfo{author}{Sarica, A.},
  \bibinfo{author}{Vasta, R.}, \bibinfo{author}{Cerasa, A.},
  \bibinfo{author}{Quattrone, A.}, \bibinfo{author}{Yeatman, J.},
  \bibinfo{author}{Khan, A.R.}, \bibinfo{author}{Hodges, W.},
  \bibinfo{author}{Alexander, S.}, \bibinfo{author}{Romascano, D.},
  \bibinfo{author}{Barakovic, M.}, \bibinfo{author}{Aur{\'i}a, A.},
  \bibinfo{author}{Esteban, O.}, \bibinfo{author}{Lemkaddem, A.},
  \bibinfo{author}{Thiran, J.P.}, \bibinfo{author}{Cetingul, H.E.},
  \bibinfo{author}{Odry, B.L.}, \bibinfo{author}{Mailhe, B.},
  \bibinfo{author}{Nadar, M.S.}, \bibinfo{author}{Pizzagalli, F.},
  \bibinfo{author}{Prasad, G.}, \bibinfo{author}{{Villalon-Reina}, J.E.},
  \bibinfo{author}{Galvis, J.}, \bibinfo{author}{Thompson, P.M.},
  \bibinfo{author}{Requejo, F.D.S.}, \bibinfo{author}{Laguna, P.L.},
  \bibinfo{author}{Lacerda, L.M.}, \bibinfo{author}{Barrett, R.},
  \bibinfo{author}{Dell'Acqua, F.}, \bibinfo{author}{Catani, M.},
  \bibinfo{author}{Petit, L.}, \bibinfo{author}{Caruyer, E.},
  \bibinfo{author}{Daducci, A.}, \bibinfo{author}{Dyrby, T.B.},
  \bibinfo{author}{{Holland-Letz}, T.}, \bibinfo{author}{Hilgetag, C.C.},
  \bibinfo{author}{Stieltjes, B.}, \bibinfo{author}{Descoteaux, M.},
  \bibinfo{year}{2017}.
\newblock \bibinfo{title}{The challenge of mapping the human connectome based
  on diffusion tractography}.
\newblock \bibinfo{journal}{Nature Communications} \bibinfo{volume}{8},
  \bibinfo{pages}{1349}.
\newblock \DOIprefix\doi{10.1038/s41467-017-01285-x}.
%Type = Inproceedings
\bibitem[{Masci et~al.(2016)Masci, Rodol{\`a}, Boscaini, Bronstein and
  Li}]{masci_geometric_2016}
\bibinfo{author}{Masci, J.}, \bibinfo{author}{Rodol{\`a}, E.},
  \bibinfo{author}{Boscaini, D.}, \bibinfo{author}{Bronstein, M.M.},
  \bibinfo{author}{Li, H.}, \bibinfo{year}{2016}.
\newblock \bibinfo{title}{Geometric deep learning}, in:
  \bibinfo{booktitle}{{{SIGGRAPH ASIA}} 2016 {{Courses}}}, pp.
  \bibinfo{pages}{1--50}.
\newblock \DOIprefix\doi{10.1145/2988458.2988485}.
%Type = Article
\bibitem[{Mazoyer et~al.(2016)Mazoyer, Mellet, Perchey, Zago, Crivello, Jobard,
  Delcroix, Vigneau, Leroux, Petit, Joliot and
  {Tzourio-Mazoyer}}]{mazoyer_bilgin_2016}
\bibinfo{author}{Mazoyer, B.}, \bibinfo{author}{Mellet, E.},
  \bibinfo{author}{Perchey, G.}, \bibinfo{author}{Zago, L.},
  \bibinfo{author}{Crivello, F.}, \bibinfo{author}{Jobard, G.},
  \bibinfo{author}{Delcroix, N.}, \bibinfo{author}{Vigneau, M.},
  \bibinfo{author}{Leroux, G.}, \bibinfo{author}{Petit, L.},
  \bibinfo{author}{Joliot, M.}, \bibinfo{author}{{Tzourio-Mazoyer}, N.},
  \bibinfo{year}{2016}.
\newblock \bibinfo{title}{{{BIL}}{$\&$}{{GIN}}: {{A}} neuroimaging, cognitive,
  behavioral, and genetic database for the study of human brain
  lateralization}.
\newblock \bibinfo{journal}{NeuroImage} \bibinfo{volume}{124},
  \bibinfo{pages}{1225--1231}.
\newblock \DOIprefix\doi{10.1016/j.neuroimage.2015.02.071}.
%Type = Article
\bibitem[{Mori et~al.(1999)Mori, Crain, Chacko and
  Van~Zijl}]{mori_three-dimensional_1999}
\bibinfo{author}{Mori, S.}, \bibinfo{author}{Crain, B.J.},
  \bibinfo{author}{Chacko, V.P.}, \bibinfo{author}{Van~Zijl, P.},
  \bibinfo{year}{1999}.
\newblock \bibinfo{title}{Three-dimensional tracking of axonal projections in
  the brain by magnetic resonance imaging}.
\newblock \bibinfo{journal}{Annals of Neurology} \bibinfo{volume}{45},
  \bibinfo{pages}{265--269}.
\newblock
  \DOIprefix\doi{10.1002/1531-8249(199902)45:2<265::AID-ANA21>3.0.CO;2-3}.
%Type = Article
\bibitem[{Nath et~al.(2020)Nath, Schilling, Parvathaneni, Huo, Blaber,
  Hainline, Barakovic, Romascano, Rafael-Patino, Frigo, Girard, Thiran,
  Daducci, Rowe, Rodrigues, Pr{\v c}kovska, Aydogan, Sun, Shi, Parker,
  Ould~Ismail, Verma, Cabeen, Toga, Newton, Wasserthal, Neher, Maier-Hein,
  Savini, Palesi, Kaden, Wu, He, Feng, Paquette, Rheault, Sidhu, Lebel,
  Leemans, Descoteaux, Dyrby, Kang and Landman}]{nath_tractography_2020}
\bibinfo{author}{Nath, V.}, \bibinfo{author}{Schilling, K.G.},
  \bibinfo{author}{Parvathaneni, P.}, \bibinfo{author}{Huo, Y.},
  \bibinfo{author}{Blaber, J.A.}, \bibinfo{author}{Hainline, A.E.},
  \bibinfo{author}{Barakovic, M.}, \bibinfo{author}{Romascano, D.},
  \bibinfo{author}{Rafael-Patino, J.}, \bibinfo{author}{Frigo, M.},
  \bibinfo{author}{Girard, G.}, \bibinfo{author}{Thiran, J.P.},
  \bibinfo{author}{Daducci, A.}, \bibinfo{author}{Rowe, M.},
  \bibinfo{author}{Rodrigues, P.}, \bibinfo{author}{Pr{\v c}kovska, V.},
  \bibinfo{author}{Aydogan, D.B.}, \bibinfo{author}{Sun, W.},
  \bibinfo{author}{Shi, Y.}, \bibinfo{author}{Parker, W.A.},
  \bibinfo{author}{Ould~Ismail, A.A.}, \bibinfo{author}{Verma, R.},
  \bibinfo{author}{Cabeen, R.P.}, \bibinfo{author}{Toga, A.W.},
  \bibinfo{author}{Newton, A.T.}, \bibinfo{author}{Wasserthal, J.},
  \bibinfo{author}{Neher, P.}, \bibinfo{author}{Maier-Hein, K.},
  \bibinfo{author}{Savini, G.}, \bibinfo{author}{Palesi, F.},
  \bibinfo{author}{Kaden, E.}, \bibinfo{author}{Wu, Y.}, \bibinfo{author}{He,
  J.}, \bibinfo{author}{Feng, Y.}, \bibinfo{author}{Paquette, M.},
  \bibinfo{author}{Rheault, F.}, \bibinfo{author}{Sidhu, J.},
  \bibinfo{author}{Lebel, C.}, \bibinfo{author}{Leemans, A.},
  \bibinfo{author}{Descoteaux, M.}, \bibinfo{author}{Dyrby, T.B.},
  \bibinfo{author}{Kang, H.}, \bibinfo{author}{Landman, B.A.},
  \bibinfo{year}{2020}.
\newblock \bibinfo{title}{Tractography reproducibility challenge with empirical
  data ({{TraCED}}): {{The}} 2017 {{ISMRM}} diffusion study group challenge}.
\newblock \bibinfo{journal}{Journal of Magnetic Resonance Imaging}
  \bibinfo{volume}{51}, \bibinfo{pages}{234--249}.
\newblock \DOIprefix\doi{10.1002/jmri.26794}.
%Type = Article
\bibitem[{Neher et~al.(2015)Neher, Descoteaux, Houde, Stieltjes and
  {Maier-Hein}}]{neher_strengths_2015}
\bibinfo{author}{Neher, P.F.}, \bibinfo{author}{Descoteaux, M.},
  \bibinfo{author}{Houde, J.C.}, \bibinfo{author}{Stieltjes, B.},
  \bibinfo{author}{{Maier-Hein}, K.H.}, \bibinfo{year}{2015}.
\newblock \bibinfo{title}{Strengths and weaknesses of state of the art fiber
  tractography pipelines \textendash{} {{A}} comprehensive in-vivo and phantom
  evaluation study using {{Tractometer}}}.
\newblock \bibinfo{journal}{Medical Image Analysis} \bibinfo{volume}{26},
  \bibinfo{pages}{287--305}.
\newblock \DOIprefix\doi{10.1016/j.media.2015.10.011}.
%Type = Inproceedings
\bibitem[{Neher et~al.(2018)Neher, Stieltjes and
  {Maier-Hein}}]{neher_anchor-constrained_2018}
\bibinfo{author}{Neher, P.F.}, \bibinfo{author}{Stieltjes, B.},
  \bibinfo{author}{{Maier-Hein}, K.H.}, \bibinfo{year}{2018}.
\newblock \bibinfo{title}{Anchor-{{Constrained Plausibility}} ({{ACP}}): {{A
  Novel Concept}} for {{Assessing Tractography}} and {{Reducing
  False-Positives}}}, in: \bibinfo{booktitle}{Medical {{Image Computing}} and
  {{Computer Assisted Intervention}} \textendash{} {{MICCAI}} 2018}, pp.
  \bibinfo{pages}{20--27}.
\newblock \DOIprefix\doi{10.1007/978-3-030-00931-1_3}.
%Type = Inproceedings
\bibitem[{Nie and Shi(2019)}]{nie_topographic_2019}
\bibinfo{author}{Nie, X.}, \bibinfo{author}{Shi, Y.}, \bibinfo{year}{2019}.
\newblock \bibinfo{title}{Topographic {{Filtering}} of {{Tractograms}} as
  {{Vector Field Flows}}}, in: \bibinfo{booktitle}{Medical {{Image Computing}}
  and {{Computer Assisted Intervention}} \textendash{} {{MICCAI}} 2019}, pp.
  \bibinfo{pages}{564--572}.
\newblock \DOIprefix\doi{10.1007/978-3-030-32248-9_63}.
%Type = Article
\bibitem[{{Ocampo-Pineda} et~al.(2021){Ocampo-Pineda}, Schiavi, Rheault,
  Girard, Petit, Descoteaux and Daducci}]{ocampo-pineda_hierarchical_2021}
\bibinfo{author}{{Ocampo-Pineda}, M.}, \bibinfo{author}{Schiavi, S.},
  \bibinfo{author}{Rheault, F.}, \bibinfo{author}{Girard, G.},
  \bibinfo{author}{Petit, L.}, \bibinfo{author}{Descoteaux, M.},
  \bibinfo{author}{Daducci, A.}, \bibinfo{year}{2021}.
\newblock \bibinfo{title}{Hierarchical {{Microstructure Informed
  Tractography}}}.
\newblock \bibinfo{journal}{Brain Connectivity}
  \DOIprefix\doi{10.1089/brain.2020.0907}.
%Type = Inproceedings
\bibitem[{O'Donnell et~al.(2012)O'Donnell, Wells, Golby and
  Westin}]{odonnell_unbiased_2012}
\bibinfo{author}{O'Donnell, L.J.}, \bibinfo{author}{Wells, W.M.},
  \bibinfo{author}{Golby, A.J.}, \bibinfo{author}{Westin, C.F.},
  \bibinfo{year}{2012}.
\newblock \bibinfo{title}{Unbiased {{Groupwise Registration}} of {{White Matter
  Tractography}}}, in: \bibinfo{booktitle}{Medical {{Image Computing}} and
  {{Computer-Assisted Intervention}} \textendash{} {{MICCAI}} 2012}, pp.
  \bibinfo{pages}{123--130}.
\newblock \DOIprefix\doi{10.1007/978-3-642-33454-2_16}.
%Type = Article
\bibitem[{O'Donnell and Westin(2007)}]{odonnell_automatic_2007}
\bibinfo{author}{O'Donnell, L.J.}, \bibinfo{author}{Westin, C.F.},
  \bibinfo{year}{2007}.
\newblock \bibinfo{title}{Automatic {{Tractography Segmentation Using}} a
  {{High-Dimensional White Matter Atlas}}}.
\newblock \bibinfo{journal}{IEEE Transactions on Medical Imaging}
  \bibinfo{volume}{26}, \bibinfo{pages}{1562--1575}.
\newblock \DOIprefix\doi{10.1109/TMI.2007.906785}.
%Type = Inproceedings
\bibitem[{Olivetti and Avesani(2011)}]{olivetti_supervised_2011}
\bibinfo{author}{Olivetti, E.}, \bibinfo{author}{Avesani, P.},
  \bibinfo{year}{2011}.
\newblock \bibinfo{title}{Supervised segmentation of fiber tracts}, in:
  \bibinfo{booktitle}{Proceedings of the {{First}} International Conference on
  {{Similarity-based}} Pattern Recognition}, pp. \bibinfo{pages}{261--274}.
\newblock \DOIprefix\doi{10.1007/978-3-642-24471-1_19}.
%Type = Article
\bibitem[{Pang et~al.(2021)Pang, Shen, Cao and Hengel}]{pang_deep_2021}
\bibinfo{author}{Pang, G.}, \bibinfo{author}{Shen, C.}, \bibinfo{author}{Cao,
  L.}, \bibinfo{author}{Hengel, A.V.D.}, \bibinfo{year}{2021}.
\newblock \bibinfo{title}{Deep learning for anomaly detection: {{A}} review}.
\newblock \bibinfo{journal}{ACM Computing Surveys (CSUR)} \bibinfo{volume}{54},
  \bibinfo{pages}{1--38}.
%Type = Inproceedings
\bibitem[{Paszke et~al.(2019)Paszke, Gross, Massa, Lerer, Bradbury, Chanan,
  Killeen, Lin, Gimelshein and Antiga}]{paszke_pytorch_2019}
\bibinfo{author}{Paszke, A.}, \bibinfo{author}{Gross, S.},
  \bibinfo{author}{Massa, F.}, \bibinfo{author}{Lerer, A.},
  \bibinfo{author}{Bradbury, J.}, \bibinfo{author}{Chanan, G.},
  \bibinfo{author}{Killeen, T.}, \bibinfo{author}{Lin, Z.},
  \bibinfo{author}{Gimelshein, N.}, \bibinfo{author}{Antiga, L.},
  \bibinfo{year}{2019}.
\newblock \bibinfo{title}{{{PyTorch}}: {{An}} imperative style,
  high-performance deep learning library}, in: \bibinfo{booktitle}{Advances in
  {{Neural Information Processing Systems}}}, pp. \bibinfo{pages}{8024--8035}.
%Type = Article
\bibitem[{Pestilli et~al.(2014)Pestilli, Yeatman, Rokem, Kay and
  Wandell}]{pestilli_evaluation_2014}
\bibinfo{author}{Pestilli, F.}, \bibinfo{author}{Yeatman, J.D.},
  \bibinfo{author}{Rokem, A.}, \bibinfo{author}{Kay, K.N.},
  \bibinfo{author}{Wandell, B.A.}, \bibinfo{year}{2014}.
\newblock \bibinfo{title}{Evaluation and statistical inference for human
  connectomes}.
\newblock \bibinfo{journal}{Nature Methods} \bibinfo{volume}{11},
  \bibinfo{pages}{1058--1063}.
\newblock \DOIprefix\doi{10.1038/nmeth.3098}.
%Type = Article
\bibitem[{Petit et~al.(2022)Petit, Ali, Rheault, Bor{\'e}, Cremona, Corsini,
  De~Benedictis, Descoteaux and Sarubbo}]{petit_structural_2022}
\bibinfo{author}{Petit, L.}, \bibinfo{author}{Ali, K.M.},
  \bibinfo{author}{Rheault, F.}, \bibinfo{author}{Bor{\'e}, A.},
  \bibinfo{author}{Cremona, S.}, \bibinfo{author}{Corsini, F.},
  \bibinfo{author}{De~Benedictis, A.}, \bibinfo{author}{Descoteaux, M.},
  \bibinfo{author}{Sarubbo, S.}, \bibinfo{year}{2022}.
\newblock \bibinfo{title}{The structural connectivity of the human angular
  gyrus as revealed by microdissection and diffusion tractography}.
\newblock \bibinfo{journal}{Brain Structure and Function}
  \DOIprefix\doi{10.1007/s00429-022-02551-5}.
%Type = Article
\bibitem[{Pierpaoli et~al.(1996)Pierpaoli, Jezzard, Basser, Barnett and
  Di~Chiro}]{pierpaoli_diffusion_1996}
\bibinfo{author}{Pierpaoli, C.}, \bibinfo{author}{Jezzard, P.},
  \bibinfo{author}{Basser, P.J.}, \bibinfo{author}{Barnett, A.},
  \bibinfo{author}{Di~Chiro, G.}, \bibinfo{year}{1996}.
\newblock \bibinfo{title}{Diffusion tensor {{MR}} imaging of the human brain}.
\newblock \bibinfo{journal}{Radiology} \bibinfo{volume}{201},
  \bibinfo{pages}{637--648}.
\newblock \DOIprefix\doi{10.1148/radiology.201.3.8939209}.
%Type = Article
\bibitem[{Presseau et~al.(2015)Presseau, Jodoin, Houde and
  Descoteaux}]{presseau_new_2015}
\bibinfo{author}{Presseau, C.}, \bibinfo{author}{Jodoin, P.M.},
  \bibinfo{author}{Houde, J.C.}, \bibinfo{author}{Descoteaux, M.},
  \bibinfo{year}{2015}.
\newblock \bibinfo{title}{A new compression format for fiber tracking
  datasets}.
\newblock \bibinfo{journal}{NeuroImage} \bibinfo{volume}{109},
  \bibinfo{pages}{73--83}.
\newblock \DOIprefix\doi{10.1016/j.neuroimage.2014.12.058}.
%Type = Inproceedings
\bibitem[{Qi et~al.(2017)Qi, Su, Mo and Guibas}]{qi_pointnet_2017}
\bibinfo{author}{Qi, C.R.}, \bibinfo{author}{Su, H.}, \bibinfo{author}{Mo, K.},
  \bibinfo{author}{Guibas, L.J.}, \bibinfo{year}{2017}.
\newblock \bibinfo{title}{{{PointNet}}: {{Deep Learning}} on {{Point Sets}} for
  {{3D Classification}} and {{Segmentation}}}, in:
  \bibinfo{booktitle}{Proceedings of the {{IEEE Conference}} on {{Computer
  Vision}} and {{Pattern Recognition}}}, pp. \bibinfo{pages}{652--660}.
\newblock \href{http://arxiv.org/abs/1612.00593}{\tt arXiv:1612.00593}.
%Type = Article
\bibitem[{Raffelt et~al.(2012)Raffelt, Tournier, Rose, Ridgway, Henderson,
  Crozier, Salvado and Connelly}]{raffelt_apparent_2012}
\bibinfo{author}{Raffelt, D.}, \bibinfo{author}{Tournier, J.D.},
  \bibinfo{author}{Rose, S.}, \bibinfo{author}{Ridgway, G.R.},
  \bibinfo{author}{Henderson, R.}, \bibinfo{author}{Crozier, S.},
  \bibinfo{author}{Salvado, O.}, \bibinfo{author}{Connelly, A.},
  \bibinfo{year}{2012}.
\newblock \bibinfo{title}{Apparent {{Fibre Density}}: {{A}} novel measure for
  the analysis of diffusion-weighted magnetic resonance images}.
\newblock \bibinfo{journal}{NeuroImage} \bibinfo{volume}{59},
  \bibinfo{pages}{3976--3994}.
\newblock \DOIprefix\doi{10.1016/j.neuroimage.2011.10.045}.
%Type = Article
\bibitem[{Reddy and Rathi(2016)}]{reddy_joint_2016}
\bibinfo{author}{Reddy, C.P.}, \bibinfo{author}{Rathi, Y.},
  \bibinfo{year}{2016}.
\newblock \bibinfo{title}{Joint {{Multi-Fiber NODDI Parameter Estimation}} and
  {{Tractography Using}} the {{Unscented Information Filter}}}.
\newblock \bibinfo{journal}{Frontiers in Neuroscience} \bibinfo{volume}{10}.
\newblock \DOIprefix\doi{10.3389/fnins.2016.00166}.
%Type = Article
\bibitem[{Rheault et~al.(2020a)Rheault, De~Benedictis, Daducci, Maffei, Tax,
  Romascano, Caverzasi, Morency, Corrivetti, Pestilli, Girard, Theaud,
  Zemmoura, Hau, Glavin, Jordan, Pomiecko, Chamberland, Barakovic, Goyette,
  Poulin, Chenot, Panesar, Sarubbo, Petit and
  Descoteaux}]{rheault_tractostorm_2020}
\bibinfo{author}{Rheault, F.}, \bibinfo{author}{De~Benedictis, A.},
  \bibinfo{author}{Daducci, A.}, \bibinfo{author}{Maffei, C.},
  \bibinfo{author}{Tax, C.M.W.}, \bibinfo{author}{Romascano, D.},
  \bibinfo{author}{Caverzasi, E.}, \bibinfo{author}{Morency, F.C.},
  \bibinfo{author}{Corrivetti, F.}, \bibinfo{author}{Pestilli, F.},
  \bibinfo{author}{Girard, G.}, \bibinfo{author}{Theaud, G.},
  \bibinfo{author}{Zemmoura, I.}, \bibinfo{author}{Hau, J.},
  \bibinfo{author}{Glavin, K.}, \bibinfo{author}{Jordan, K.M.},
  \bibinfo{author}{Pomiecko, K.}, \bibinfo{author}{Chamberland, M.},
  \bibinfo{author}{Barakovic, M.}, \bibinfo{author}{Goyette, N.},
  \bibinfo{author}{Poulin, P.}, \bibinfo{author}{Chenot, Q.},
  \bibinfo{author}{Panesar, S.S.}, \bibinfo{author}{Sarubbo, S.},
  \bibinfo{author}{Petit, L.}, \bibinfo{author}{Descoteaux, M.},
  \bibinfo{year}{2020}a.
\newblock \bibinfo{title}{Tractostorm: {{The}} what, why, and how of
  tractography dissection reproducibility}.
\newblock \bibinfo{journal}{Human Brain Mapping} \bibinfo{volume}{41},
  \bibinfo{pages}{1859--1874}.
\newblock \DOIprefix\doi{10.1002/hbm.24917}.
%Type = Article
\bibitem[{Rheault et~al.(2020b)Rheault, Poulin, Valcourt~Caron, {St-Onge} and
  Descoteaux}]{rheault_common_2020}
\bibinfo{author}{Rheault, F.}, \bibinfo{author}{Poulin, P.},
  \bibinfo{author}{Valcourt~Caron, A.}, \bibinfo{author}{{St-Onge}, E.},
  \bibinfo{author}{Descoteaux, M.}, \bibinfo{year}{2020}b.
\newblock \bibinfo{title}{Common misconceptions, hidden biases and modern
  challenges of {{dMRI}} tractography}.
\newblock \bibinfo{journal}{Journal of Neural Engineering}
  \bibinfo{volume}{17}, \bibinfo{pages}{011001}.
\newblock \DOIprefix\doi{10.1088/1741-2552/ab6aad}.
%Type = Article
\bibitem[{Rheault et~al.(2019)Rheault, {St-Onge}, Sidhu, {Maier-Hein},
  {Tzourio-Mazoyer}, Petit and Descoteaux}]{rheault_bundle-specific_2019}
\bibinfo{author}{Rheault, F.}, \bibinfo{author}{{St-Onge}, E.},
  \bibinfo{author}{Sidhu, J.}, \bibinfo{author}{{Maier-Hein}, K.},
  \bibinfo{author}{{Tzourio-Mazoyer}, N.}, \bibinfo{author}{Petit, L.},
  \bibinfo{author}{Descoteaux, M.}, \bibinfo{year}{2019}.
\newblock \bibinfo{title}{Bundle-specific tractography with incorporated
  anatomical and orientational priors}.
\newblock \bibinfo{journal}{NeuroImage} \bibinfo{volume}{186},
  \bibinfo{pages}{382--398}.
\newblock \DOIprefix\doi{10.1016/j.neuroimage.2018.11.018}.
%Type = Article
\bibitem[{Schiavi et~al.(2020)Schiavi, {Ocampo-Pineda}, Barakovic, Petit,
  Descoteaux, Thiran and Daducci}]{schiavi_new_2020}
\bibinfo{author}{Schiavi, S.}, \bibinfo{author}{{Ocampo-Pineda}, M.},
  \bibinfo{author}{Barakovic, M.}, \bibinfo{author}{Petit, L.},
  \bibinfo{author}{Descoteaux, M.}, \bibinfo{author}{Thiran, J.P.},
  \bibinfo{author}{Daducci, A.}, \bibinfo{year}{2020}.
\newblock \bibinfo{title}{A new method for accurate in vivo mapping of human
  brain connections using microstructural and anatomical information}.
\newblock \bibinfo{journal}{Science Advances} \bibinfo{volume}{6},
  \bibinfo{pages}{eaba8245}.
\newblock \DOIprefix\doi{10.1126/sciadv.aba8245}.
%Type = Article
\bibitem[{Schilling et~al.(2019)Schilling, Nath, Hansen, Parvathaneni, Blaber,
  Gao, Neher, Aydogan, Shi, {Ocampo-Pineda}, Schiavi, Daducci, Girard,
  Barakovic, {Rafael-Patino}, Romascano, Rensonnet, Pizzolato, Bates, Fischi,
  Thiran, {Canales-Rodr{\'i}guez}, Huang, Zhu, Zhong, Cabeen, Toga, Rheault,
  Theaud, Houde, Sidhu, Chamberland, Westin, Dyrby, Verma, Rathi, Irfanoglu,
  Thomas, Pierpaoli, Descoteaux, Anderson and Landman}]{schilling_limits_2019}
\bibinfo{author}{Schilling, K.G.}, \bibinfo{author}{Nath, V.},
  \bibinfo{author}{Hansen, C.}, \bibinfo{author}{Parvathaneni, P.},
  \bibinfo{author}{Blaber, J.}, \bibinfo{author}{Gao, Y.},
  \bibinfo{author}{Neher, P.}, \bibinfo{author}{Aydogan, D.B.},
  \bibinfo{author}{Shi, Y.}, \bibinfo{author}{{Ocampo-Pineda}, M.},
  \bibinfo{author}{Schiavi, S.}, \bibinfo{author}{Daducci, A.},
  \bibinfo{author}{Girard, G.}, \bibinfo{author}{Barakovic, M.},
  \bibinfo{author}{{Rafael-Patino}, J.}, \bibinfo{author}{Romascano, D.},
  \bibinfo{author}{Rensonnet, G.}, \bibinfo{author}{Pizzolato, M.},
  \bibinfo{author}{Bates, A.}, \bibinfo{author}{Fischi, E.},
  \bibinfo{author}{Thiran, J.P.}, \bibinfo{author}{{Canales-Rodr{\'i}guez},
  E.J.}, \bibinfo{author}{Huang, C.}, \bibinfo{author}{Zhu, H.},
  \bibinfo{author}{Zhong, L.}, \bibinfo{author}{Cabeen, R.},
  \bibinfo{author}{Toga, A.W.}, \bibinfo{author}{Rheault, F.},
  \bibinfo{author}{Theaud, G.}, \bibinfo{author}{Houde, J.C.},
  \bibinfo{author}{Sidhu, J.}, \bibinfo{author}{Chamberland, M.},
  \bibinfo{author}{Westin, C.F.}, \bibinfo{author}{Dyrby, T.B.},
  \bibinfo{author}{Verma, R.}, \bibinfo{author}{Rathi, Y.},
  \bibinfo{author}{Irfanoglu, M.O.}, \bibinfo{author}{Thomas, C.},
  \bibinfo{author}{Pierpaoli, C.}, \bibinfo{author}{Descoteaux, M.},
  \bibinfo{author}{Anderson, A.W.}, \bibinfo{author}{Landman, B.A.},
  \bibinfo{year}{2019}.
\newblock \bibinfo{title}{Limits to anatomical accuracy of diffusion
  tractography using modern approaches}.
\newblock \bibinfo{journal}{NeuroImage} \bibinfo{volume}{185},
  \bibinfo{pages}{1--11}.
\newblock \DOIprefix\doi{10.1016/j.neuroimage.2018.10.029}.
%Type = Article
\bibitem[{Schilling et~al.(2020)Schilling, Petit, Rheault, Remedios, Pierpaoli,
  Anderson, Landman and Descoteaux}]{schilling_brain_2020}
\bibinfo{author}{Schilling, K.G.}, \bibinfo{author}{Petit, L.},
  \bibinfo{author}{Rheault, F.}, \bibinfo{author}{Remedios, S.},
  \bibinfo{author}{Pierpaoli, C.}, \bibinfo{author}{Anderson, A.W.},
  \bibinfo{author}{Landman, B.A.}, \bibinfo{author}{Descoteaux, M.},
  \bibinfo{year}{2020}.
\newblock \bibinfo{title}{Brain connections derived from diffusion {{MRI}}
  tractography can be highly anatomically accurate\textemdash if we know where
  white matter pathways start, where they end, and where they do not go}.
\newblock \bibinfo{journal}{Brain Structure and Function}
  \bibinfo{volume}{225}, \bibinfo{pages}{2387--2402}.
\newblock \DOIprefix\doi{10.1007/s00429-020-02129-z}.
%Type = Article
\bibitem[{Smith et~al.(2020)Smith, Raffelt, Tournier and
  Connelly}]{smith_quantitative_2020}
\bibinfo{author}{Smith, R.}, \bibinfo{author}{Raffelt, D.},
  \bibinfo{author}{Tournier, J.D.}, \bibinfo{author}{Connelly, A.},
  \bibinfo{year}{2020}.
\newblock \bibinfo{title}{Quantitative streamlines tractography: Methods and
  inter-subject normalisation} \DOIprefix\doi{10.31219/osf.io/c67kn}.
%Type = Article
\bibitem[{Smith et~al.(2012)Smith, Tournier, Calamante and
  Connelly}]{smith_anatomically-constrained_2012}
\bibinfo{author}{Smith, R.E.}, \bibinfo{author}{Tournier, J.D.},
  \bibinfo{author}{Calamante, F.}, \bibinfo{author}{Connelly, A.},
  \bibinfo{year}{2012}.
\newblock \bibinfo{title}{Anatomically-constrained tractography: {{Improved}}
  diffusion {{MRI}} streamlines tractography through effective use of
  anatomical information}.
\newblock \bibinfo{journal}{NeuroImage} \bibinfo{volume}{62},
  \bibinfo{pages}{1924--1938}.
\newblock \DOIprefix\doi{10.1016/j.neuroimage.2012.06.005}.
%Type = Article
\bibitem[{Smith et~al.(2015a)Smith, Tournier, Calamante and
  Connelly}]{smith_effects_2015}
\bibinfo{author}{Smith, R.E.}, \bibinfo{author}{Tournier, J.D.},
  \bibinfo{author}{Calamante, F.}, \bibinfo{author}{Connelly, A.},
  \bibinfo{year}{2015}a.
\newblock \bibinfo{title}{The effects of {{SIFT}} on the reproducibility and
  biological accuracy of the structural connectome}.
\newblock \bibinfo{journal}{NeuroImage} \bibinfo{volume}{104},
  \bibinfo{pages}{253--265}.
\newblock \DOIprefix\doi{10.1016/j.neuroimage.2014.10.004}.
%Type = Article
\bibitem[{Smith et~al.(2015b)Smith, Tournier, Calamante and
  Connelly}]{smith_sift2_2015}
\bibinfo{author}{Smith, R.E.}, \bibinfo{author}{Tournier, J.D.},
  \bibinfo{author}{Calamante, F.}, \bibinfo{author}{Connelly, A.},
  \bibinfo{year}{2015}b.
\newblock \bibinfo{title}{{{SIFT2}}: {{Enabling}} dense quantitative assessment
  of brain white matter connectivity using streamlines tractography}.
\newblock \bibinfo{journal}{NeuroImage} \bibinfo{volume}{119},
  \bibinfo{pages}{338--351}.
\newblock \DOIprefix\doi{10.1016/j.neuroimage.2015.06.092}.
%Type = Article
\bibitem[{Smith et~al.(2013)Smith, Tournier, Calamante and
  Connelly}]{smith_sift:_2013}
\bibinfo{author}{Smith, R.E.}, \bibinfo{author}{Tournier, J.D.D.},
  \bibinfo{author}{Calamante, F.}, \bibinfo{author}{Connelly, A.},
  \bibinfo{year}{2013}.
\newblock \bibinfo{title}{{{SIFT}}: {{Spherical-deconvolution}} informed
  filtering of tractograms.}
\newblock \bibinfo{journal}{NeuroImage} \bibinfo{volume}{67},
  \bibinfo{pages}{298--312}.
%Type = Article
\bibitem[{Takemura et~al.(2016)Takemura, Caiafa, Wandell and
  Pestilli}]{takemura_ensemble_2016}
\bibinfo{author}{Takemura, H.}, \bibinfo{author}{Caiafa, C.F.},
  \bibinfo{author}{Wandell, B.A.}, \bibinfo{author}{Pestilli, F.},
  \bibinfo{year}{2016}.
\newblock \bibinfo{title}{Ensemble {{Tractography}}}.
\newblock \bibinfo{journal}{PLoS computational biology} \bibinfo{volume}{12}.
%Type = Article
\bibitem[{Thomas et~al.(2014)Thomas, Ye, Irfanoglu, Modi, Saleem, Leopold and
  Pierpaoli}]{thomas_anatomical_2014}
\bibinfo{author}{Thomas, C.}, \bibinfo{author}{Ye, F.Q.},
  \bibinfo{author}{Irfanoglu, M.O.}, \bibinfo{author}{Modi, P.},
  \bibinfo{author}{Saleem, K.S.}, \bibinfo{author}{Leopold, D.A.},
  \bibinfo{author}{Pierpaoli, C.}, \bibinfo{year}{2014}.
\newblock \bibinfo{title}{Anatomical accuracy of brain connections derived from
  diffusion {{MRI}} tractography is inherently limited}.
\newblock \bibinfo{journal}{Proceedings of the National Academy of Sciences of
  the United States of America} \bibinfo{volume}{111},
  \bibinfo{pages}{16574--16579}.
\newblock \DOIprefix\doi{10.1073/pnas.1405672111}.
%Type = Article
\bibitem[{Tournier et~al.(2007)Tournier, Calamante and
  Connelly}]{tournier_robust_2007}
\bibinfo{author}{Tournier, J.D.}, \bibinfo{author}{Calamante, F.},
  \bibinfo{author}{Connelly, A.}, \bibinfo{year}{2007}.
\newblock \bibinfo{title}{Robust determination of the fibre orientation
  distribution in diffusion {{MRI}}: {{Non-negativity}} constrained
  super-resolved spherical deconvolution}.
\newblock \bibinfo{journal}{NeuroImage} \bibinfo{volume}{35},
  \bibinfo{pages}{1459--1472}.
\newblock \DOIprefix\doi{10.1016/j.neuroimage.2007.02.016}.
%Type = Article
\bibitem[{Tournier et~al.(2019)Tournier, Smith, Raffelt, Tabbara, Dhollander,
  Pietsch, Christiaens, Jeurissen, Yeh and Connelly}]{tournier_mrtrix3_2019}
\bibinfo{author}{Tournier, J.D.}, \bibinfo{author}{Smith, R.},
  \bibinfo{author}{Raffelt, D.}, \bibinfo{author}{Tabbara, R.},
  \bibinfo{author}{Dhollander, T.}, \bibinfo{author}{Pietsch, M.},
  \bibinfo{author}{Christiaens, D.}, \bibinfo{author}{Jeurissen, B.},
  \bibinfo{author}{Yeh, C.H.}, \bibinfo{author}{Connelly, A.},
  \bibinfo{year}{2019}.
\newblock \bibinfo{title}{{{MRtrix3}}: {{A}} fast, flexible and open software
  framework for medical image processing and visualisation}.
\newblock \bibinfo{journal}{NeuroImage} \bibinfo{volume}{202},
  \bibinfo{pages}{116137}.
\newblock \DOIprefix\doi{10.1016/j.neuroimage.2019.116137}.
%Type = Article
\bibitem[{{Van der Maaten} and Hinton(2008)}]{van_der_maaten_visualizing_2008}
\bibinfo{author}{{Van der Maaten}, L.}, \bibinfo{author}{Hinton, G.},
  \bibinfo{year}{2008}.
\newblock \bibinfo{title}{Visualizing data using t-{{SNE}}.}
\newblock \bibinfo{journal}{Journal of Machine Learning Research}
  \bibinfo{volume}{9}.
%Type = Article
\bibitem[{Van~Essen et~al.(2013)Van~Essen, Smith, Barch, Behrens, Yacoub and
  Ugurbil}]{van_essen_wu-minn_2013}
\bibinfo{author}{Van~Essen, D.C.}, \bibinfo{author}{Smith, S.M.},
  \bibinfo{author}{Barch, D.M.}, \bibinfo{author}{Behrens, T.E.J.},
  \bibinfo{author}{Yacoub, E.}, \bibinfo{author}{Ugurbil, K.},
  \bibinfo{year}{2013}.
\newblock \bibinfo{title}{The {{WU-Minn Human Connectome Project}}: {{An}}
  overview}.
\newblock \bibinfo{journal}{NeuroImage} \bibinfo{volume}{80},
  \bibinfo{pages}{62--79}.
\newblock \DOIprefix\doi{10.1016/j.neuroimage.2013.05.041}.
%Type = Article
\bibitem[{Vincent et~al.(2010)Vincent, Larochelle, Lajoie, Bengio, Manzagol and
  Bottou}]{vincent_stacked_2010}
\bibinfo{author}{Vincent, P.}, \bibinfo{author}{Larochelle, H.},
  \bibinfo{author}{Lajoie, I.}, \bibinfo{author}{Bengio, Y.},
  \bibinfo{author}{Manzagol, P.A.}, \bibinfo{author}{Bottou, L.},
  \bibinfo{year}{2010}.
\newblock \bibinfo{title}{Stacked denoising autoencoders: {{Learning}} useful
  representations in a deep network with a local denoising criterion.}
\newblock \bibinfo{journal}{Journal of Machine Learning Research}
  \bibinfo{volume}{11}.
%Type = Article
\bibitem[{Wang et~al.(2018)Wang, Aydogan, Varma, Toga and
  Shi}]{wang_modeling_2018}
\bibinfo{author}{Wang, J.}, \bibinfo{author}{Aydogan, D.B.},
  \bibinfo{author}{Varma, R.}, \bibinfo{author}{Toga, A.W.},
  \bibinfo{author}{Shi, Y.}, \bibinfo{year}{2018}.
\newblock \bibinfo{title}{Modeling topographic regularity in structural brain
  connectivity with application to tractogram filtering}.
\newblock \bibinfo{journal}{NeuroImage} \bibinfo{volume}{183},
  \bibinfo{pages}{87--98}.
\newblock \DOIprefix\doi{10.1016/j.neuroimage.2018.07.068}.
%Type = Inproceedings
\bibitem[{Wang et~al.(2007)Wang, Benner, Sorensen and
  Wedeen}]{wang_diffusion_2007}
\bibinfo{author}{Wang, R.}, \bibinfo{author}{Benner, T.},
  \bibinfo{author}{Sorensen, A.G.}, \bibinfo{author}{Wedeen, V.J.},
  \bibinfo{year}{2007}.
\newblock \bibinfo{title}{Diffusion toolkit: A software package for diffusion
  imaging data processing and tractography}, in:
  \bibinfo{booktitle}{Proceedings of {{International Society}} of {{Magnetic
  Resonance}} in {{Medicine}} ({{ISMRM}})}.
%Type = Article
\bibitem[{Wang et~al.(2019)Wang, Sun, Liu, Sarma, Bronstein and
  Solomon}]{wang_dynamic_2019}
\bibinfo{author}{Wang, Y.}, \bibinfo{author}{Sun, Y.}, \bibinfo{author}{Liu,
  Z.}, \bibinfo{author}{Sarma, S.E.}, \bibinfo{author}{Bronstein, M.M.},
  \bibinfo{author}{Solomon, J.M.}, \bibinfo{year}{2019}.
\newblock \bibinfo{title}{Dynamic {{Graph CNN}} for {{Learning}} on {{Point
  Clouds}}}.
\newblock \bibinfo{journal}{ACM Transactions on Graphics} \bibinfo{volume}{38},
  \bibinfo{pages}{146}.
%Type = Article
\bibitem[{Wasserthal et~al.(2018)Wasserthal, Neher and
  {Maier-Hein}}]{wasserthal_tractseg_2018}
\bibinfo{author}{Wasserthal, J.}, \bibinfo{author}{Neher, P.},
  \bibinfo{author}{{Maier-Hein}, K.H.}, \bibinfo{year}{2018}.
\newblock \bibinfo{title}{{{TractSeg}} - {{Fast}} and accurate white matter
  tract segmentation.}
\newblock \bibinfo{journal}{NeuroImage} \bibinfo{volume}{183},
  \bibinfo{pages}{239--253}.
%Type = Article
\bibitem[{Wasserthal et~al.(2019)Wasserthal, Neher, Hirjak and
  {Maier-Hein}}]{wasserthal_combined_2019}
\bibinfo{author}{Wasserthal, J.}, \bibinfo{author}{Neher, P.F.},
  \bibinfo{author}{Hirjak, D.}, \bibinfo{author}{{Maier-Hein}, K.H.},
  \bibinfo{year}{2019}.
\newblock \bibinfo{title}{Combined tract segmentation and orientation mapping
  for bundle-specific tractography}.
\newblock \bibinfo{journal}{Medical Image Analysis} \bibinfo{volume}{58},
  \bibinfo{pages}{101559}.
\newblock \DOIprefix\doi{10.1016/j.media.2019.101559}.
%Type = Article
\bibitem[{Xia and Shi(2020)}]{xia_groupwise_2020}
\bibinfo{author}{Xia, Y.}, \bibinfo{author}{Shi, Y.}, \bibinfo{year}{2020}.
\newblock \bibinfo{title}{Groupwise track filtering via iterative message
  passing and pruning}.
\newblock \bibinfo{journal}{NeuroImage} \bibinfo{volume}{221},
  \bibinfo{pages}{117147}.
\newblock \DOIprefix\doi{10.1016/j.neuroimage.2020.117147}.
%Type = Article
\bibitem[{Xu et~al.(2019)Xu, Dong, Lee, O'Hara, Asano and
  Jeong}]{xu_objective_2019}
\bibinfo{author}{Xu, H.}, \bibinfo{author}{Dong, M.}, \bibinfo{author}{Lee,
  M.H.}, \bibinfo{author}{O'Hara, N.}, \bibinfo{author}{Asano, E.},
  \bibinfo{author}{Jeong, J.W.}, \bibinfo{year}{2019}.
\newblock \bibinfo{title}{Objective {{Detection}} of {{Eloquent Axonal
  Pathways}} to {{Minimize Postoperative Deficits}} in {{Pediatric Epilepsy
  Surgery Using Diffusion Tractography}} and {{Convolutional Neural
  Networks}}}.
\newblock \bibinfo{journal}{IEEE Transactions on Medical Imaging}
  \bibinfo{volume}{38}, \bibinfo{pages}{1910--1922}.
\newblock \DOIprefix\doi{10.1109/TMI.2019.2902073}.
%Type = Article
\bibitem[{Yang et~al.(2021)Yang, Yeh, Poupon and
  Calamante}]{yang_diffusion_2021}
\bibinfo{author}{Yang, J.Y.M.}, \bibinfo{author}{Yeh, C.H.},
  \bibinfo{author}{Poupon, C.}, \bibinfo{author}{Calamante, F.},
  \bibinfo{year}{2021}.
\newblock \bibinfo{title}{Diffusion {{MRI}} tractography for neurosurgery: The
  basics, current state, technical reliability and challenges}.
\newblock \bibinfo{journal}{Physics in Medicine {$\&$} Biology}
  \DOIprefix\doi{10.1088/1361-6560/ac0d90}.
%Type = Article
\bibitem[{Yeh et~al.(2020)Yeh, Jones, Liang, Descoteaux and
  Connelly}]{yeh_mapping_2020}
\bibinfo{author}{Yeh, C.H.}, \bibinfo{author}{Jones, D.K.},
  \bibinfo{author}{Liang, X.}, \bibinfo{author}{Descoteaux, M.},
  \bibinfo{author}{Connelly, A.}, \bibinfo{year}{2020}.
\newblock \bibinfo{title}{Mapping {{Structural Connectivity Using Diffusion
  MRI}}: {{Challenges}} and {{Opportunities}}}.
\newblock \bibinfo{journal}{Journal of Magnetic Resonance Imaging} ,
  \bibinfo{pages}{jmri.27188}\DOIprefix\doi{10.1002/jmri.27188}.
%Type = Article
\bibitem[{Yeh et~al.(2016)Yeh, Smith, Liang, Calamante and
  Connelly}]{yeh_correction_2016}
\bibinfo{author}{Yeh, C.H.}, \bibinfo{author}{Smith, R.E.},
  \bibinfo{author}{Liang, X.}, \bibinfo{author}{Calamante, F.},
  \bibinfo{author}{Connelly, A.}, \bibinfo{year}{2016}.
\newblock \bibinfo{title}{Correction for diffusion {{MRI}} fibre tracking
  biases: {{The}} consequences for structural connectomic metrics}.
\newblock \bibinfo{journal}{NeuroImage} \bibinfo{volume}{142},
  \bibinfo{pages}{150--162}.
\newblock \DOIprefix\doi{10.1016/j.neuroimage.2016.05.047}.
%Type = Article
\bibitem[{Yeh et~al.(2019)Yeh, Panesar, Barrios, Fernandes, Abhinav, Meola and
  {Fernandez-Miranda}}]{yeh_automatic_2019}
\bibinfo{author}{Yeh, F.C.}, \bibinfo{author}{Panesar, S.},
  \bibinfo{author}{Barrios, J.}, \bibinfo{author}{Fernandes, D.},
  \bibinfo{author}{Abhinav, K.}, \bibinfo{author}{Meola, A.},
  \bibinfo{author}{{Fernandez-Miranda}, J.C.}, \bibinfo{year}{2019}.
\newblock \bibinfo{title}{Automatic {{Removal}} of {{False Connections}} in
  {{Diffusion MRI Tractography Using Topology-Informed Pruning}} ({{TIP}})}.
\newblock \bibinfo{journal}{Neurotherapeutics} \bibinfo{volume}{16},
  \bibinfo{pages}{52--58}.
\newblock \DOIprefix\doi{10.1007/s13311-018-0663-y}.
%Type = Article
\bibitem[{Yendiki et~al.(2011)Yendiki, Panneck, Srinivasan, Stevens,
  Z{\"o}llei, Augustinack, Wang, Salat, Ehrlich, Behrens, Jbabdi, Gollub and
  Fischl}]{yendiki_automated_2011}
\bibinfo{author}{Yendiki, A.}, \bibinfo{author}{Panneck, P.},
  \bibinfo{author}{Srinivasan, P.}, \bibinfo{author}{Stevens, A.},
  \bibinfo{author}{Z{\"o}llei, L.}, \bibinfo{author}{Augustinack, J.},
  \bibinfo{author}{Wang, R.}, \bibinfo{author}{Salat, D.},
  \bibinfo{author}{Ehrlich, S.}, \bibinfo{author}{Behrens, T.},
  \bibinfo{author}{Jbabdi, S.}, \bibinfo{author}{Gollub, R.},
  \bibinfo{author}{Fischl, B.}, \bibinfo{year}{2011}.
\newblock \bibinfo{title}{Automated {{Probabilistic Reconstruction}} of
  {{White-Matter Pathways}} in {{Health}} and {{Disease Using}} an {{Atlas}} of
  the {{Underlying Anatomy}}}.
\newblock \bibinfo{journal}{Frontiers in Neuroinformatics} \bibinfo{volume}{5}.
\newblock \DOIprefix\doi{10.3389/fninf.2011.00023}.
%Type = Article
\bibitem[{Zalesky et~al.(2016)Zalesky, Fornito, Cocchi, Gollo, {van den Heuvel}
  and Breakspear}]{zalesky_connectome_2016}
\bibinfo{author}{Zalesky, A.}, \bibinfo{author}{Fornito, A.},
  \bibinfo{author}{Cocchi, L.}, \bibinfo{author}{Gollo, L.L.},
  \bibinfo{author}{{van den Heuvel}, M.P.}, \bibinfo{author}{Breakspear, M.},
  \bibinfo{year}{2016}.
\newblock \bibinfo{title}{Connectome sensitivity or specificity: Which is more
  important?}
\newblock \bibinfo{journal}{NeuroImage} \bibinfo{volume}{142},
  \bibinfo{pages}{407--420}.
\newblock \DOIprefix\doi{10.1016/j.neuroimage.2016.06.035}.
%Type = Article
\bibitem[{Zhang et~al.(2020)Zhang, Cetin~Karayumak, Hoffmann, Rathi, Golby and
  O'Donnell}]{zhang_deep_2020}
\bibinfo{author}{Zhang, F.}, \bibinfo{author}{Cetin~Karayumak, S.},
  \bibinfo{author}{Hoffmann, N.}, \bibinfo{author}{Rathi, Y.},
  \bibinfo{author}{Golby, A.J.}, \bibinfo{author}{O'Donnell, L.J.},
  \bibinfo{year}{2020}.
\newblock \bibinfo{title}{Deep white matter analysis ({{DeepWMA}}): {{Fast}}
  and consistent tractography segmentation}.
\newblock \bibinfo{journal}{Medical Image Analysis} \bibinfo{volume}{65},
  \bibinfo{pages}{101761}.
\newblock \DOIprefix\doi{10.1016/j.media.2020.101761}.
%Type = Article
\bibitem[{Zhang et~al.(2022)Zhang, Daducci, He, Schiavi, Seguin, Smith, Yeh,
  Zhao and O'Donnell}]{zhang_quantitative_2022}
\bibinfo{author}{Zhang, F.}, \bibinfo{author}{Daducci, A.},
  \bibinfo{author}{He, Y.}, \bibinfo{author}{Schiavi, S.},
  \bibinfo{author}{Seguin, C.}, \bibinfo{author}{Smith, R.E.},
  \bibinfo{author}{Yeh, C.H.}, \bibinfo{author}{Zhao, T.},
  \bibinfo{author}{O'Donnell, L.J.}, \bibinfo{year}{2022}.
\newblock \bibinfo{title}{Quantitative mapping of the brain's structural
  connectivity using diffusion {{MRI}} tractography: {{A}} review}.
\newblock \bibinfo{journal}{NeuroImage} \bibinfo{volume}{249},
  \bibinfo{pages}{118870}.
\newblock \DOIprefix\doi{10.1016/j.neuroimage.2021.118870}.
%Type = Article
\bibitem[{Zhang et~al.(2018)Zhang, Wu, Norton, Rigolo, Rathi, Makris and
  O'Donnell}]{zhang_anatomically_2018}
\bibinfo{author}{Zhang, F.}, \bibinfo{author}{Wu, Y.}, \bibinfo{author}{Norton,
  I.}, \bibinfo{author}{Rigolo, L.}, \bibinfo{author}{Rathi, Y.},
  \bibinfo{author}{Makris, N.}, \bibinfo{author}{O'Donnell, L.J.},
  \bibinfo{year}{2018}.
\newblock \bibinfo{title}{An anatomically curated fiber clustering white matter
  atlas for consistent white matter tract parcellation across the lifespan}.
\newblock \bibinfo{journal}{NeuroImage} \bibinfo{volume}{179},
  \bibinfo{pages}{429--447}.
\newblock \DOIprefix\doi{10.1016/j.neuroimage.2018.06.027}.

\end{thebibliography}
}

\clearpage
\appendix

\section{Additional Experiments with Verifyber}

\subsection{Verifyber Test on BIL\&GIN data}
\label{suppl:vf_gin}

\paragraph{BILGIN-EP} BIL\&GIN~\citep{mazoyer_bilgin_2016} data source. Exclusive labeling~\citep{petit_structural_2022}. Single subject, with tractogram of 1.5M streamlines obtained via PF-ACT tractography~\citep{girard_cortical_2020}. 

\newcolumntype{C}{>{\centering\arraybackslash}p{1.15cm}}
\begin{table}[h]
	\caption[Generalization test on BILGIN-EP]{Generalization test on BILGIN-EP. Models have been trained on one of the 5-fold split of HCP-EP. We do not report standard deviation, as BILGIN-EP is a single-subject dataset. The results confirm Veryfiber as best performing model. It maintains very high precision, while it loose some points in terms of recall. This is probably due to the different distribution of plausible/non-plausible.    
    }
    \label{tab:res_bilgin}
    \centering
    \begin{tabular}{ l | C | C | C | C }
        \toprule
        Method & Accuracy & Precision & Recall & DSC \\
        \midrule
        bLSTM   & 90.7 & 94.4 & 93.1 & 93.7 \\
        PN      & 91.2 & 95.1 & 93.1 & 94.1 \\
        DGCNN     & 91.4 & 95.5 & 92.8 & 94.1 \\
        \textbf{VF} & \textbf{92.5} & \textbf{96.2} & \textbf{93.7} & \textbf{94.9} \\
        \bottomrule
    \end{tabular}
    
\end{table}
%\FloatBarrier

\subsection{Training Verifyber with Signal-based Supervision}
\label{suppl:vf_sift}

\begin{figure*}[h]
\centering
  \includegraphics[width=.835\textwidth]{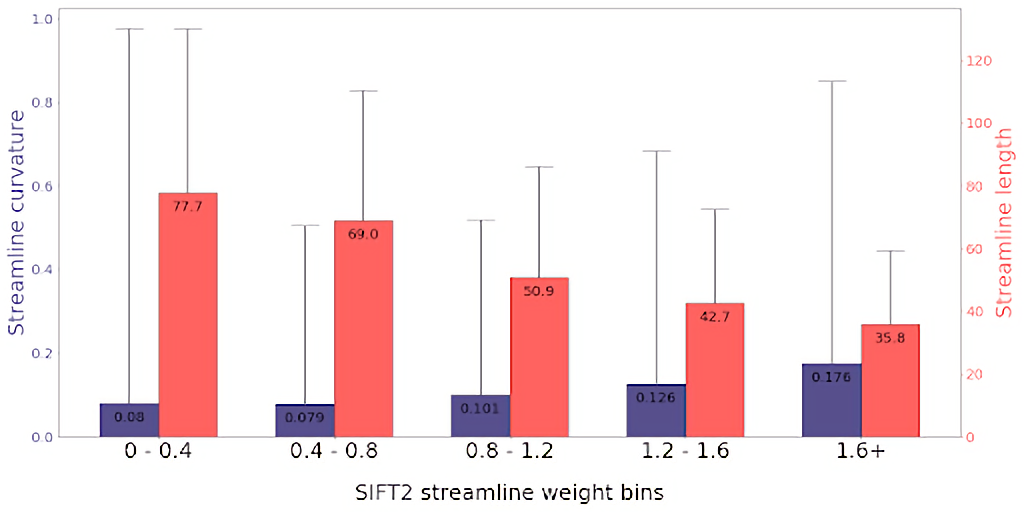}
  \caption[SIFT2 weights per streamline categories]{SIFT2 weights per streamline categories in a single tractogram. Relationship between streamline anatomy and SIFT2 streamline weights. Each weight interval  contains at least 10\% of fibers. The trend of length and curvature shows that high-weight is assigned to very curved and short fibers. Viceversa, low-weight is given to straight and long fibers. This behavior is almost opposite compared to the anatomical labeling of Extractor~\citep{petit_structural_2022}. Note also a very high standard deviation.}
  \label{fig:sift_sl_categories}
\end{figure*}

\begin{figure*}[h]
\centering
  \includegraphics[width=\textwidth]{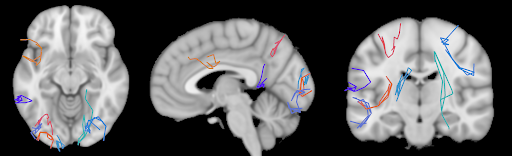}
  \caption[Visual example of non-plausible fibers according to SIFT2]{Visual example of non-plausible fibers according to SIFT2. 8 streamlines with weights $> 3$, lengths $> 92$ mm, and curvatures $> 2$ (MNI space, downsampled to 16 points). The figure shows that SIFT2 weights are not a good indicator for the anatomical plausibility of a streamline.}
  \label{fig:sift_visual_np}
\end{figure*}

\begin{figure*}[h]
\centering
  \includegraphics[height=.95\textheight]{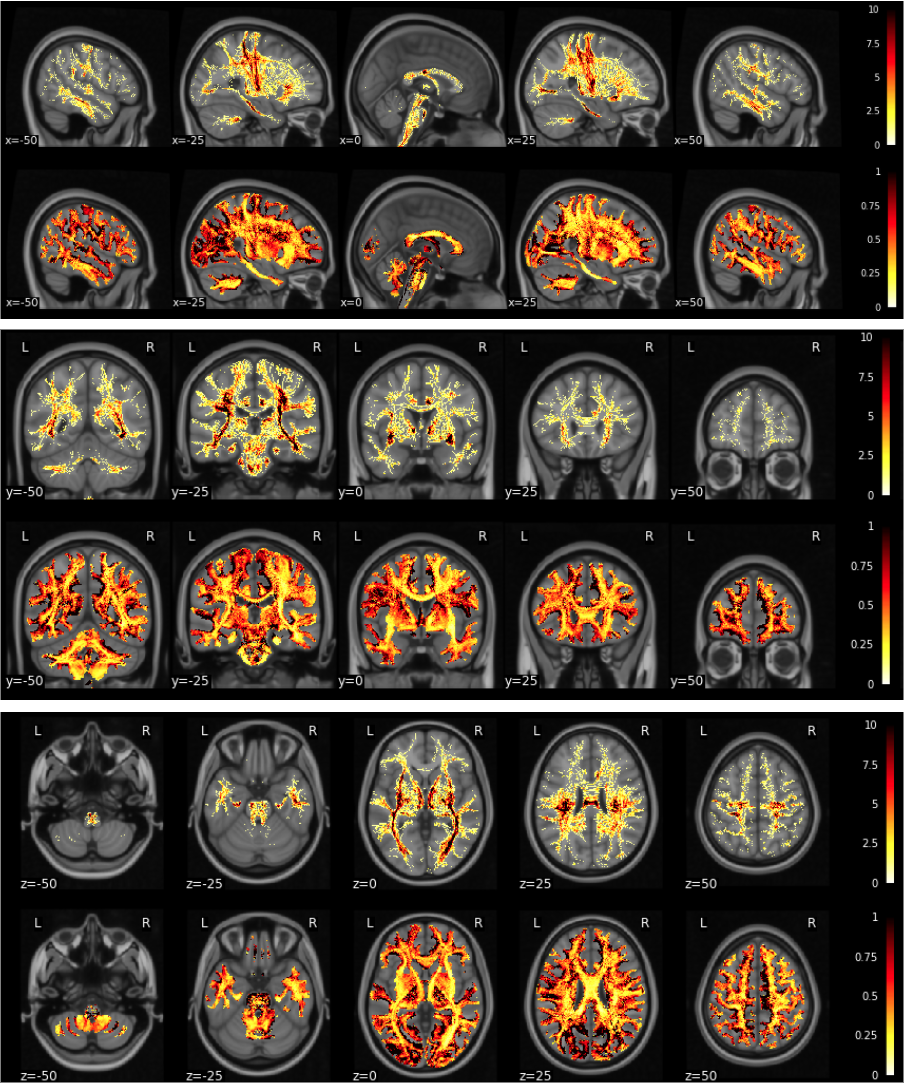}
  \caption[Spatial distribution of SIFT2 weights vs. fiber density]{Spatial distribution of SIFT2 weights vs. fiber density. SIFT weights distribution (top rows) compared to streamline density map (bottom rows). Observing the two types of maps, it is easy to see the high degree of correlation.}
  \label{fig:sift_density}
\end{figure*}

\outline{adoption of SIFT2 as alternative labeling}
In this study, we adopted fiber {\em weights} obtained with SIFT2~\citep{smith_sift2_2015} as target for Verifyber, which was trained for regression. The results suggested that VF can learn weights regression decently well, but cannot capture the density-based filtering criteria. In fact, with tractograms having similar fiber density map, VF was able to predict weights, but as soon as new tractograms had different fiber density our methods failed. See Figures~\ref{fig:sift_sl_categories}, \ref{fig:sift_visual_np}, \ref{fig:sift_density}

\begin{table}[h]
\centering
\caption[Regression results on SIFT2 weights for HCP-EP]{Regression results on SIFT2 weights for HCP-EP. Both Verifyber and PointNet~\citep{qi_pointnet_2017} were trained on the training subjects of HCP-EP, for which we computed the SIFT2 weights. The table shows the test results averaged over 4 test subjects in terms of Mean Absolute Error(MAE) --- the same metric optimized during training ---, and Spearman's rank correlation coefficient. The former indicate the absolute regression error, while the latter indicate how much the ranking of fibers obtained for SIFT2 weights is preserved after weights prediction.}
\label{tab:sift2_results}
\begin{tabular}{@{}cccc@{}}
\toprule
Model & CSD & MAE & Spear. Corr. \\ \midrule
Verifyber & DiPy & 0.173 ($\pm$0.011) & 0.766 ($\pm$0.039) \\
PointNet & DiPy & 0.175 ($\pm$0.014) & 0.757 ($\pm$0.033) \\ \bottomrule
\end{tabular}%

\end{table}

\outline{exp and results}

\outline{exp and results}

\begin{figure*}[h]
\centering
  \includegraphics[height=.95\textheight]{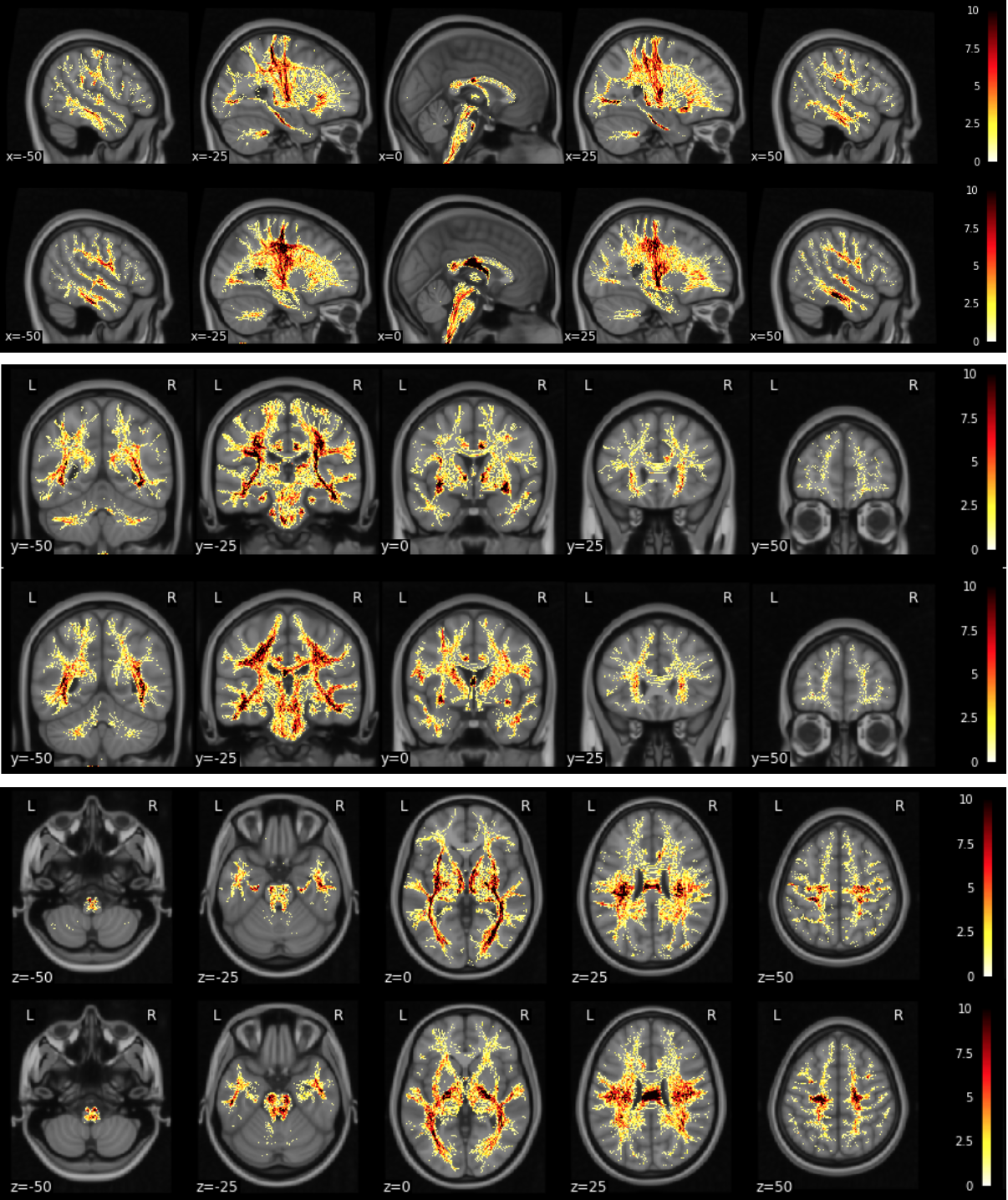}
  \caption[Fiber density map after different tractography algorithms]{Fiber density map obtained after Ensemble Tractography~\citep{takemura_ensemble_2016} (top rows) and Particle Filtering tractography~\citep{girard_towards_2014} (bottom rows). This figure shows that we cannot expect generalization of VF$^{\mathrm{SIFT2}}$ across the two kind of tractograms.}
  \label{fig:density_comparison_track}
\end{figure*}

\FloatBarrier
\nobalance

\section{Investigation of FINTA}

\begin{figure*}[b!]
\centering
  \includegraphics[width=.42\textwidth]{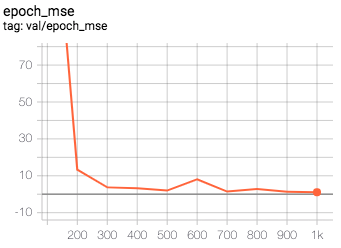}
  \includegraphics[width=.42\textwidth]{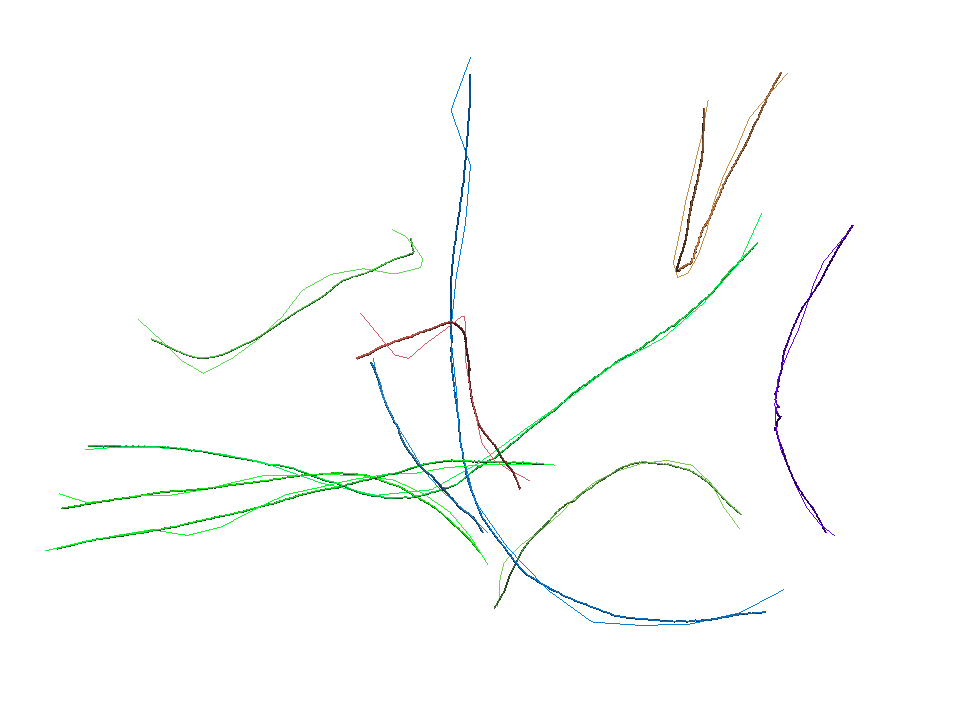}
  \caption[Reproduced FINTA autoencoder]{Reproduced FINTA autoencoder. The plot on the left refer to a training performed on the dataset HCP-IZ. In this experiment, according to the setting used by the Authors of FINTA, we split the streamlines of the HCP-IZ tractogram into 80/20 train/test. The train set was then split in 4 buckets, 3 used for train and 1 for validation, which is performed once every 100 epochs. The training is stopped after 1000 epochs as it has converged to an MSE value in validation $\sim$1mm. The image on the right depicts the qualitative reconstruction using the trained autoencoder. Thin fibers are the original, while fat are the reconstructed. It can be observed that the reconstruction is smooth and reasonably correct.}
  \label{fig:finta_train}
\end{figure*}

\begin{figure*}[t]
\centering
  \vspace{-.8cm}
  \includegraphics[scale=0.4]{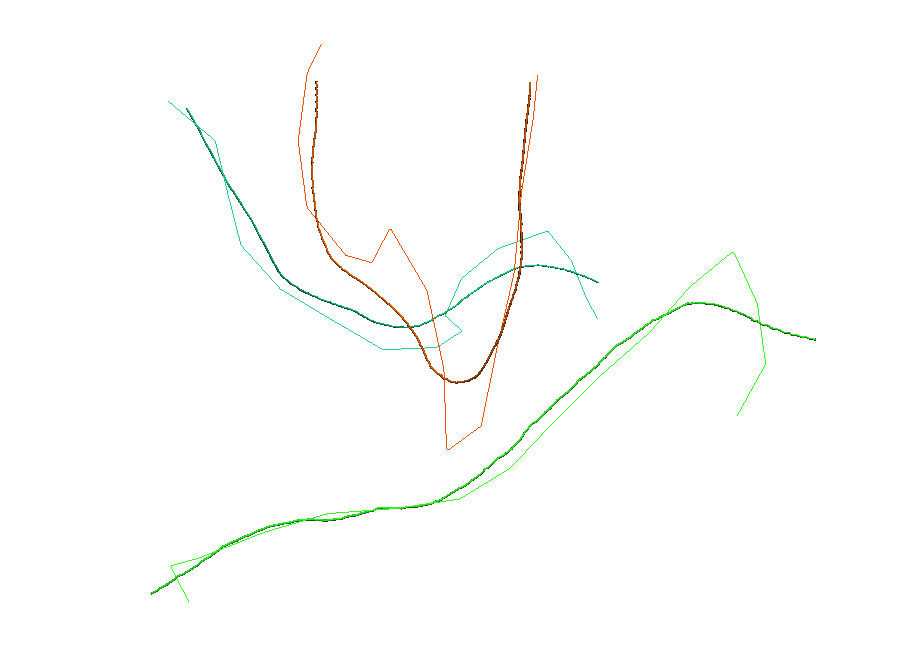}
  \vspace{-1cm}
  \caption[Non-plausible fibers (HCP-EP) reconstructed using FINTA autoencoder]{Non-plausible fibers (HCP-EP) reconstructed using FINTA autoencoder. As can be noted original streamlines (thin) have a clear non-plausible trajectory with sharp curves. The denoising action performed by the autoencoder reconstruct smoother trajectories (fat), which we may observe to be more similar to plausible than to non-plausible fibers. Such denoising behaviour is expected when adopting an autoencoder model. However, as the output reconstructions reflect the learned embedding, such a behaviour can be a source of error for the task of tractogram filtering.}
  \label{fig:finta_smooth}
\end{figure*}

\begin{figure*}[b]
\centering
  \includegraphics[width=.835\textwidth]{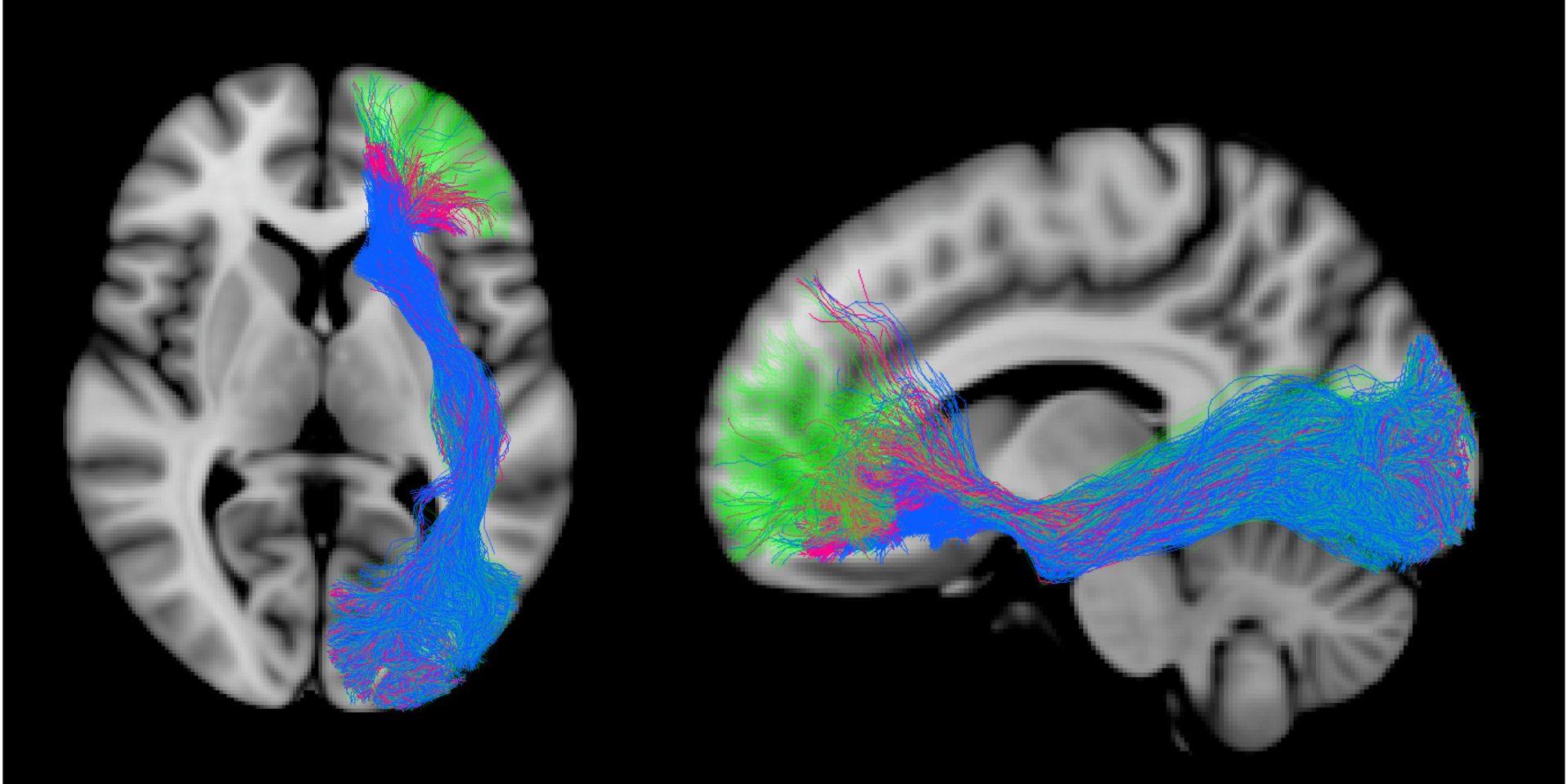}
  \caption[Qualitative FINTA versus Verifyber]{IFOF left from HCP-IW \cite{wasserthal_tractseg_2018} after classification of streamlines as plausible ($p$) and non-plausible ($np$) using FINTA trained on HCP-IZ \cite{zhang_anatomically_2018}. Considering the same classification performed with Verifyber (see Figure 16 of the paper) as a reference labeling, we investigate the possible false positive prediction of FINTA i.e,. fibers that are $np$ according to Verifyber, but that FINTA classifies as $p$. In the figure we indicate such fibers in fuchsia, while we color with green and blue fibers classified by both methods as $p$ and $np$ respectively.}
  \label{fig:finta_ifof}
%  Left: FINTA$\mathrm{^{IZ}_{BAcc}}$, right: FINTA$\mathrm{^{IZ}_{DSC}}$. First row: FINTA prediction, green are {\em p}, blue {\em np}. Second row: Verifyber predicted {\em np}, yellow are predicted as {\em p} by FINTA, blue as {\em np}. Third row: Verifyber predicted {\em p}, dark blue are predicted as {\em np} by FINTA, green as {\em p}.}
\end{figure*}

Additional results obtained with the reproduced FINTA~\citep{legarreta_filtering_2021}. See Figures~\ref{fig:finta_train}, \ref{fig:finta_smooth}, \ref{fig:finta_ifof}.
%\onecolumn

\end{document}